\documentclass[10pt,journal,compsoc]{IEEEtran}
\newcommand{\HL}[1]{\textcolor{blue}{#1}}
\newcommand{\HLR}[1]{\textcolor{red}{#1}}

\usepackage[pdftex]{graphicx}
\usepackage{subfigure}
\usepackage{cancel}
\usepackage{pdfsync}
\usepackage{balance}
\usepackage{color}
\usepackage{mathtools}
\usepackage{algpseudocode}
\usepackage{amssymb,mathrsfs,amsmath}
\usepackage{bm}
\usepackage{subfigure}
\usepackage{diagbox}
\usepackage{pifont}
\usepackage{fixltx2e}
\usepackage{multirow}
\usepackage{url}
\usepackage{hyperref}
\usepackage{verbatim}
\usepackage{mathabx}	
\usepackage{physics}
\usepackage{booktabs}
\usepackage{siunitx}
\usepackage{caption}
\usepackage{rotating}
\usepackage{color, colortbl} 

\usepackage[margin=4pt,font=footnotesize,labelfont=bf,labelsep=endash,tableposition=top]{caption}

\usepackage[flushleft]{threeparttable}
\ifCLASSOPTIONcompsoc
  \usepackage[nocompress]{cite}
\else
  \usepackage{cite}
\fi
\graphicspath{{figures/}}
\DeclareGraphicsExtensions{.png,.jpg,.eps,.pdf}

\ifCLASSINFOpdf
\else
\fi
\usepackage{url}

\begin{document}
%
\title{R$^3$LIVE++: A Robust, Real-time, Radiance reconstruction package with a tightly-coupled LiDAR-Inertial-Visual state Estimator}
%
%
%
%

\author{ Jiarong Lin and Fu Zhang
\IEEEcompsocitemizethanks{\IEEEcompsocthanksitem J. Lin and F. Zhang are with the Department of Mechanical Engineering, The University of Hong Kong, Hong Kong SAR, China.\protect\\
E-mail: $\{$jiarong.lin,  fuzhang$\}$@hku.hk
}
}

\IEEEtitleabstractindextext{
\setcounter{figure}{0}
\includegraphics[width=0.95\textwidth]{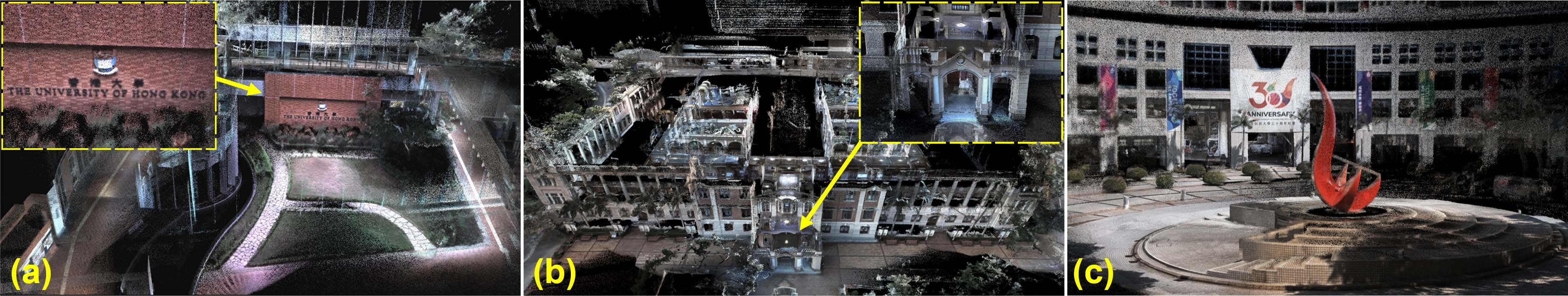}
\vspace{-0.7cm}
\captionof{figure}{The radiance map of HKU (a and b) and HKUST campuses (c) reconstructed by R$^3$LIVE++ in real-time (see our accompanying video on YouTube: \href{https://youtu.be/qXrnIfn-7yA}{\tt{youtu.be/qXrnIfn-7yA}}).}
\label{fig_main_cover}
\vspace{0.3cm}
\begin{abstract}
This work proposed a LiDAR-inertial-visual fusion framework termed R$^3$LIVE++ to achieve robust and accurate state estimation while simultaneously reconstructing the radiance map on the fly. R$^3$LIVE++ consists of a LiDAR-inertial odometry (LIO) and a visual-inertial odometry (VIO), both running in real-time. The LIO subsystem utilizes the measurements from a LiDAR for reconstructing the geometric structure, while the VIO subsystem simultaneously recovers the radiance information of the geometric structure from the input images. R$^3$LIVE++ is developed based on R$^3$LIVE and further improves the accuracy in localization and mapping by accounting for the camera photometric calibration and the online estimation of camera exposure time. We conduct more extensive experiments on public and private datasets to compare our proposed system against other state-of-the-art SLAM systems. Quantitative and qualitative results show that R$^3$LIVE++ has significant improvements over others in both accuracy and robustness. Moreover, to demonstrate the extendability of R$^3$LIVE++, we developed several applications based on our reconstructed maps, such as high dynamic range (HDR) imaging, virtual environment exploration, and 3D video gaming. Lastly, to share our findings and make contributions to the community, we release our codes, hardware design, and dataset on our Github: \href{https://github.com/hku-mars/r3live}{\tt github.com/hku-mars/r3live}.
\end{abstract}

}

\maketitle

\iftrue
\section{Introduction}
\IEEEPARstart{S}{imultaneous} localization and mapping (SLAM) is a technology that utilizes a sequence of sensor (e.g., camera, LiDAR, IMU, etc.) data to estimate the sensor poses and simultaneously reconstruct the 3D map of surrounding environments. Since SLAM can estimate poses in real-time, it has been widely applied in localization and feedback control for autonomous robots (e.g., unmanned aerial vehicles\cite{gao2019flying, kong2021avoiding}, automated ground vehicles\cite{lategahn2011visual, beinschob2015graph,lin2020decentralized}, and self-driving cars\cite{levinson2011towards, ros2012visual, singandhupe2019review}). Meanwhile, with the capacity to reconstruct the map in real-time, SLAM is also crucial in various robots navigation, virtual and augmented reality (VR/AR), surveying, and mapping applications. Different applications usually require a different level of mapping details: sparse feature map, 3D dense point cloud map, and 3D radiance map (i.e.,  a 3D point cloud map with radiance information). For example, the sparse visual feature map is suitable and has been widely used for camera-based localization, where the sparse features observed in images can be used for calculating the camera's pose\cite{pink2008visual, burki2018map}. The 3D dense point cloud can capture the geometrical structure of the environment even for tiny objects. Hence it is widely used in robot navigation and obstacle avoidance\cite{nagy2018real, kong2021avoiding}. Finally, radiance maps containing both geometry and radiance information are used in mobile mapping, AR/VR, video gaming, 3D simulation,  and surveying. These applications require both geometric structures and textures to provide virtual environments alike the real world\cite{li2019aads, bao2022high}.

Existing SLAM systems can be mainly categorized into two classes based on the used sensor: visual SLAM and LiDAR SLAM. Visual SLAM is based on low-cost and SWaP (size, weight, and power)-efficient camera sensors and has achieved satisfactory results in localization accuracy. The rich colorful information measured by cameras also makes the reconstructed map suitable for human interpretation. However, due to the lack of direct accurate depth measurements, the mapping accuracy and resolution of visual SLAM are usually lower than LiDAR SLAM. To be more specific, visual SLAM maps the environments by triangulating disparities from multi-view images (e.g., structure from motion for mono-camera, stereo-vision for stereo-camera), an extremely computationally expensive process that often requires hardware acceleration or server clusters. Moreover, limited by the measurement noises and the baseline of multi-view images, the computed depth accuracy drops quadratically with the measurement distance, making visual SLAM difficult to reconstruct large-scale outdoor scenes. Furthermore, visual SLAM can only work in scenarios with good illuminations and will degenerate in high-occlusion or texture-less environments.  

On the other hand, LiDAR SLAM is based on LiDAR sensors. Benefiting from the high measurement accuracy (a few centimeters) and the long measurement range  (hundreds of meters) of LiDAR sensors, LiDAR SLAM can achieve much higher accuracy and efficiency on both localization and map reconstruction than visual SLAM. However, LiDAR SLAM easily fails in scenarios with insufficient geometry features, such as in long tunnel-like corridors, facing a single big wall, etc. Moreover, LiDAR SLAM can only reconstruct the geometric structure of the environment, but lacks the color information.

Fusing both LiDAR and camera measurements in the SLAM could overcome the degeneration issues of each sensor in localization and produce an accurate, textured, and high-resolution 3D map that suffices the needs of various mapping applications.  Motivated by this, we propose R$^3$LIVE++, which has the following features:

\begin{itemize}
	\item It is a LiDAR-Inertial-Visual fusion framework that tightly couples two subsystems: the LiDAR-inertial odometry (LIO) subsystem and the visual-inertial odometry (VIO) subsystem. The two subsystems jointly and incrementally build a 3D radiance map of the environment in real-time. In particular, the LIO subsystem reconstructs the geometric structure by registering new points in each LiDAR scan to the map, and the VIO subsystem recovers the radiance information by rendering pixel colors in each image to points in the map.
	\item It has a novel VIO design, which tracks the camera pose (and estimates other system state) by minimizing the radiance difference between points from the radiance map and a sparse set of pixels in the current image. The frame-to-map alignment effectively lowers the odometry drift, and the direct photometric error on a sparse set of individual pixels effectively constrains the computation load. Moreover, based on the photometric errors, the VIO is able to estimate the camera exposure time online, which enables the recovery of environments' true radiance information.  
	\item It is extensively validated in real-world experiments in terms of localization accuracy, robustness, and radiance map reconstruction accuracy. Benchmark results on 25 sequences from an open dataset (the NCLT-dataset) show that R$^3$LIVE++ achieves the highest overall accuracy among all other state-of-the-art SLAM systems (e.g., LVI-SAM, LIO-SAM, FAST-LIO2, etc) under comparison. The evaluations on our private dataset show that R$^3$LIVE++ is robust to extremely challenging scenarios that LiDAR and/or camera measurements degenerate (e.g., when the device is facing a single texture-less wall). Finally, compared with other counterparts, R$^3$LIVE++ estimates the camera exposure time more accurately and reconstructs the true radiance information of the environment with significantly smaller errors when compared to the measured pixels in images. 
	\item It is, to our best knowledge, the first radiance map reconstruction framework that can achieve real-time performance on a PC equipped with a standard CPU without any hardware or GPU accelerations. The system is completely open sourced to ease the reproduction of this work and benefit the following-up researches. Based on a set of offline utilities for mesh reconstruction and texturing further developed, the system shows high potentials in a variety of real-world applications, such as 3D HDR imaging, physics simulation, and video gaming. 
\end{itemize}

\section{Related Works}\label{sect_intro}
In this chapter, we review existing works related to our method or system, including LiDAR SLAM, visual SLAM, and LiDAR-visual fused SLAM. Due to the large number of existing works, any attempts to give a full review would be incomplete, hence we only select the most relevant ones of each branch for review.

\subsection{LiDAR(-inertial) SLAM} \label{sect_LiDAR_Inertial}
In recent years, with the rapid development of LiDAR technologies, the reliability and performance of LiDAR sensors have been greatly improved while the cost significantly lowered, drawing increasing amount of researches on LiDAR SLAM\cite{lin2020loam}. Zhang \textit{et al.} propose a real-time LiDAR odometry and mapping framework, LOAM \cite{zhang2014loam}, which achieves localization by scan-to-scan point registration and mapping by scan-to-map registration. In both registrations, only edge and plane feature points are considered to lower the computation load. To make the algorithm run in real-time at computation-limited platforms, Shan \textit{et al.} \cite{shan2018lego} propose a lightweight and ground-optimized LOAM (LeGO-LOAM), which discards unreliable features in the step of ground plane segmentation. These works \cite{zhang2014loam ,shan2018lego} are mainly based on multi-line spinning LiDARs. For emerging solid-state LiDARs that have irregular scanning and very small FoV, our previous works \cite{lin2020loam, lin2019fast} use direct scan-to-map registration to achieve localization and mapping. 

To further improve the accuracy and robustness of LiDAR SLAM systems, many frameworks that fuse LiDAR measurements with inertial sensors have been proposed. In LOAM \cite{zhang2014loam}, an IMU could be used to de-skew the LiDAR scan and give a motion prior for the scan-to-scan registration. It is a loosely-coupled method since the IMU bias (and the full state vector) is not involved in the scan registration process. Compared with loosely-coupled methods, tightly-coupled methods show higher robustness and accuracy, therefore drawing increasing research interests recently. Authors in \cite{ye2019tightly} propose LIOM, which uses a graph optimization based on priors from LiDAR-Inertial odometry and a rotation-constrained refinement method. Compared with the former algorithms, LIO-SAM \cite{shan2020lio} optimizes a sliding window of keyframe poses in a factor graph to achieve higher accuracy. Similarly, Li \textit{et al.} propose LiLi-OM \cite{li2020towards} for both multi-line and solid-state LiDARs based on a sliding window optimization technique. LINS \cite{qin2020lins} is the first tightly-coupled LIO that solves the 6 DOF ego-motion via iterated Kalman filtering. To lower the high computation load in calculating the Kalman gain, FAST-LIO \cite{xu2020fast} proposes a new formula for the Kalman gain computation. The resultant computation complexity depends on the state dimension instead of measurement dimension. Its successor FAST-LIO2\cite{fastlio2} further improves the computation efficiency by proposing an incremental k-d tree. Such a data structure can significantly reduce the time cost of nearest points search and allow the registration of raw points (instead of feature points, such as planes and edges, in the past works). The method using raw points is termed as a ``direct" method and could exploit subtle features in the environment and hence increase the localization accuracy and robustness. 

The LIO subsystem of R$^3$LIVE++ is largely based on  FAST-LIO2\cite{fastlio2} since it achieves the best overall performance among its counterparts in terms of accuracy, efficiency, and robustness. Moreover, to address the LiDAR degeneration problem and further improve the localization accuracy, we fuse the LIO subsystem with our VIO subsystem in a tightly-coupled manner.

\subsection{Visual(-inertial) SLAM}
Depending on how a camera measurement is formulated, we review the works of visual SLAM by categorizing them into two branches based on the criteria proposed in \cite{engel2017direct}: indirect and direct. These two types of methods have very different pipelines: the former one (indirect method) includes feature extraction, data association, and minimization of feature re-projection error. In contrast, the latter one (direct method) directly minimizes the photometric error (or intensity discrepancy) between consecutive images.

Indirect visual SLAM is also called the feature-based method, which has a quite long history. MonoSLAM\cite{davison2007monoslam} proposed by Davison \textit{et al.} is the first monocular visual SLAM, which recovers the 3D trajectory of a camera in real-time by creating a sparse but persistent map of natural landmarks within a probabilistic framework. PTAM \cite{klein2007parallel} proposed by Klein and Murray split the tracking and mapping in parallel threads. Visual landmarks in the map are selected from only a few frames to allow efficient bundle-adjustment (BA) optimization that estimates the camera pose and landmark position. Following this idea, a more complete and reliable framework ORB-SLAM\cite{mur2015orb} was proposed. ORB-SLAM utilizes the same feature (i.e., ORB feature) for all the involved tasks, including tracking, mapping, relocalization, and loop closing. Its further work ORB-SLAM2\cite{mur2017orb} improves the accuracy by utilizing the metric scale provided by stereo or RGB-D cameras. The scale issue in pure visual SLAM can also be addressed by fusing inertial sensor data, such as VINS-mono\cite{qin2018vins} and ORB-SLAM3\cite{campos2021orb}, which achieve high-accuracy localization by fusing the IMU measurements and image features in a sliding window bundle adjustment optimization.

Direct visual SLAM is also called photometric-based method, which minimizes the intensity differences rather than a geometric error. It has been successfully applied in 2D sparse feature tracking (e.g., Lucas–Kanade optical flow\cite{lucas1981iterative}) and then extended to visual SLAM. LSD-SLAM\cite{engel2014lsd} proposed by Engel \textit{et al.} is a direct monocular SLAM algorithm with both tracking and mapping directly operating on image intensities. It incrementally tracks the camera pose using direct image alignment and simultaneously performs a pose graph optimization to keep the entire camera trajectory globally consistent. In DSO\cite{engel2017direct}, authors proposed a fully direct probabilistic model that integrates a full photometric calibration. By incorporating a  photometric bundle adjustment, the system outperforms other state-of-the-art works in terms of both accuracy and robustness. To achieve real-time performance on a standard CPU, the authors also exploit the sparsity structures of the corresponding Hessian matrix. While the photometric model provides accurate pose estimation over short-term tracking without data association, the geometric model gives robustness for a large baseline. Hybrid approaches that use both photometric and geometric errors have been proposed, with the most representative work SVO\cite{forster2014svo} proposed by Forster \textit{et al.}, where the short-term tracking is solved by minimizing the photometric error, while the long-term drift is constrained by a windowed bundle adjustment on visual landmarks.

There have been many discussions in the literature to answer the question: \textit{Which is better?} While it is difficult to answer this question now, it is true that the direct method often shows better short-term performance in low-textured environments \cite{engel2017direct, campos2021orb}. Besides, the direct method is often more computation efficient due to the removal of feature extraction \cite{forster2014svo}. To leverage these advantages, R$^3$LIVE++ uses a photometric-based VIO subsystem. Unlike the pure visual (or visual-inertial) direct SLAM systems, which perform bundle adjustment on photometric errors \cite{engel2017direct} or feature reprojection errors \cite{forster2014svo} to restrain long-term drift, the VIO in R$^3$LIVE++ makes fully use of the geometry structure reconstructed from LiDAR point cloud by minimizing the radiance errors between map points and image pixels. Such a frame-to-map alignment effectively lowers the odometry drift at a low computation cost. Moreover, pure visual (or visual-inertial) direct methods construct photometric errors on dense images \cite{engel2014lsd} or a sparse set of image patches \cite{engel2017direct, forster2014svo}, while the photometric errors of R$^3$LIVE++ VIO subsystem are on a sparse set of individual pixels. Furthermore, the VIO in R$^3$LIVE++  accounts for the camera photometric calibration (e.g., non-linear response function and lens vignetting) and estimates the camera exposure time online, which help improve the odometry accuracy and recover the true radiance information of the environment. 

\begin{figure*}
	\centering
	\includegraphics[width=0.95\linewidth]{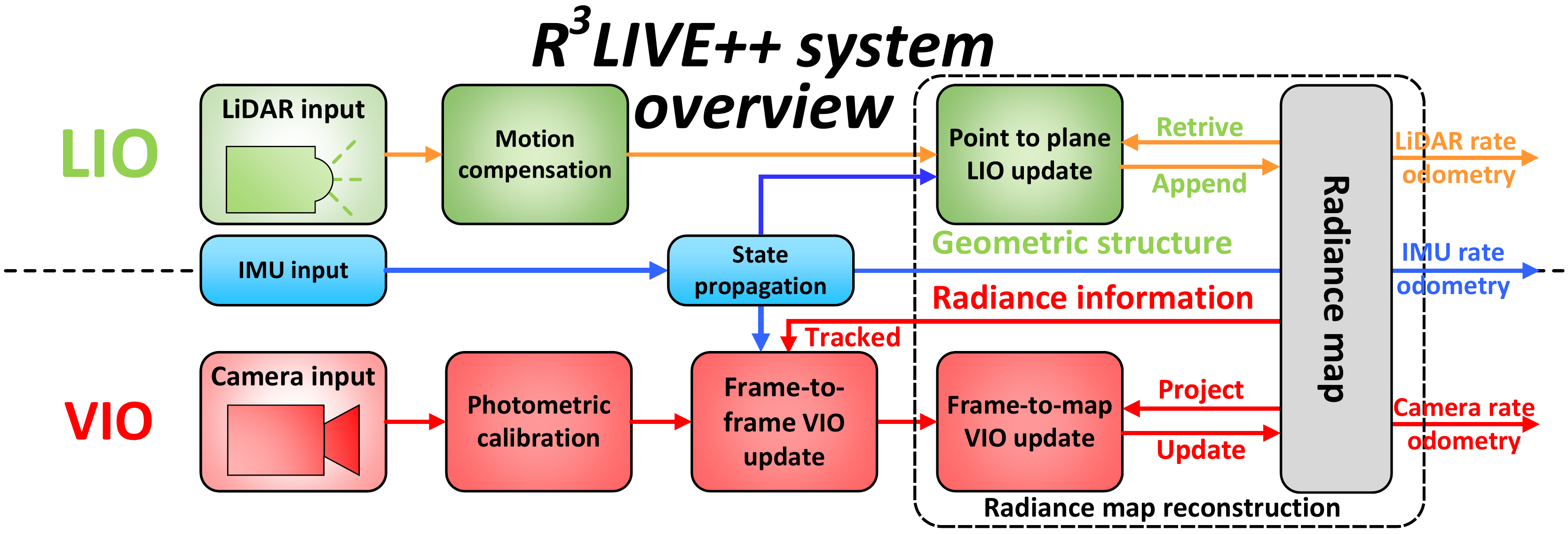}
	\vspace{-0.2cm}
	\caption{The overview of our proposed system.}
	\label{fig_overview}
\end{figure*}

\subsection{LiDAR-visual fused SLAM} \label{sect_LiDAR_Inertial_Visual}

On the basis of LiDAR-inertial methods, LiDAR-inertial-visual odometry incorporating measurements from visual sensors shows higher robustness and accuracy.  Zhang and Singh in \cite{zhang2018laser} propose a LiDAR-inertial-visual system that uses a loosely-coupled VIO as the motion model to initialize the LiDAR mapping subsystem. Similarly, Shao \textit{et al.} in \cite{shao2019stereo} propose a stereo visual-inertial LiDAR SLAM that incorporates the tightly-coupled stereo VIO with LiDAR mapping and LiDAR-enhanced visual loop closure. The overall system is still a loosely-coupled fusion since the LiDAR data are not jointly optimized along with the visual-inertial measurements.

There are also some RGB-D-inertial SLAM systems, such as \cite{laidlow2017dense, shao2019stereo}. Designed for RGB-D cameras, these methods are difficult to be applied on LiDARs due to the significant differences in the measurement pattern, range and density between RGB-D cameras and LiDAR sensors.
In \cite{zhu2020camvox}, LiDAR measurements are used to provide depth information for camera images at each frame, forming a system similar to RGB-D camera and hence can leverage existing visual SLAM works such as ORB-SLAM2 \cite{mur2017orb}. This is also a loosely-coupled method as it ignores the direct constraints imposed by LiDAR measurements. 

For works mentioned above, the measurement of LiDAR and camera are fused in a loosely-coupled manner. To achieve higher accuracy and robustness, frameworks that fuse sensor data in a tightly-coupled way are proposed in recent years. Zuo \textit{et al.} \cite{zuo2019lic} propose a LIC-fusion framework combining IMU measurements, sparse visual features, and LiDAR plane and edge features with online spatial and temporal calibration based on the MSCKF framework. The system exhibits higher accuracy and robustness than other state-of-the-art methods in their experiment results. Later on, their further work termed LIC-Fusion 2.0 \cite{zuo2020lic} refines a novel plane-feature tracking algorithm across multiple LiDAR scans within a sliding window to make LiDAR scan matching more robust. Shan \textit{et al.} in \cite{lvisam} propose LVI-SAM that fuses the LiDAR, visual and inertial sensors in a tightly-coupled smooth and mapping framework, which is built atop a factor graph. The LiDAR-inertial and visual-inertial subsystems of LVI-SAM can function independently when failure is detected in one of them, or jointly when enough features are detected. A similar tightly-coupled system is our previous work R$^2$LIVE \cite{r2live}, which fuses the LiDAR and camera measurements in an on-manifold iterated Kalman filter. R$^2$LIVE can run in various challenging scenarios even with small LiDAR FoV, aggressive motions, sensor failures, and narrow tunnel-like environments with moving objects. 

The above LiDAR-inertial-visual systems all use feature-based methods in both LIO and VIO. In contrast, R$^3$LIVE++ uses direct methods in both LIO and VIO to best exploit any subtle features in the environments even in case of extreme scenarios (e.g., structure-less and/or texture-less environments).  Moreover, the above LiDAR-inertial-visual systems mainly focus on the localization part and has very limited consideration on the mapping efficiency and accuracy. Hence, their visual and LiDAR subsystem often maintains two separate map for the LIO and VIO, preventing the data fusion at a deeper level and the reconstruction of high-accuracy colored 3D maps. R$^3$LIVE++ is designed to perform both localization and radiance map reconstruction in real-time. The central of these two tasks is a single radiance map shared among and maintained by both LIO and VIO. In particular, the LIO subsystem reconstructs the geometric structure of the map and the VIO subsystem recovers the radiance information of the map.

This paper is an extension of the previously published work R$^3$LIVE \cite{r3live}. The extended works of this paper include (1) a full incorporation of the camera photometric calibration, which corrects the camera nonlinear response function and lens vignetting effect; (2) online estimation of the camera exposure time. The estimated exposure time and the camera photometric calibration enables the system to recover the true radiance information of the environment; (3) a more comprehensive evaluation of the system on both open and private dataset in terms of localization accuracy, robustness and radiance map reconstruction accuracy; and (4) release of the system codes, private dataset, and the in-house designed hardware devices for collecting this dataset. 

\section{Basic models}
\subsection{Notations}
In this paper, we use notations shown in Table~\ref{table_I_NOMENCLATURE}. 

\vspace{-0.1cm}
\begin{table}[htbp]
	\caption{NOMENCLATURE}
	\vspace{-0.0cm}
	\setlength\tabcolsep{2.5pt}
	\begin{tabular}{r l p{10cm} }
		\toprule
		\textbf{Notation} & \textbf{\hspace{2cm} Explanation} \\
		\toprule
		&\hspace{2cm}{\textit{Expressions}}\\
		\toprule
		$\boxplus / \boxminus$ &  The encapsulated ``boxplus'' and \\  & \quad ``boxminus'' operations on manifold \cite{he2021embedding}  \\		
		${^G}(\cdot)$ &  				The value of $(\cdot)$ expressed in global frame \\		
		${^C}(\cdot)$ &  				The value of $(\cdot)$ expressed in camera frame \\		
		$\mathtt{Exp}(\cdot) / \mathtt{Log}(\cdot)$ & The Rodrigues' transformation between the\\ & \quad rotation matrix and rotation vector \\
		$\delta\left(\cdot\right)$ & The estimated error of $(\cdot)$ parameterized \\& \quad   in  tangent space.\\
		$\boldsymbol{\Sigma}_{(\cdot)}$ & The covariance matrix of vector $(\cdot)$ \\
		\toprule
		&\hspace{2cm}{\textit{Variables}} \\
		\toprule
		$\mathbf{b}_{\mathbf{g}}, \mathbf{b}_{\mathbf{a}}$ &  The bias of gyroscope and accelerometer in an IMU \\
		${^G}\mathbf{g}$ & The gravitational acceleration in global frame \\
		${^G}\mathbf{v}$ & The linear velocity in global frame \\
		$({^G}\mathbf{R}_{I}, {^G}\mathbf{p}_{I})$ & The IMU attitude and position w.r.t. global frame \\
		$({^I}\mathbf{R}_{C}, {^I}\mathbf{p}_{C})$ & The extrinsic between camera and IMU \\
		$\mathbf{x}$ &  The ground-true state \\
		$\hat{\mathbf{x}}$ & The prior estimation of $\mathbf{x}$ \\
		$\check{\mathbf{x}}$ & The current estimate of $\mathbf{x}$ in each ESIKF iteration  \\  
		\bottomrule \\
		\label{table_I_NOMENCLATURE}
	\end{tabular}
	\vspace{-1.0cm}
\end{table}

\subsection{System overview}
To simultaneously estimate the sensor pose and reconstruct the environment radiance map, we design a tightly-coupled LiDAR-inertial-visual sensor fusion framework, as shown in Fig. \ref{fig_overview}. The proposed framework contains two subsystems: the LIO subsystem (upper part) and the VIO subsystem (lower part). The LIO subsystem constructs the geometric structure of the radiance map by registering point cloud measurements of each input LiDAR scan. The VIO subsystem recovers the radiance information of the map in two steps: the frame-to-frame VIO update estimates the system state by minimizing the frame-to-frame PnP reprojection error, while the frame-to-map VIO update minimizes the radiance error between map points and the current image. The two subsystems are tightly coupled within an on-manifold error-state iterated Kalman filter framework (ESIKF) \cite{he2021embedding}, where the LiDAR and camera visual measurements are fused to the same system state (Section \ref{state_vector}) at their respective data reception time (Section \ref{sect_LIO_subsystem} and Section \ref{sect_VIO_subsystem}).

\subsubsection{Point}\label{sect_raidance_map_point}
Our radiance map is composed of map points in the global frame, each point $\mathbf{P}$ is a structure as below:
\begin{align}
	\mathbf{P} = \left[ {^G}\mathbf{p}_x, {^G}\mathbf{p}_y, {^G}\mathbf{p}_z, \boldsymbol{\gamma}_r, \boldsymbol{\gamma}_g, \boldsymbol{\gamma}_b \right]^T = \left[{^G}\mathbf{p}^T, \boldsymbol{\gamma}^T \right]^T \in \mathbb{R}^6
\end{align}
where the head sub-vector ${^G}\mathbf{p} = \left[ {^G}\mathbf{p}_x, {^G}\mathbf{p}_y, {^G}\mathbf{p}_z\right]^T \in \mathbb{R}^3$ denotes the point 3D position, and the tail sub-vector $ \boldsymbol{\gamma} = \left[  \boldsymbol{\gamma}_r,  \boldsymbol{\gamma}_g,  \boldsymbol{\gamma}_b\right]^T \in \mathbb{R}^3 $ is the point radiance consisting of three independent channels (i.e., red, green, and blue channel) accounting for the camera photometric calibration (see Section~\ref{sect_camera_photography}). Besides, we also record other necessary information of this point, such as the $3\times 3$ matrix $\boldsymbol{\Sigma}_{\mathbf{p}}$ and $\boldsymbol{\Sigma}_{\boldsymbol{\gamma}}$, which denote the covariance of the estimation errors of ${^G}\mathbf{p}$ and $\boldsymbol{\gamma}$, respectively, and the timestamps when this point was created and updated.

\subsubsection{Voxel}

To inquiry a point in the radiance map efficiently (e.g., for camera pose tracking in Section \ref{Frame_to_map} and map point radiance recovery in Section \ref{sect_render_texture_of_maps}), we put map points in fix-size (e.g. \SI{0.1}{\meter\times}\SI{0.1}{\meter\times}\SI{0.1}{\meter}) voxels. If a voxel has points appended recently (e.g. in recent 1 second), we mark this voxel as \textit{activated}. Otherwise, this voxel is marked as \textit{deactivated}.


\subsection{Color camera photometric model}\label{sect_camera_photography}

\begin{figure}[t]
	\centering
	\includegraphics[width=1.00\linewidth]{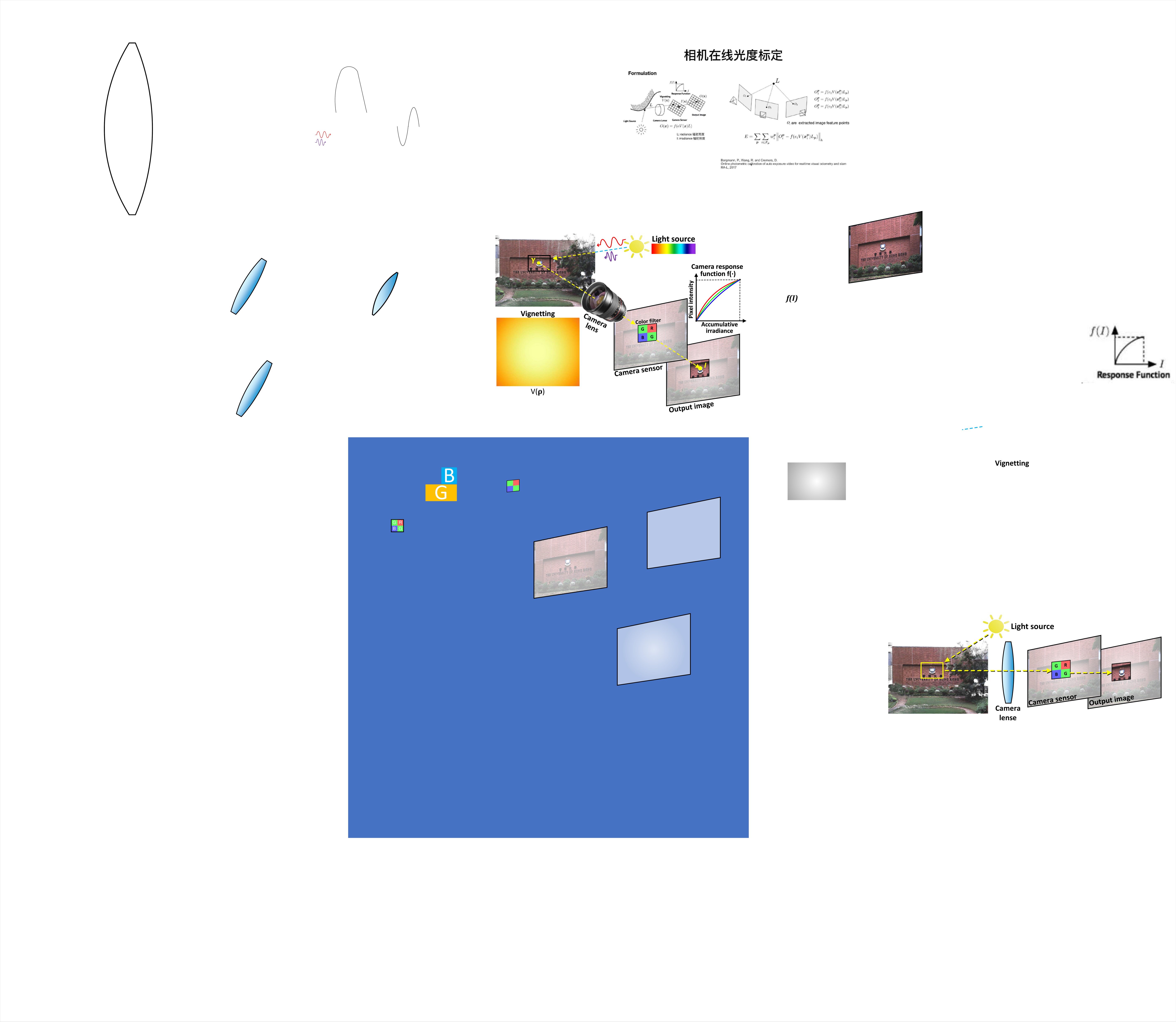}
	\caption{The image formation process of a color camera.}
	\label{fig_camera_photography}
	\vspace{-0.5cm}
\end{figure}

A camera observes the radiance of the real world in the form of images that consists of 2D arrays of pixel intensities. In our work, we model the image formation process of a camera based on \cite{bergmann2017online} and further extend the gray camera model to a color camera. As shown in Fig.~\ref{fig_camera_photography}, for a point $\mathbf P$ in the world, it reflects the incoming lights emitted from a light source (e.g., the sun). The reflected lights then pass through the camera lens and finally arrive at the CMOS sensor, which records the intensity of the reflected lights and creates a pixel channel in the output image. The recorded intensity is determined by the {\it radiance}, a measure of the power reflected at the point $\mathbf P$. 

To model the above imaging process, we denote $\boldsymbol{\gamma}$ the radiance at point $\mathbf P$. Since a color camera has three channels in its CMOS sensor: red, green, and blue, the radiance $\boldsymbol{\gamma}$ has three components: $\boldsymbol{\gamma}_r,  \boldsymbol{\gamma}_g,  \boldsymbol{\gamma}_b$, respectively. For each channel $i$, the lights passing through the camera lens has power
\begin{align}
	\mathbf{O}_i(\boldsymbol{\rho}) = {V}(\boldsymbol{\rho}) \boldsymbol{\gamma}_i
\end{align}
where ${V}(\boldsymbol{\rho})\in 
[0,1]$ is called the vignetting factor accounting for the lens vignetting effect. Since the vignetting effect is different at different area of the lens, the vignetting factor ${V}(\boldsymbol{\rho})$ is a function of the pixel location $\boldsymbol{\rho}$.  

$\mathbf{O}_i(\boldsymbol{\rho})$ is the amount of power that can be received by the sensor and is called the irradiance. When taking an image, the captured irradiance $\mathbf{O}(\boldsymbol{\rho})$ is integrated over time (i.e., the exposure time $\tau$). The accumulated irradiance $\boldsymbol{\theta}_i = \tau  {V}(\boldsymbol{\rho}) \boldsymbol{\gamma}_i  $ is then converted as the output of pixel intensity $\mathbf{I}_i(\boldsymbol{\rho})$ via the camera response function (CRF) $\mathbf{f}_i(\cdot)$:
\begin{align}
	\mathbf{I}_i(\boldsymbol{\rho}) = \mathbf{f}_i(\tau  {V}(\boldsymbol{\rho}) \boldsymbol{\gamma}_i ), ~~
	\mathbf{I}_i \in \left[0, 255\right]. \label{eq_image_output}
\end{align}
Since a real camera sensor has a limited dynamic range and that the physical scale of the radiance $\boldsymbol{\gamma}$ can not be recovered anyway, the pixel intensities can be normalized within $[0, 1]$ without loss of generality. 

As noted in (\ref{eq_image_output}), different channels often has different non-linear response function (CRF) $\mathbf{f}_i(\cdot)$ and they can be calibrated offline along with the vignetting factor ${V}(\boldsymbol{\rho})$ based on the method in \cite{engel2016photometrically}. The exposure time $\tau$ is estimated online in our work.  With the calibration and estimation results, the radiance of point $\mathbf{P}$ from the observed pixel value $\mathbf{I}(\boldsymbol{\rho})$ can be computed as:
\begin{align}
	\boldsymbol{\gamma}_i =  \dfrac{\mathbf{f}_i^{-1}(\mathbf{I}_i(\boldsymbol{\rho}))}{\tau {V} (\boldsymbol{\rho})} 
\end{align}

{\it Remark:} Under the assumption of constant continuous light sources and a Lambertian reflection model, the radiance at point $\mathbf P$ is a constant physical value that is invariant to the camera pose. Such invariance to time and camera pose enables us to infer the camera ego-motion from the radiance difference between the map and the current image (with photometric calibration), as detailed in Section \ref{Frame_to_map}.

\subsection{State}\label{state_vector}
In our work, we define the full state $\mathbf{x}$ as: 
\begin{align}
	\begin{split}
		\hspace{-0.0cm}\mathbf{x} &= 
		\left(
	^G\mathbf{R}_{I}, {^G}\mathbf{p}_{I},  {^G}\mathbf{v}, \mathbf{b}_{\mathbf{g}},\mathbf{b}_{\mathbf{a}},{^G}\mathbf{g}, {^I}\mathbf{R}_{C}, {^I}\mathbf{p}_{C}, \epsilon, {^I}t_{C} , \boldsymbol{\phi}
		\right)
	\end{split}
	\label{eq_full_state}
\end{align}
where the notations ${}^G\mathbf{R}_{I}$, ${^G}\mathbf{p}_{I}$,  ${^G}\mathbf{v}$, $\mathbf{b}_{\mathbf{g}}$, $\mathbf{b}_{\mathbf{a}}$, ${^G}\mathbf{g}$, ${^I}\mathbf{R}_{C}$, ${^I}\mathbf{p}_{C}$ are explained in Table~\ref{table_I_NOMENCLATURE}, ${^I}t_{C}$ is the time-offset between IMU and camera while LiDAR is assumed to be synced with the IMU already, $\epsilon = 1/\tau$ is the inverse camera exposure time, $\boldsymbol{\phi} = \begin{bmatrix}
	f_x, f_y, c_x, c_y		
\end{bmatrix}^T$ are the camera intrinsics, where $(f_x, f_y)$ denote the camera focal length and $(c_x, c_y)$ the offsets of the principal point from the top-left corner of the image plane. The camera extrinsic $({^I}\mathbf{R}_{C}, {^I}\mathbf{p}_{C})$, intrinsic $\boldsymbol{\phi}$ and time-offset ${^I}t_{C}$ would usually have their rough values available (e.g., offline calibration, CAD model, manufacturer's manual). To cope with the possible calibration errors (e.g., extrinsic $({^I}\mathbf{R}_{C}, {^I}\mathbf{p}_{C})$ and intrinsic $\boldsymbol{\phi}$) or online drifting (e.g., time-offset ${^I}t_{C}$), we also include them in the state $\mathbf{x}$ such that they will be estimated online. Beside, we also estimate the camera exposure time online in order to recover the true radiance value of each map point. 


\section{LiDAR-Inertial odometry (LIO)}\label{sect_LIO_subsystem}

Our LIO subsystem reconstructs the geometry structure of the environment by registering each new LiDAR scan to the global map. We use the generalized-iterative closet point (GICP) method \cite{segal2009generalized} to iteratively estimate the LiDAR pose (and other system state) by minimizes the distance of each point in the scan to a plane fitted from the corresponding points in the map. The estimated state estimate is then used to append the new points to the map. 

\subsection{LiDAR point-to-plane residual}
As shown in Fig. \ref{fig_overview}, our LIO subsystem constructs the geometric structure of the global map. 
For the $k$-th input LiDAR scan, we first compensate the in-frame motion with a IMU backward propagation introduced in \cite{xu2020fast}. Let  ${\boldsymbol{\mathcal{L}}}_{k} = \left\{^{L}\mathbf{p}_{1}, ...,  ^{L}\mathbf{p}_{m}\right\}$ be the set of $m$ LiDAR points after motion compensation, we compute the residual of each raw point (or a downsampled subset) of ${^{L}\mathbf{p}_s} \in {\boldsymbol{\mathcal{L}}}_{k}$ where $s$ is the index of point and the superscript $L$ denotes that the point is represented in the LiDAR-reference frame.

With $\check{\mathbf x}_{k}$ being the estimate of ${\mathbf x}_{k}$ at the current iteration, we transform ${^{L}\mathbf{p}_s}$ from LiDAR frame to the global frame:
\begin{align}
	{^{G}\mathbf{p}_s} = {^G{\check{\mathbf{R}}}_{I_{k}}}( ^I\mathbf{R}_{L}{^{L}{\mathbf{p}}_{s}}  + {^{I}\mathbf{p}_L}) +  {^G\check{\mathbf{p}}_{I_{k}}} \label{eq_p_LiDAR_frame_to_global_frame}
\end{align}
To register the point to the global map, we search for the nearest five points in the map. To accelerate the nearest neighbor search, map points are organized into an incremental k-d tree (see \cite{fastlio2}). The found nearest neighbor points are used to fit a plane with normal $\mathbf{u}_{s}$ and centroid ${{\mathbf{q}}_{s}}$, then the LiDAR measurement residual $\mathbf{r}_l(\check{\mathbf{x}}_{k}, {^{L}{\mathbf{p}}}_{s})$ is:
\begin{align}
	\mathbf{r}_l(\check{\mathbf{x}}_{k}, {^{L}{\mathbf{p}}}_{s})=& \mathbf{u}_{s}^T\left({^{G}\mathbf{p}_{s}} - {{\mathbf{q}}_{s} } \right) \label{eq_def_rl_x_kplus1}      
\end{align}
	
	\subsection{LIO ESIKF update}\label{sec_LIO_update}
	
	The residual in (\ref{eq_def_rl_x_kplus1}) should be zero ideally. However, due to the estimation error in $\check{\mathbf{x}}_{k}$ and the LiDAR measurement noise, this residual is often not zero and can be used to refine the state estimate $\check{\mathbf{x}}_{k}$. Specifically, let $\mathbf n_{s}$ be the measurement noise of the point ${^{L}\mathbf{p}_{s}}$, we have the relation between the true point location ${^{L}\mathbf{p}^{\mathtt{gt}}_{s}}$ and the measured one ${^{L}\mathbf{p}_{s}}$ as below:
\begin{align}
	{^{L}\mathbf{p}_{s}} = {^{L}\mathbf{p}^{\mathtt{gt}}_{s}} + {\mathbf{n}_{s}},  {\mathbf{n}_{s}} \sim \mathcal{N}(\mathbf{0}, \boldsymbol{\Sigma}_{\mathbf{n}_{s}}). 
\end{align}
This true point location together with the true state $\mathbf x_{k}$ should lead to zero residual in (\ref{eq_def_rl_x_kplus1}), i.e.,
\begin{align}
	\mathbf 0 = \mathbf{r}_l(\mathbf{x}_{k}, {^{L}\mathbf{p}}^{\mathtt{gt}}_{s}) &\approx \mathbf{r}_l(\check{\mathbf{x}}_{k}, {^{L}{\mathbf{p}}}_{s}) + \mathbf{H}^{{l}}_{s} \delta \check{\mathbf{x}}_{k} + \boldsymbol{\alpha}_{s}, \label{eq_lidar_meas}
\end{align}
where $\mathbf x_{k}$ is parameterized by its error $\delta \check{\mathbf{x}}_{k}$ in the tangent space of $\check{\mathbf{x}}_{k}$ (i.e., $\mathbf x_{k} = \check{\mathbf{x}}_{k} \boxplus \delta \check{\mathbf{x}}_{k}$), $\boldsymbol{\alpha}_{s} \sim \mathcal{N}(\mathbf{0}, \boldsymbol{\Sigma}_{\boldsymbol{\alpha}_{s}})$ is the lumped noise due to $\mathbf n_s$, and $\mathbf{H}^{{l}}_{s}$ is the Jacobian of the residual w.r.t. $\delta \check{\mathbf{x}}_{k}$. 

	Equation (\ref{eq_lidar_meas}) constitutes an observation distribution for ${\mathbf{x}}_{k}$ (or equivalently $\delta \check{\mathbf{x}}_{k} \triangleq {\mathbf{x}}_{k} \boxminus \check{\mathbf{x}}_{k}$), which is combined with the prior distribution from the IMU propagation:
	\begin{align}
		\begin{split}
				\hspace{-0.5cm}\mathop{\min}_{\delta \check{\mathbf{x}}_{k}} &\left(  \left\| \left( \check{\mathbf{x}}_{k} \boxplus \delta \check{\mathbf{x}}_{k} \right) \boxminus \hat{\mathbf{x}}_{k}  \right\|_{\boldsymbol{\Sigma}_{\delta \hat{\mathbf{x}}_{k}} }^2 \right.  \\
				+&\left. \sum\nolimits_{s=1}^{m}\left\|  {  \mathbf{r}_l(\check{\mathbf{x}}_{k}, {^{L}{\mathbf{p}}}_{s}) + \mathbf{H}^{l}_{s} \delta \check{\mathbf{x}}_{k} } \right\|^2_{{\boldsymbol{\Sigma}_{\boldsymbol{\alpha}_{s}}}}  \right) 
			\label{eq_optimial_LiDAR_MAP}
		\end{split}
	\end{align}
	where $\left\| \mathbf{x} \right\|_{\boldsymbol{\Sigma}}^2 = \mathbf{x}^T \boldsymbol{\Sigma}^{-1} \mathbf{x}  $ is the squared Mahalanobis distance with covariance $\boldsymbol{\Sigma}$,  $\hat{\mathbf{x}}_{k}$ is the IMU propagated state estimate, and $\boldsymbol{\Sigma}_{\delta \hat{\mathbf{x}}_{k}}$ is the IMU propagated state covariance. 
	{The detailed derivation of first item in (\ref{eq_optimial_LiDAR_MAP}) can be found in Section~  IV-E of R$^2$LIVE\cite{r2live}}.
	
	Solving (\ref{eq_optimial_LiDAR_MAP}) leads to the Maximum A-Posteriori (MAP) estimate of $\delta \check{\mathbf{x}}_{k}^o$ which is then added to $\check{\mathbf{x}}_{k}$ as below
	\begin{align}
		\check{\mathbf{x}}_{k} \leftarrow \check{\mathbf{x}}_{k}\boxplus \delta \check{\mathbf{x}}_{k}^o \label{eq_x_k_update_frame_to_frame}
	\end{align}

	\indent The above iteration process is iterated until convergence (i.e., the update $\delta \check{\mathbf{x}}_{k}^o$ is smaller than a given threshold).  The converged state estimate $\check{\mathbf{x}}_{k}$ is then used as the starting point of the IMU propagation until the reception of the next LiDAR scan or camera image. Furthermore, the converged estimate $\check{\mathbf{x}}_{k}$ is used to append points in the current LiDAR scan to the global map as follows. For the $s$-th point ${^{L}\mathbf{p}_s} \in {\boldsymbol{\mathcal{L}}}_{k}$, its position in global frame ${^{G}\mathbf{p}_s}$ is first obtained by (\ref{eq_p_LiDAR_frame_to_global_frame}).  If ${^{G}\mathbf{p}_s}$ has nearby points in the map with distance \SI{1}{\centi\meter} (see Section~\ref{sect_raidance_map_point}), ${^{G}\mathbf{p}_s}$ will be discarded to maintain a spatial resolution of \SI{1}{\centi\meter}. Otherwise, a new point structure $\mathbf{P}_{s}$ will be created in the map with:
	\begin{align}
		\mathbf{P}_{s} = \left[  {^G}\mathbf{p}_{s}^T, \boldsymbol{\gamma}_{s}^T \right]^T = \left[  ^{G}\mathbf{p}_s, \mathbf{0} \right]^T
	\end{align}
	where the radiance vector $\boldsymbol{\gamma}_{s}$ is set as zero and will be initialized at the first time it is observed in forthcoming images (see Section~\ref{sect_render_texture_of_maps}). Finally, we mark the voxel containing ${^{G}\mathbf{p}_s}$ as \textit{activated} such that the radiance of points in this voxel can be updated by the forthcoming images (see Section~\ref{sect_render_texture_of_maps}).
	
	\section{Visual-Inertial odometry (VIO)}\label{sect_VIO_subsystem}
	While our LIO subsystem reconstructs the geometric structure of the environment, our VIO subsystem recovers the radiance information from the input color images. To be more specific, our VIO subsystem projects a certain number of points (i.e., tracked points) from the global map to the current image, then it iteratively estimates the camera pose (and other system state) by minimizing the radiance error of these points. Only a sparse set of tracked map points is used for the sake of computation efficiency. 
	
	Our proposed framework is different from previous photometric-based methods \cite{forster2014svo, wang2017stereo}, which constitute the residual of a point by considering the photometric error over all its neighborhood pixels (i.e., a patch). These patch-based methods achieve stronger robustness and faster convergence speed than that without. However, the patch-based method is not invariant to either translation or rotation, which requires estimating the relative transform when aligning one patch to another. Plus, the calculation of the residual is not completely precise by assuming the depths of all pixels in the patch are the same as the mid-point. On the other hand, our VIO is operated at an individual pixel, which utilizes the radiance of a single map point to compute the residual. The radiance, which is updated simultaneously in the VIO, is an inherent property of a point in the world and is invariant to both camera translation and rotation. To ensure a robust and fast convergence, we design a two-step pipeline shown in Fig. \ref{fig_overview}, where in the first step (i.e., frame-to-frame VIO) we leverage a frame-to-frame optical flow to track map points observed in the last frame and obtain a rough estimate of the system's state by minimizing the Perspective-n-Point (PnP) reprojection error of the tracked points (Section \ref{section_frame_to_frame_vio}). Then, in the second step (i.e., frame-to-map VIO), the state estimate is further refined by minimizing the difference between the radiance of map points and the pixel intensities at their projected location in the current image (Section \ref{Frame_to_map}). With the converged state estimate and the raw input image, we finally update map points radiance according to the current image measurement (Section \ref{sect_render_texture_of_maps}).
	
	\subsection{Photometric correction}
	For each incoming image $\mathbf{I}$, we first correct the image non-linear CRF $\mathbf{f}_i(\cdot)$ and the vignetting factor $V(\cdot)$, which are calibrated in advance (see Section~\ref{sect_camera_photography}), to obtain the photometrically corrected image $\boldsymbol{\Gamma}$, whose $i$-th channel at pixel location $\boldsymbol{\rho}$ is: 
	\begin{align}\label{eq:photometric_cali}
		\boldsymbol{\Gamma}_{i} (\boldsymbol{\rho})  =  \dfrac{\mathbf{f}_i^{-1}(\mathbf{I}_i(\boldsymbol{\rho}))}{{V}(\boldsymbol{\rho})}.
	\end{align}
	The photometrically corrected image $\boldsymbol{\Gamma}$ is then used in the following VIO pipelines including the frame-to-frame VIO, frame-to-map VIO and radiance recovery.

	\subsection{Frame-to-frame Visual-Inertial odometry}\label{section_frame_to_frame_vio}
	\subsubsection{Perspective-n-Point reprojection error}
	\begin{figure}[t]
		\centering
		\includegraphics[width=1.0\linewidth]{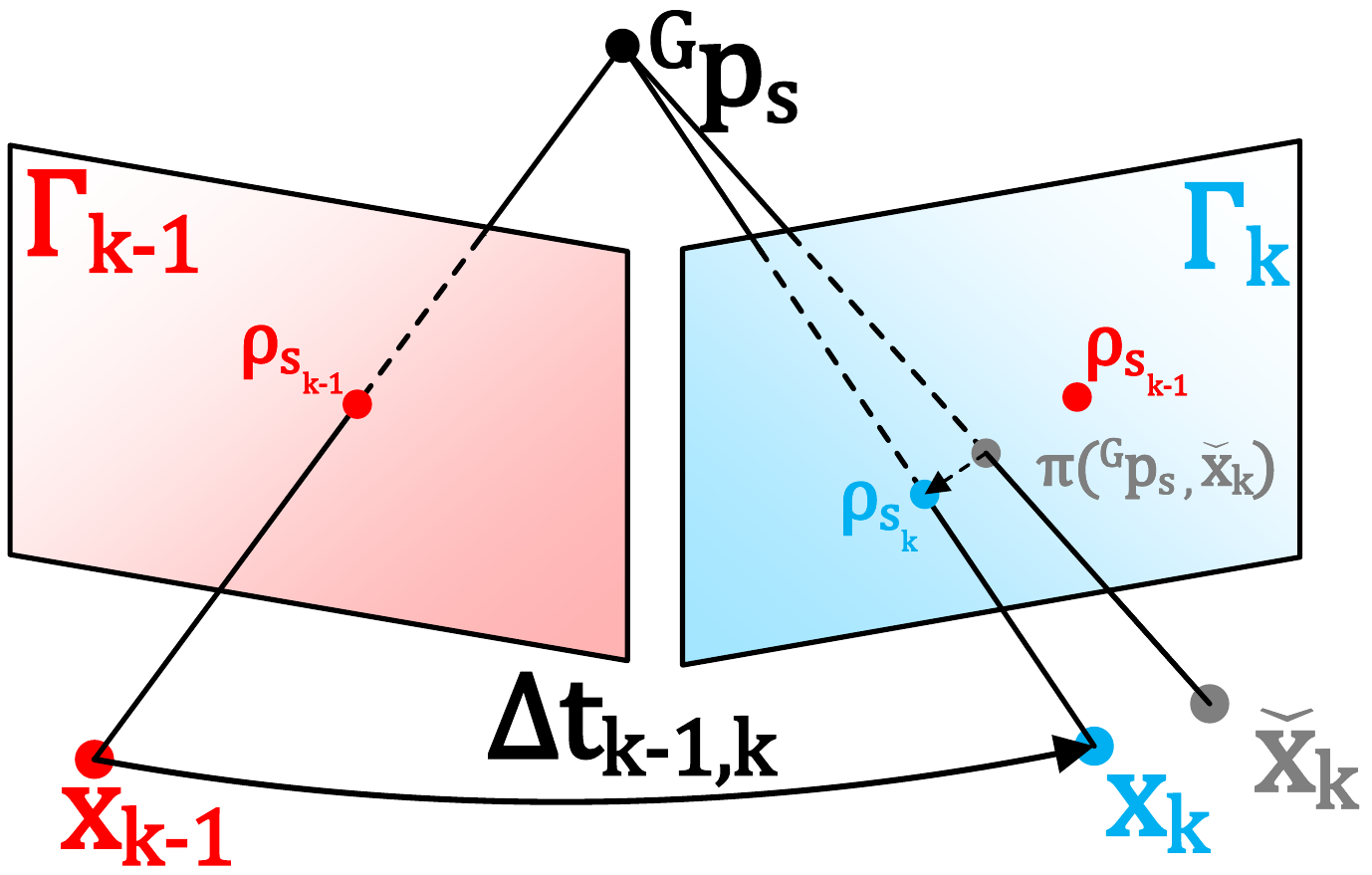}
		\caption{Frame-to-frame VIO estimates the system's state by minimizing the PnP reprojection error of map points observed in the last frame.}
		\label{fig_frame_to_frame}
		\vspace{-0.5cm}
	\end{figure}
	
	Assume we have tracked $m$ map points $\boldsymbol{\mathcal{P}} = \{ \mathbf{P}_1, ..., \mathbf{P}_m \}$ in the last image frame $\mathbf{I}_{k-1}$ with their projected location in $\mathbf{I}_{k-1}$ being $\{ \boldsymbol{\rho}_{1_{k-1}},...,  \boldsymbol{\rho}_{m_{k-1}} \}$, we leverage the Lucas$-$Kanade optical flow to find out their corresponding location in the current image $\mathbf{I}_k$, denoted as $\{ \boldsymbol{\rho}_{1_k},...,  \boldsymbol{\rho}_{m_k}\}$. Then, we iteratively minimize the reprojection errors of  $\boldsymbol{\mathcal{P}}$ to obtain a rough estimate of the state (see Fig. \ref{fig_frame_to_frame}). Specifically, taking the $s$-th point $\mathbf{P}_s =  \left[ {^G}\mathbf{p}_s^T, \boldsymbol{\gamma}_s^T \right]^T  \in \boldsymbol{\mathcal{P}}$ as example, let $\check{\mathbf{x}}_{k}$ be the state estimate at the current iteration, the projection error $\mathbf{r}_c \left(\check{\mathbf{x}}_{k}, \boldsymbol{\rho}_{s_k} , {^G\mathbf{p}_{s}}\right)$ is
	\vspace{-0.2cm}

\vspace{-0.15cm}
\begin{align}
		\mathbf{r}_c \left(\check{\mathbf{x}}_{k}, \boldsymbol{\rho}_{s_k} , {^G\mathbf{p}_{s}}\right) = 
		\boldsymbol{\rho}_{s_k} - \boldsymbol{\pi}({^G\mathbf{p}_{s}}, \check{\mathbf{x}}_k ) \label{eq_frame_to_map_projection_err} 
	\end{align}
    where $ \boldsymbol{\pi}({^G\mathbf{p}_{s}}, \check{\mathbf{x}}_k ) \in \mathbb{R}^2 $ is the predicted pixel location computed as below:

	\begin{align}
		\begin{split}
			\boldsymbol{\pi}({^G\mathbf{p}_{s}}, \check{\mathbf{x}}_k )  &= \boldsymbol{\pi}_{\text{ph}}({^G\mathbf{p}_{s}}, \check{\mathbf{x}}_k )	+ \dfrac{{^I}\check{t}_{C_k}}{\Delta t_{k-1,k}}( \boldsymbol{\rho}_{s_k} - \boldsymbol{\rho}_{s_{k-1}} ) \label{eq_projection}
		\end{split}
	\end{align}
	where the first term $\boldsymbol{\pi}_{\text{ph}}({^G\mathbf{p}_{s}}, \check{\mathbf{x}}_k )$ is the standard camera pin-hole model, the second one is the temporal correction factor \cite{qin2018online}, and $\Delta t_{k-1,k}$ is the time interval between the last and current image.

	\subsubsection{Frame-to-frame VIO Update}\label{sec:update}
	
	Similar to the LIO update, the state estimation error in $\check{\mathbf{x}}_{k}$ and the camera measurement noise will lead to a certain residual in (\ref{eq_frame_to_map_projection_err}), from which we can update the state estimate $\check{\mathbf{x}}_{k}$ as follows. First, the measurement noise in the residual (\ref{eq_frame_to_map_projection_err}) consists of two sources: one is the pixel tracking error in $\boldsymbol{\rho}_{s_k}$ and the other lies in the map point location error ${^G\mathbf{p}_{s}}$, 
	\begin{align}
		{^G\mathbf{p}_{s}}	= {^G\mathbf{p}_{s}^{\mathtt{gt}}}  + \mathbf{n}_{\mathbf{p}_{s}},~& \mathbf{n}_{\mathbf{p}_{s}} \sim \mathcal{N}(\mathbf{0}, \boldsymbol{\Sigma}_{\mathbf{n}_{\mathbf{p}_{s}}})  \\
		\boldsymbol{\rho}_{s_k} = {\boldsymbol{\rho}_{s_k}^{\mathtt{gt}}} + \mathbf{n}_{\boldsymbol{\rho}_{s_k}},~& \mathbf{n}_{\boldsymbol{\rho}_{s_k}} \sim \mathcal{N}(\mathbf{0}, \boldsymbol{\Sigma}_{\mathbf{n}_{\boldsymbol{\rho}_{s_k}}}) 
	\end{align}
	where $ {^G\mathbf{p}_{s}^{\mathtt{gt}}}$ and ${\boldsymbol{\rho}_{s_k}^{\mathtt{gt}}}$ are the true values of ${^G\mathbf{p}_{s}}$ and $\boldsymbol{\rho}_{s_k}$, respectively. Then, correcting such noises and using the true system state should lead to zero residual, i.e.,
	\begin{equation}
		\begin{split}
			\hspace{-0.3cm}	\mathbf 0 &= \mathbf{r}_c (\mathbf{x}_{k}, \boldsymbol{\rho}_{s_k}^{\mathtt{gt}}, {^G\mathbf{p}_{s}^{\mathtt{gt}}}) \approx \mathbf{r}_c\left(\check{\mathbf{x}}_{k}, \boldsymbol{\rho}_{s_k} , {^G\mathbf{p}_{s}}\right) + \mathbf{H}^{r}_{s} \delta \check{\mathbf{x}}_{k} + \boldsymbol{\beta}_{s} \label{eq_visual_meas}
		\end{split}
	\end{equation}
	where $\mathbf{H}^{r}_{s}$ is the Jacobian of the residual w.r.t. $\delta \check{\mathbf{x}}_{k}$ and $\boldsymbol{\beta}_{s} \sim \mathcal{N}(\mathbf{0}, {\boldsymbol{\Sigma}_{\boldsymbol{\beta}_{s}}}  )$ is the lumped noise due to $\mathbf n_{\mathbf p_s}$ and $\mathbf n_{\boldsymbol{\rho}_s}$.
	
	Equation (\ref{eq_visual_meas}) constitutes an observation distribution for $\mathbf x_k$, which is combined with the IMU propagation to obtain the MAP estimate of the state in the same way as the LIO update detailed in Section \ref{sec_LIO_update}. The converged state estimate is then refined in the frame-to-map VIO in the next section. 
	
	{\it Remark:} Since the camera pin-hole model  $\boldsymbol{\pi}_{\text{ph}}({^G\mathbf{p}_{s}}, {\mathbf{x}}_k )$ in (\ref{eq_projection}) is related to camera pose (consisting of the IMU pose $({^G}\mathbf{R}_{I}, {^G}\mathbf{p}_{I})$ and camera extrinsic $({^I}\mathbf{R}_{C}, {^I}\mathbf{p}_{C})$) and intrinsic $\boldsymbol{\phi}$, so the projection model  $\boldsymbol{\pi}({^G\mathbf{p}_{s}}, {\mathbf{x}}_k )$ is also related to these state components. In addition,  $\boldsymbol{\pi}({^G\mathbf{p}_{s}}, {\mathbf{x}}_k )$ is also related to the temporal offset ${^I}t_{C}$ due to the temporal correction factor.  This will cause $\mathbf{H}^{r}_{s}$ to contain nonzero elements corresponding to the IMU pose $({^G}\mathbf{R}_{I}, {^G}\mathbf{p}_{I})$, camera extrinsic $({^I}\mathbf{R}_{C}, {^I}\mathbf{p}_{C})$, intrinsic $\boldsymbol{\phi}$, and temporal offset ${^I}t_{C}$, and hence an update of them in the state estimation.

	\begin{figure}[t]
		\centering
		\includegraphics[width=1.0\linewidth]{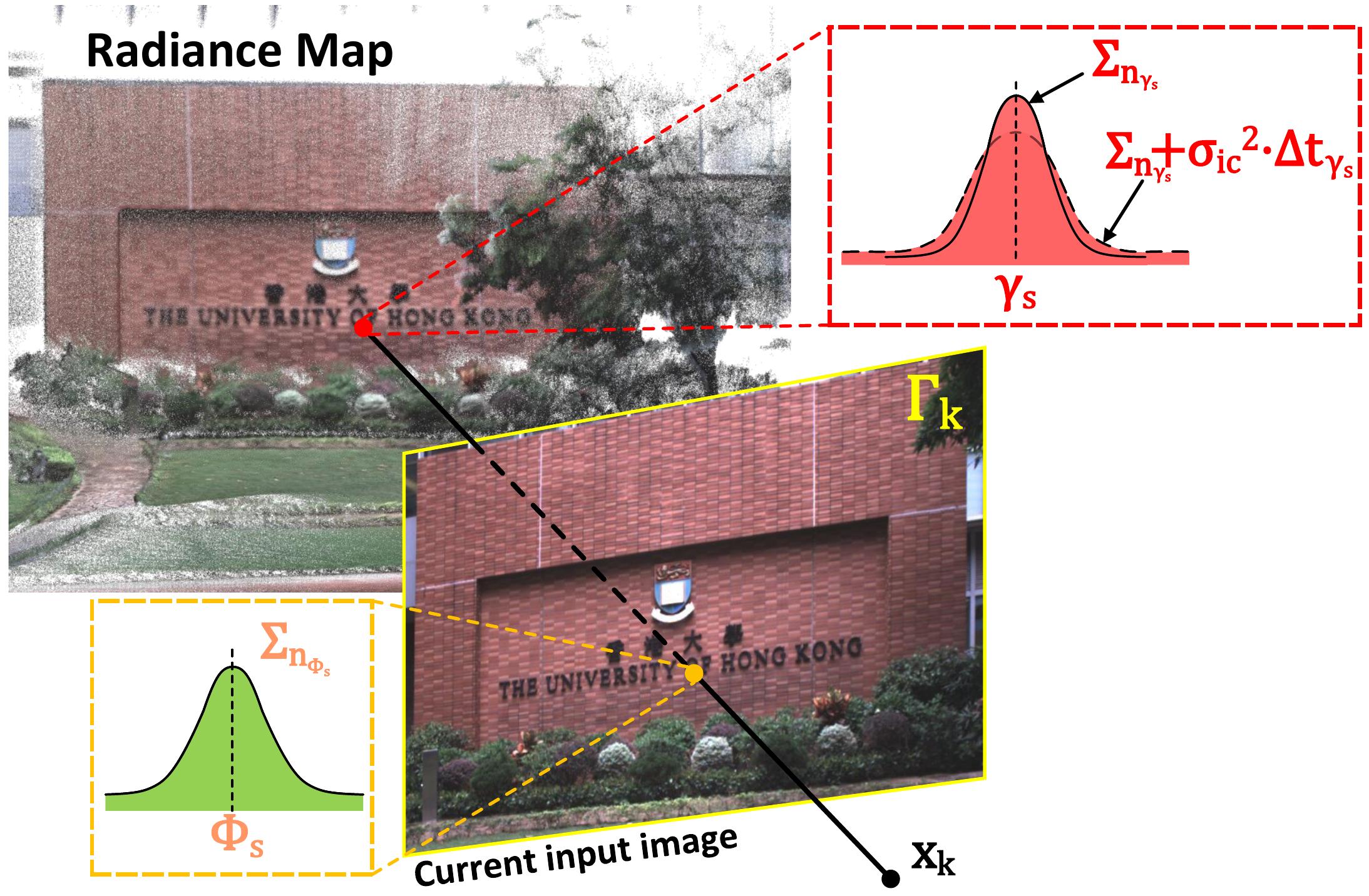}
		\caption{Frame-to-map VIO refines the state estimate by minimizing the radiance error between the map point and the observed radiance in the current image.}
		\label{fig_frame_to_frame_PE}
		\vspace{-0.5cm}
	\end{figure}
	
	\subsection{Frame-to-map Visual-Inertial odometry}\label{Frame_to_map}
	\subsubsection{Frame-to-map radiance error}
	\label{sect_frame_to_map_PE}
	The frame-to-frame VIO update can provide a good state estimate $\check{\mathbf{x}}_k$, which is further refined by the frame-to-map VIO update by minimizing the radiance error of the tracked map points $\boldsymbol{\mathcal{P}}$. Let $\boldsymbol{\Gamma}_k$ the photometrically calibrated image at the $k$-th step (see (\ref{eq:photometric_cali})). With the state estimate at the current iteration,  $\check{\mathbf{x}}_{k}$, which contains the estimated camera pose, extrinsic, intrinsic, and exposure time, we project a tracked map point $\mathbf{P}_s \in \boldsymbol{\mathcal{P}}$ to the image plane to obtain its pixel location $\check{\boldsymbol{\rho}}_{s_k} = \boldsymbol{\pi}({^G\mathbf{p}_{s}}, \check{\mathbf{x}}_k )$ (see (\ref{eq_projection}) and Fig. \ref{fig_frame_to_frame_PE}).  Then, the observed radiance denoted by $\boldsymbol{\Phi}_s$ can be computed from the exposure time component $\check{\epsilon}_{k}$ in $\check{\mathbf x}_{k}$ as: $\boldsymbol{\Phi}_s = \check{\epsilon}_{k} \boldsymbol{\Gamma}_k (\check{\boldsymbol{\rho}}_{s_k}) $. Finally, the frame-to-map radiance error is the difference between the radiance component $\boldsymbol{\gamma}_s$ of the point $\mathbf{P}_s$ and the observed value $\boldsymbol{\Phi}_s$:
	\vspace{-0.15cm}
	\begin{align}
		\hspace{-0.3cm}{\mathbf{r}_c}( \check{\mathbf{x}}_k, {^G}\mathbf{p}_s, \boldsymbol{\gamma}_s  ) &=  \boldsymbol{\Phi}_s  - {\boldsymbol{\gamma}}_s, \quad \boldsymbol{\Phi}_s = \check{\epsilon}_{k} \boldsymbol{\Gamma}_k (\check{\boldsymbol{\rho}}_{s_k}) ,  \label{eq_frame_to_map_photo_err}
	\end{align}
	where $\boldsymbol{\Phi}_s, \boldsymbol{\Gamma}_k (\check{\boldsymbol{\rho}}_{s_k})$ and $\boldsymbol{\gamma}_s$ both contain three channels: red, green, and blue. 
	
	\subsubsection{Frame-to-map VIO update}\label{sect_frame_to_map_ESIKF_update}
	
	The measurement noise in (\ref{eq_frame_to_map_photo_err}) come from both the component $\boldsymbol{\gamma}_s$ and $\boldsymbol{\Phi}_s$. For the component $\boldsymbol{\gamma}_s$, we model it as:
	\begin{align}
		\begin{split}
			\boldsymbol{\gamma}_s = \boldsymbol{\gamma}_s^{gt} + \mathbf{n}_{\boldsymbol{\gamma}_s} +  \mathbf n_{\text{ic}} \label{eq_measurement_noise_C_s} , ~& \mathbf{n}_{\boldsymbol{\gamma}_s} \sim {\mathcal{N}}(\mathbf{0}, \boldsymbol{\Sigma}_{\mathbf{n}_{\boldsymbol{\gamma}_s}} ) \\  & 
			\hspace{-1.2cm} \mathbf n_{\text{ic}} \sim  {\mathcal{N}}(\mathbf{0},  \boldsymbol{\sigma}^2_{\text{ic}} \cdot \Delta t_{\boldsymbol{\gamma}_s} )
		\end{split}
	\end{align}
	where $\boldsymbol{\gamma}^{gt}_s$ is the ground truth of $\boldsymbol{\gamma}_s$, the first noise $\mathbf{n}_{\boldsymbol{\gamma}_s}$ is due to the radiance estimation error (Section \ref{sect_render_texture_of_maps}), and the second noise $\mathbf n_{\text{ic}}$ accounts for the radiance temporal change caused by illumination change. Since the illumination often changes slowly over time, we model it as a random walk, hence its covariance is linear to $ \Delta t_{\boldsymbol{\gamma}_s}$, the time interval between current time and last update time of $\mathbf{P}_s$. 
	
	For the second component $\boldsymbol{\Phi}_s$ in (\ref{eq_frame_to_map_photo_err}), it is computed from the state estimate $\check{\mathbf x}_k$ and the current image $\boldsymbol{\Gamma}_k$ as $\boldsymbol{\Phi}_s = \check{\epsilon}_{k} \boldsymbol{\Gamma}_k \left(\boldsymbol{\pi}({^G\mathbf{p}_{s}}, \check{\mathbf{x}}_k ) \right)$, hence its noise consists of two sources: one is the state estimation error (from $\check{\mathbf{x}}_k$) and the other is the image measurement noise (from $\boldsymbol{\Gamma}_k$):
	\vspace{-0.15cm}
	\begin{align}
		\boldsymbol{\Phi}_s  = \boldsymbol{\Phi}^{gt}_s + \mathbf{n}_{{\boldsymbol{\Phi}_s}}, ~&  \mathbf{n}_{{\boldsymbol{\Phi}_s}} \sim \mathcal{N}( \mathbf{0}, \boldsymbol{\Sigma}_{\mathbf{n}_{{\boldsymbol{\Phi}_s}}} ) \label{eq_measurement_noise_Phi_s} 
	\end{align} 
	\vspace{-0.15cm}
	\hspace{-0.21cm}
	where $\boldsymbol{\Sigma}_{\mathbf{n}_{{\boldsymbol{\Phi}_s}}}$ denotes the covariance due to these two noise sources.   
	
	Combining (\ref{eq_frame_to_map_photo_err}), (\ref{eq_measurement_noise_C_s}) and (\ref{eq_measurement_noise_Phi_s}), we obtain the first order Taylor expansion of the true zero residual $\mathbf r_c( \mathbf{x}_k, {^G}\mathbf{p}^{gt}_s,  \boldsymbol{\gamma}^{gt}_s )$:
	\begin{align}
		\begin{split}
			\hspace{-0.2cm}\mathbf{0} &= \mathbf r_c ( \mathbf{x}_k, {^G}\mathbf{p}^{gt}_s, \boldsymbol{\gamma}^{gt}_s ) \approx \mathbf r_c ( \check{\mathbf{x}}_k, {^G}\mathbf{p}_s, \mathbf{c}_s ) + \mathbf{H}_{s}^{c} \delta \check{\mathbf{x}}_k + \boldsymbol{\zeta}_{s} \label{eq_visual_pe_residual}
		\end{split}
	\end{align} 
	{where $\mathbf{H}^{r}_{s}$ is the Jacobian of the residual w.r.t. $\delta \check{\mathbf{x}}_{k}$} and $\boldsymbol{\zeta}_{s} \sim \mathcal{N}(\mathbf{0}, \boldsymbol{\Sigma}_{\zeta_{s}})$ is the lumped noise due to noises in $\boldsymbol{\gamma}_s$ and $\boldsymbol{\Phi}_s$.
	
	Similar as before, (\ref{eq_visual_pe_residual}) constitutes an observation distribution for state $\mathbf x_k$, which is combined with the IMU propgation to obtain the MAP estimate of the state. 
	
	{\it Remark:} Since the $\boldsymbol{\Phi}_s$ in (\ref{eq_frame_to_map_photo_err}) is related to camera exposure time $\epsilon$, it will cause $\mathbf{H}^{c}_{s}$ to contain nonzero elements corresponding to the exposure time and hence an update of them in the state estimation.
	
	\subsection{Recovery of radiance information}\label{sect_render_texture_of_maps} 
	
	\begin{figure}[t]
		\centering
		\includegraphics[width=1.0\linewidth]{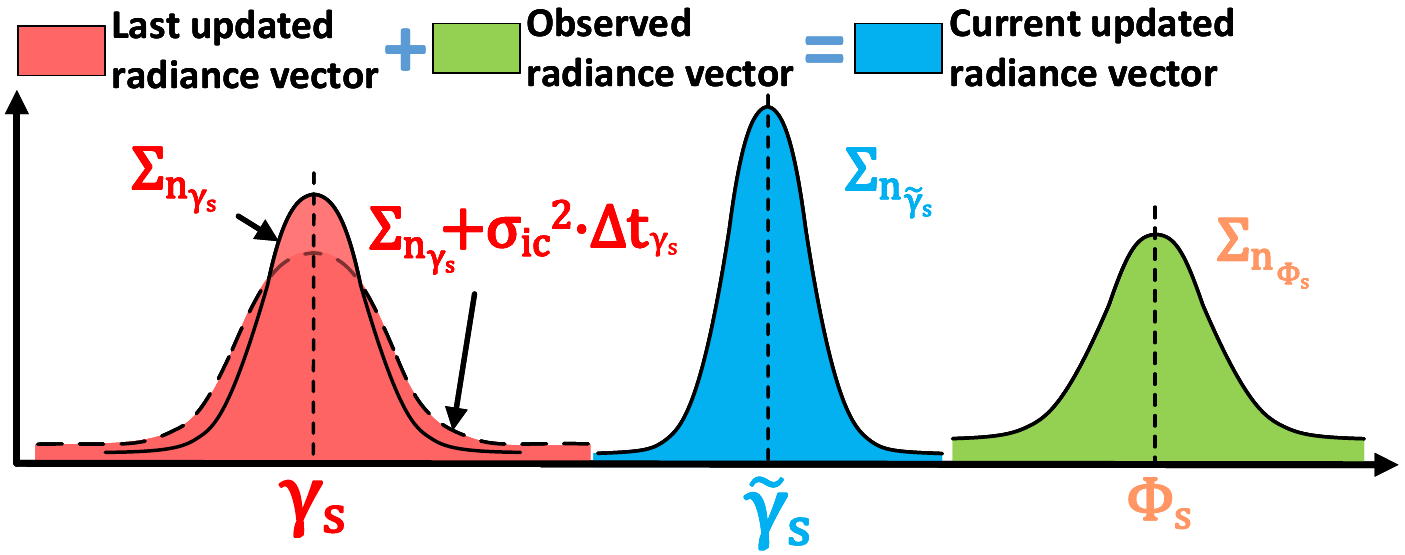}
		\caption{We update the radiance $\boldsymbol{\gamma}_s$ of a map point via Bayesian update.}
		\label{fig_color_update}
		\vspace{-0.5cm}
	\end{figure}
	
	After the frame-to-map VIO update, we have the precise pose of the current image. Then, we perform the Bayesian update to determine the optimal radiance of all map points such that the average radiance error between each point and its viewed images is minimal.

	First of all, we retrieve all the points in all \textit{activated} voxels (activated in Section \ref{sec_LIO_update}). Assume the retrieved point set is $\boldsymbol{\mathcal{Q}} = \{ \mathbf{P}_1, ..., \mathbf{P}_n \}$. For the $s$-th point $\mathbf{P}_s = \left[ {^G}\mathbf{p}_s^T, \boldsymbol{\gamma}_s^T \right]^T \in \boldsymbol{\mathcal{Q}}$ falling in the current image FoV, we first can obtain the observed radiance vector ${\boldsymbol{\Phi}}_s$ by (\ref{eq_frame_to_map_photo_err}) and its covariance $\boldsymbol{\Sigma}_{\mathbf{n}_{{{\boldsymbol{\Phi}}_s}}}$ by (\ref{eq_measurement_noise_Phi_s}). Then, if $\mathbf{P}_s$ is a new point appended by the LIO subsystem (see Section~\ref{sec_LIO_update}) with $\boldsymbol{\gamma}_{s} = \mathbf{0}$, we set:
	\begin{align}
		\boldsymbol{\gamma}_{s} = \boldsymbol{\Phi}_s,~~ {\boldsymbol{\Sigma}}_{\mathbf{n}_{\boldsymbol{\gamma}_s}} = \boldsymbol{\Sigma}_{\mathbf{n}_{{{\boldsymbol{\Phi}}_s}}}
	\end{align}
	Otherwise, the radiance vector $\boldsymbol{\gamma}_s$ saved in the map (see (\ref{eq_measurement_noise_C_s})) is fused with newly observed radiance vector $\boldsymbol{\Phi}_s$ with covariance $\boldsymbol{\Sigma}_{\mathbf{n}_{{{\boldsymbol{\Phi}}_s}}}$ via Bayesian update (see Fig. \ref{fig_color_update}):
	\begin{align}
		\hspace{-0.5cm}\boldsymbol{\Sigma}_{\mathbf{n}_{\tilde{\boldsymbol{\gamma}}_s}} = \left( \left(\boldsymbol{\Sigma}_{\mathbf{n}_{\boldsymbol{\gamma}_s}} + \boldsymbol{\sigma}^2_{\text{ic}} \cdot \Delta t_{\boldsymbol{\gamma}_s} \right)^{-1} + \boldsymbol{\Sigma}^{-1}_{\mathbf{n}_{{{{\boldsymbol{\Phi}}}_s}}} \right)^{-1} \\
		\hspace{-0.5cm}\tilde{\boldsymbol{\gamma}}_{s} = \left( \left(\boldsymbol{\Sigma}_{\mathbf{n}_{\boldsymbol{\gamma}_s}} + \boldsymbol{\sigma}^2_{\text{ic}} \cdot \Delta t_{\boldsymbol{\gamma}_s} \right)^{-1} \boldsymbol{\gamma}_s + \boldsymbol{\Sigma}^{-1}_{\mathbf{n}_{{{{\boldsymbol{\Phi}}}_s}}} {{\boldsymbol{\Phi}}}_s \right)^{-1} \boldsymbol{\Sigma}_{\mathbf{n}_{\tilde{\boldsymbol{\gamma}}_s}} \\
		\hspace{1.5cm}{\boldsymbol{\gamma}}_{s} = \tilde{\boldsymbol{\gamma}}_{s}, ~~ {\boldsymbol{\Sigma}}_{\mathbf{n}_{\boldsymbol{\gamma}_s}} = \boldsymbol{\Sigma}_{\mathbf{n}_{\tilde{\boldsymbol{\gamma}}_s}} \hspace{1.5cm}
	\end{align} 
	
	\subsection{Update of the tracking points}\label{sect_update_tracking_pts}
	After the {recovery of radiance information}, we update the tracked point set $\boldsymbol{\mathcal{P}}$ for the next frame of image use. Firstly, we remove points from current $\boldsymbol{\mathcal{P}}$ if their  projection error in (\ref{eq_frame_to_map_projection_err}) or radiance error in (\ref{eq_frame_to_map_photo_err}) are too large, and also remove the points which does not fall into the current image FoV. Secondly, we project each point in $\boldsymbol{\mathcal{Q}}$ to the current image and add it to $\boldsymbol{\mathcal{P}}$ if no other tracked points already existed in a neighborhood of $50$ pixels.
 
\section{Experiments}

In this chapter, we conduct extensive experiments to validate the advantages of our proposed system against other counterparts in threefold: 1) To verify the accuracy in localization, we quantitatively compare our system against existing state-of-the-art SLAM systems on a public dataset (NCLT-dataset). 2) To validate the robustness of our framework, we test it under various challenging scenarios where camera and LiDAR sensor degeneration occurs. 3) To evaluate the accuracy of our system in reconstructing the radiance map, we compare it against existing baselines in estimating the camera exposure time and calculating the average photometric error w.r.t. each image. In the experiments, two datasets are used for evaluation: the NCLT-dataset and the R$^3$LIVE-dataset.

\subsection{NCLT-dataset}
To compare the accuracy of our proposed method against other state-of-the-art SLAM systems, we perform quantitative evaluations on NCLT dataset\cite{carlevaris2016university}. NCLT-dataset is a large-scale, long-term autonomy dataset for robotics research that was collected on the University of Michigan’s North Campus. The dataset is comprised of 27 sequences that are collected by exploring the campus, both indoors and outdoors, on varying paths, and at different time of a day across all four seasons. Each sequence includes data from the omnidirectional camera, 3D lidar, planar lidar, GPS, and wheel encoders on a Segway robot.

We choose NCLT-dataset for three reasons: 1) NCLT-dataset is currently the largest public dataset with ground-truth trajectories of high quality. 2) NCLT-dataset provides all raw data sampled by the sensors, which meets our requirement for the input data. 3) NCLT-dataset has many challenging scenarios, such as moving obstacles (e.g., pedestrians, bicyclists, and cars), illumination changing, varying viewpoint, seasonal and weather changes (e.g., falling leaves and snow), and long-term structural changes caused by construction projects. 

In the experiments, the front-facing camera data (one of five) and the 3D LiDAR data are used for all systems under evaluation.  Moreover, we notice some time synchronization errors in two sequences (i.e., 2012-03-17 and 2012-08-04), where the LiDAR timestamp is 100ms delayed from the IMU timestamp (about one LiDAR-frame). Therefore, we exclude these two sequences from the evaluation. As a result, 25 sequences are evaluated with total traveling length up to \SI{138}{\kilo\meter} and duration up to \SI{33}{\hour}:\SI{34}{\minute}.

\subsection{Our private dataset: R$^3$LIVE-dataset}
While the large-scale NCLT-dataset is suitable for evaluating the localization accuracy, it didn't cover any scenarios with sensor degeneration, preventing us from evaluating the system robustness, which is one of the major motivations of this work. Moreover, the camera photometric calibration and the ground-true exposure time are not available in the NCLT-dataset, which are essential for the reconstruction of the radiance maps and the evaluation of the online exposure time estimation. To fill this gap, we designed a handheld data collection device and made a new dataset named R$^3$LIVE-dataset. The dataset and hardware device are released along with the codes of this work to facilitate the reproduction of our work.
 
\subsubsection{Handheld device for data collection}\label{sect_equiment_set}
\begin{figure}[h]
	\centering
	\includegraphics[width=1.0\linewidth]{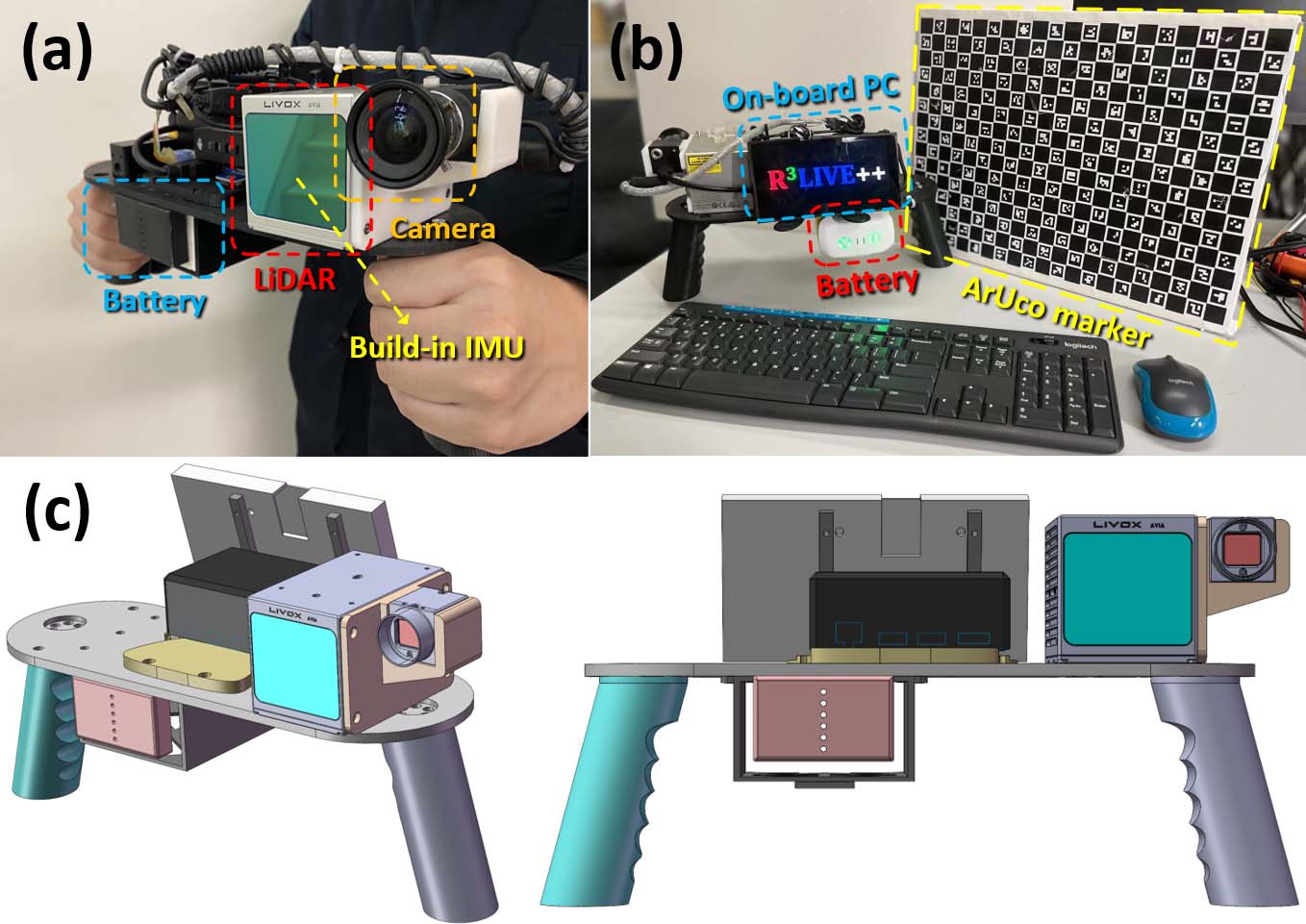}
	\caption{(a) shows our handheld device for data collection. (b) shows the ArUco marker board to provide the ground-truth for evaluating the system accuracy. (c) shows our open source schematics model.}
	\label{fig_handheld_device}
\end{figure}
Our handheld device for data collection is shown in Fig. \ref{fig_handheld_device}(a), which includes a power supply unit, an onboard computer \textit{DJI manifold-2c} (equipped with  an \textit{Intel i7-8550u} CPU and 8 GB RAM), a \textit{FLIR Blackfly BFS-u3-13y3c} global shutter camera,  and a \textit{LiVOX AVIA} 3D LiDAR. The camera FoV is  \SI{82.9}{\degree\times}\SI{66.5}{\degree} and the LiDAR FoV is \SI{70.4}{\degree\times}\SI{77.2}{\degree}. To quantitatively evaluate the accuracy of our algorithm (Section \ref{sect_experiment_2}) even in GPS denied environments, we use an ArUco marker \cite{garrido2014automatic} as a reference to calculate the sensor pose when returns to the starting point, which enables to evaluate the localization drift. All of the mechanical modules of this device are designed as FDM printable, schematics of the design are opened on our Github: \href{https://github.com/hku-mars/r3live}{\tt github.com/hku-mars/r3live}.

To correct the camera's nonlinear response function and the vignette effect, we perform photometric calibration on the camera based on the method in \cite{engel2016photometrically}. The calibrated results are shown in Fig.\ref{fig_photometric_calibration}, which are also released on Github.
\begin{figure}[h]
	\centering
	\includegraphics[width=0.45\linewidth]{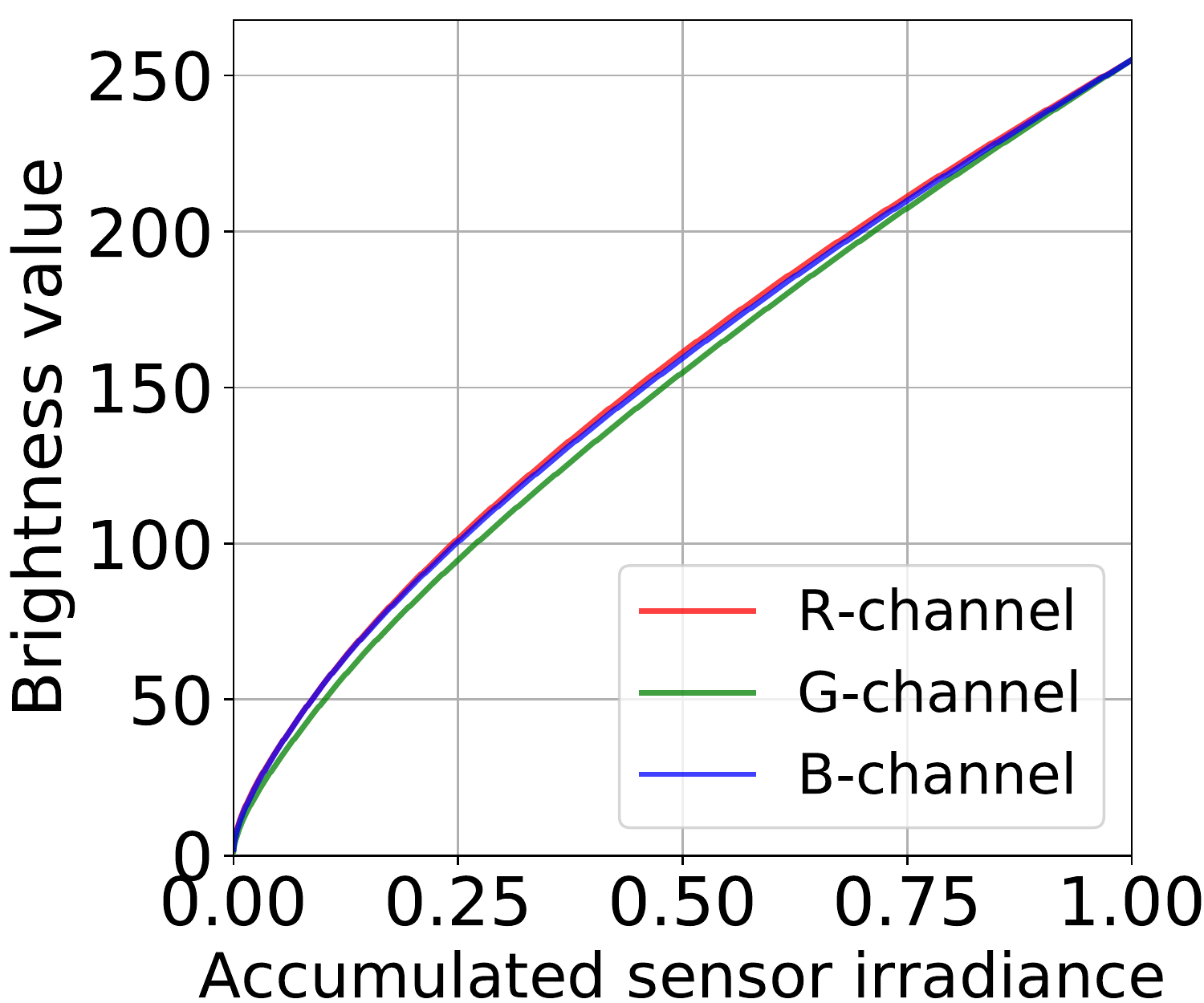}
	\includegraphics[width=0.53\linewidth]{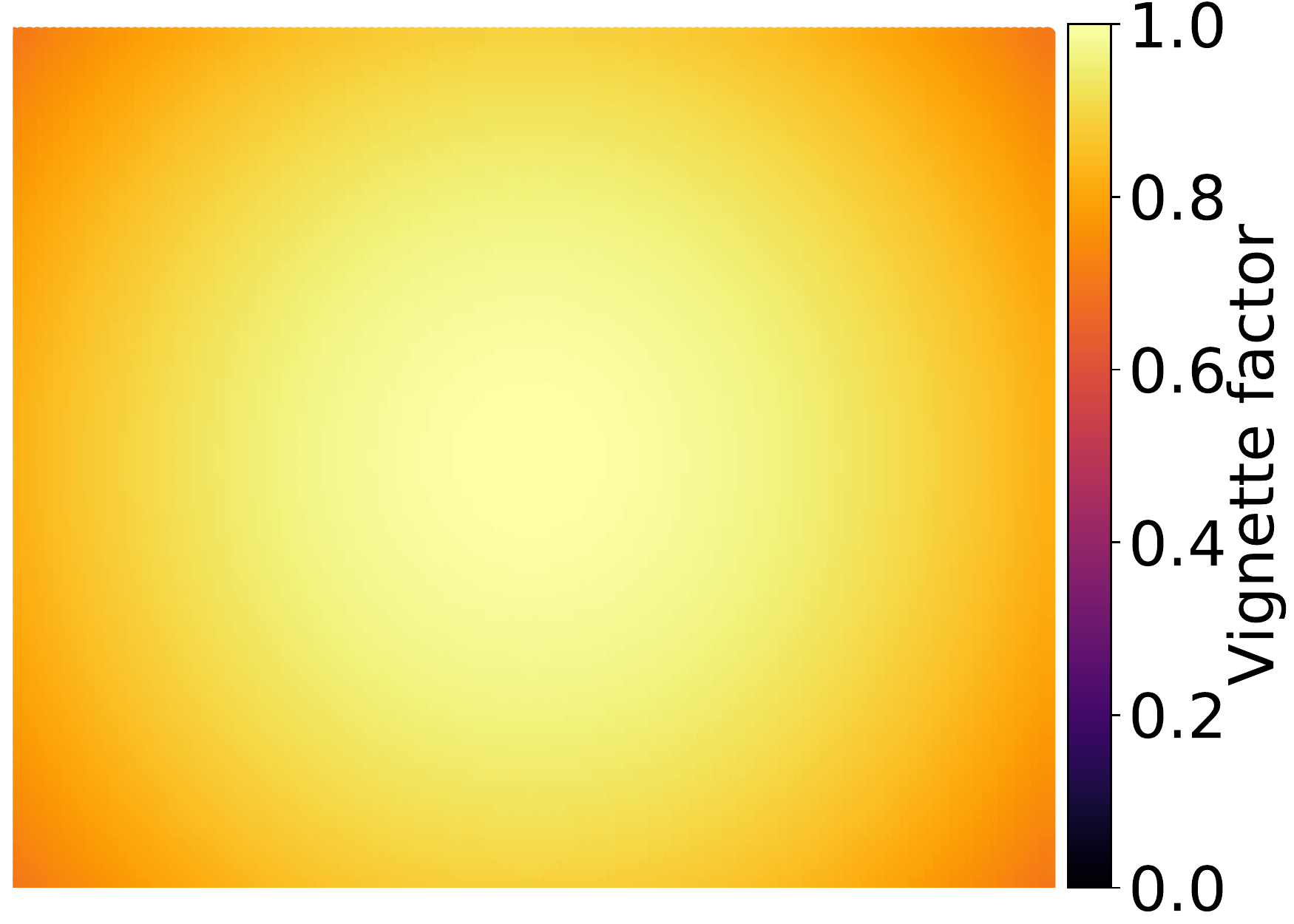}
	\caption{The left figure shows the calibrated nonlinear response function in three channels (red, green, and blue). The right one plots the calibrated vignetting factors at each image pixel. }
	\vspace{-0.5cm}
	\label{fig_photometric_calibration}
\end{figure}

\subsubsection{The R$^3$LIVE-dataset}\label{sect_r3live_dataset}

The R$^3$LIVE-dataset was collected within the campuses of the University of Hong Kong (HKU) and the Hong Kong University of Science and Technology (HKUST). As summarized in Table~\ref{TAB_R3LIVE_DATASET}, the dataset includes 13 sequences that are collected by exploring both indoor and outdoor environments, in various scenes (e.g., walkway, park, forest, etc) at different time in a day (i.e., morning, noon, and evening). This allows the dataset to capture both structured urban buildings and cluttered field environments with different lighting conditions. The dataset also includes three sequences (degenerate\_seq\_00/01/02) where the LiDAR or camera (or both) degenerate by occasionally facing the device to a single and/or texture-less plane (e.g., wall, the ground) or visually. The total traveling length reaches \SI{8.4}{\kilo\meter}, duration reaching \SI{2.4}{\hour}. More details of each sequence will be provided in sequel when it is used. 
\begin{table*}
	\definecolor{LightBlue}{rgb}{0.7,0.7,1}
	\newcolumntype{b}{>{\columncolor{LightBlue}}c}
	\centering
	\begin{threeparttable}[t]	
		\scriptsize
		\setlength\tabcolsep{8pt}
		\renewcommand{\arraystretch}{1.1} 
		\centering
		\caption{Overview of the R$^3$LIVE-dataset.}
			\begin{tabular}{cccccccc}
				\hline
				\textbf{Sequence}    & \textbf{\begin{tabular}[c]{@{}c@{}}Duration\\ (s)\end{tabular}} & \textbf{\begin{tabular}[c]{@{}c@{}}Traveling\\  Length (m)\end{tabular}} & \textbf{\begin{tabular}[c]{@{}c@{}}Sensor \\ Degeneration\end{tabular}} & \textbf{\begin{tabular}[c]{@{}c@{}}Return to \\ origin$^{\textbf{1}}$\end{tabular}} & \textbf{Aruco marker}$^{\textbf{2}}$ & \textbf{\begin{tabular}[c]{@{}c@{}}Camera exposure \\ time$^{\textbf{3}}$\end{tabular}} & \textbf{Scenarios} \\ \hline
				degenerate\_seq\_00  & 101                                                             & 74.9                                                                     & Camera, LiDAR                                                           & \checkmark                                                                       &                       &                                                                          & Indoor             \\
				degenerate\_seq\_01  & 86                                                              & 53.3                                                                     & LiDAR                                                                   & \checkmark                                                                       &                       &                                                                          & Outdoor            \\
				degenerate\_seq\_02  & 85                                                              & 75.2                                                                     & LiDAR                                                                   & \checkmark                                                                       & \checkmark                   &                                                                          & Outdoor            \\
				hku\_campus\_seq\_00 & 202                                                             & 190.6                                                                    & ----                                                                    & \checkmark                                                                       &                       &                                                                          & Indoor             \\
				hku\_campus\_seq\_01 & 304                                                             & 374.6                                                                    & ----                                                                    &                                                                           &                       &                                                                          & Outdoor            \\
				hku\_campus\_seq\_02 & 323                                                             & 354.3                                                                    & ----                                                                    & \checkmark                                                                       &                       & \checkmark                                                                      & Indoor, Outdoor    \\
				hku\_campus\_seq\_03 & 173                                                             & 181.2                                                                    & ----                                                                    & \checkmark                                                                       &                       & \checkmark                                                                      & Indoor, Outdoor    \\
				hku\_main\_building  & 1170                                                            & 1036.9                                                                   & ----                                                                    & \checkmark                                                                       &                       & \checkmark                                                                      & Indoor, Outdoor    \\
				hku\_park\_00        & 351                                                             & 401.8                                                                    & ----                                                                    & \checkmark                                                                       & \checkmark                   &                                                                          & Outdoor, Cluttered \\
				hku\_park\_01        & 228                                                             & 247.3                                                                    & ----                                                                    & \checkmark                                                                       & \checkmark                   &                                                                          & Outdoor, Cluttered \\
				hkust\_campus\_00    & 1073                                                            & 1317.2                                                                   & ----                                                                    & \checkmark                                                                       & \checkmark                   &                                                                          & Indoor, Outdoor    \\
				hkust\_campus\_01    & 1162                                                            & 1524.3                                                                   & ----                                                                    & \checkmark                                                                       & \checkmark                   &                                                                          & Indoor, Outdoor    \\
				hkust\_campus\_02    & 478                                                             & 503.8                                                                    & ----                                                                    & \checkmark                                                                       &                       & \checkmark                                                                      & Indoor, Outdoor    \\ 
				hkust\_campus\_03    & 1618                                                            & 2112.2                                                                   & ----                                                                    &                                                                           &                       & \checkmark                                                                      & Outdoor            \\
				\hline
				\textbf{Total}       & 7354                                                            & 8447.6                                                                   &                                                                         &                                                                           &                       &                                                                          &                   
				\\ \hline	
			\end{tabular}
			\label{TAB_R3LIVE_DATASET}
			\begin{tablenotes}
				\scriptsize
				\item[1] Sequences are collected by traveling a loop, with starting from and ending with the same position.
				\item[2] Sequences with ArUco marker for providing the ground-truth relative pose.
				\item[3] Sequences with ground-truth camera exposure time read from camera's API.
			\end{tablenotes}
		\end{threeparttable}
	\end{table*}

\subsection{System configurations}\label{sect_equiment_set}

For the sake of fair comparison, in the evaluation of our systems and their counterparts, each system uses the same parameters for all sequences in the same dataset. For the counterpart systems (e.g., LIO-SAM, LVI-SAM, FAST-LIO2, etc), we use their default configurations on their Github repository except for some necessary adjustments to match the hardware setup. For our system, we also make its configuration available on Github, ``\href{https://github.com/hku-mars/r3live/blob/master/config/r3live\_config.yaml}{\tt{r3live\_config\_nclt.yaml}}" for NCLT-dataset and ``\href{https://github.com/hku-mars/r3live/blob/master/config/r3live\_config.yaml}{\tt{r3live\_config.yaml}}" for R$^3$LIVE-dataset.


\begin{table*}
\begin{threeparttable}[t]
	\caption{The comparison of absolute position errors (APE, meters) on NCLT-dataset}
	\label{TABLE_APE_NCLT}
	\centering
	\normalsize
	\setlength\tabcolsep{7pt}
	\vspace{-0.25cm}
	\renewcommand{\arraystretch}{1.2} 
	\begin{tabular}{cccccccccc}
		\toprule
		\textbf{\begin{tabular}[c]{@{}c@{}}Sequence\\ (date)\end{tabular}} & \textbf{\begin{tabular}[c]{@{}c@{}}Length\\ (m)\end{tabular}} & \textbf{\begin{tabular}[c]{@{}c@{}}Duration\\ (hour:minute:second)\end{tabular}} & \textbf{Our}   & \textbf{R$^2$LIVE} & \textbf{LVI-SAM} & \textbf{Our-LIO} & \textbf{Fast-LIO2} & \textbf{LIO-SAM} \\ \hline 
		2012-01-08                                                         & 6495.69                                                       & 1 hr:25 min:35 sec                                                               & \textbf{10.81} & 22.43           & 23.43            & 20.07             & 18.50              & 21.66            \\ 
		2012-01-15                                                         & 7499.80                                                       & 1 hr:52 min:19 sec                                                               & 6.64           & 5.10            & ----             & 6.19              & \textbf{4.81}      & ----             \\ 
		2012-01-22                                                         & 6183.07                                                       & 1 hr:27 min:22 sec                                                               & 9.23           & 12.64           & 8.29             & 12.39             & \textbf{7.14}      & 8.99             \\ 
		2012-02-02                                                         & 6315.78                                                       & 1 hr:38 min:36 sec                                                               &\textbf{ 5.33}           & 6.05            & 18.08            & 7.44              & 9.12               & 15.63            \\ 
		2012-02-04                                                         & 5641.00                                                       & 1 hr:18 min:30 sec                                                               & \textbf{5.58}  & 8.36            & 9.63             & 7.78              & 7.19               & 10.78            \\ 
		2012-02-05                                                         & 6649.26                                                       & 1 hr:34 min:17 sec                                                               & 8.52           & \textbf{7.58}   & ----             & 7.67              & 7.80               & ----             \\ 
		2012-02-12                                                         & 5829.12                                                       & 1 hr:25 min:35 sec                                                               & \textbf{4.50}  & 6.47            & 40.01            & 10.48             & 8.30               & 45.02            \\ 
		2012-02-18                                                         & 6249.20                                                       & 1 hr:29 min:55 sec                                                               & \textbf{40.50} & 59.25           & ----             & 53.53             & 56.97              & ----             \\ 
		2012-02-19                                                         & 6232.68                                                       & 1 hr:29 min:11 sec                                                               & 8.52           & 6.58            & 8.87             & 6.19              & \textbf{5.98}      & 9.63             \\ 
		2012-03-17                                                         & 5907.19                                                       & 1 hr:22 min:53 sec                                                               & 4.83           & 5.94            & 12.58            & 6.77              & \textbf{4.70}      & 11.82            \\ 
		2012-03-31                                                         & 6073.71                                                       & 1 hr:27 min:53 sec                                                               & \textbf{4.94}  & 9.96            & 19.04            & 9.33              & 7.27               & 18.34            \\ 
		2012-04-29                                                         & 3183.09                                                       & 43 min:18 sec                                                                    & 6.32           & 6.43            & 5.94             & 6.27              & 6.44               & \textbf{5.67}    \\ 
		2012-05-11                                                         & 6116.74                                                       & 1 hr:25 min:5 sec                                                                & \textbf{3.70}  & 3.79            & 4.23             & 4.21              & 4.13               & 4.16             \\ 
		2012-05-26                                                         & 6340.70                                                       & 1 hr:28 min:34 sec                                                               & \textbf{4.55}  & 6.30            & 18.34            & 6.13              & 6.43               & 18.38            \\ 
		2012-06-15                                                         & 4085.89                                                       & 55 min:10 sec                                                                    & 7.74           & 6.29            & ----             & 5.65              & \textbf{5.27}      & ----             \\ 
		2012-08-04                                                         & 5492.13                                                       & 1 hr:20 min:32 sec                                                               & 3.85           & \textbf{3.73}   & 11.00            & 4.53              & 6.96               & 12.73            \\ 
		2012-08-20                                                         & 6014.51                                                       & 1 hr:23 min:48 sec                                                               & 4.50           & 4.46            & 11.20            & \textbf{4.18}     & 6.02               & 11.43            \\ 
		2012-09-28                                                         & 5574.41                                                       & 1 hr: 17 min:59 sec                                                              & 7.89           & \textbf{6.59}   & 34.42            & 6.84              & 10.42              & 36.71            \\ 
		2012-10-28                                                         & 5682.10                                                       & 1 hr:26 min:10 sec                                                               & 7.71           & 7.95            & ----             & 8.61              & \textbf{7.68}      & ----             \\ 
		2012-11-04                                                         & 4788.33                                                       & 1 hr:20 min:39 sec                                                               & 7.48           & 9.31            & 3.42             & 12.55             & \textbf{3.33}      & 3.37             \\ 
		2012-11-17                                                         & 5751.89                                                       & 1 hr:29 min:44 sec                                                               & 8.68           & 6.48            & 21.92            & 6.13              & \textbf{5.83}      & 24.17            \\ 
		2012-12-01                                                         & 4991.93                                                       & 1 hr:16 min:48 sec                                                               & 11.25          & 14.21           & 6.93             & 16.96             & \textbf{7.41}      & 7.21             \\ 
		2013-01-10                                                         & 1137.32                                                       & 17 min:4 sec                                                                     & \textbf{3.41}  & 4.57            & 4.88             & 5.30              & 3.50               & 5.08             \\ 
		2013-02-23                                                         & 5235.27                                                       & 1hr: 20min:08 sec                                                                & \textbf{11.64} & 13.39           & 12.60            & 13.79             & 11.85              & 12.20            \\ 
		2013-04-05                                                         & 4523.65                                                       & 1 hr:9 min:27 sec                                                                & 8.82           & 11.91           & 9.83             & 11.55             & \textbf{6.38}      & 9.01             \\ \hline
		\textbf{Total}:                                                             & 137994.46                                                     & 33 hr: 34 min: 52 sec                                                            &                &                 &                  &                   &                    &                  \\ 
		\textbf{Average}                                                            &                                                               &                                                                                  & \textbf{8.51}  & 10.58           & 15.03            & 10.75             & 9.59               & 15.39            \\ \toprule
	\end{tabular}
\begin{tablenotes}
	\normalsize
	\item[1] Some systems fail in midway in some sequences and are marked as "----".
\end{tablenotes}
\end{threeparttable}
\vspace{0.5cm}
\centering
\vspace{-0.4cm}
\hspace{-0.5cm}
\setcounter{figure}{9} 
\includegraphics[width=1.05\linewidth]{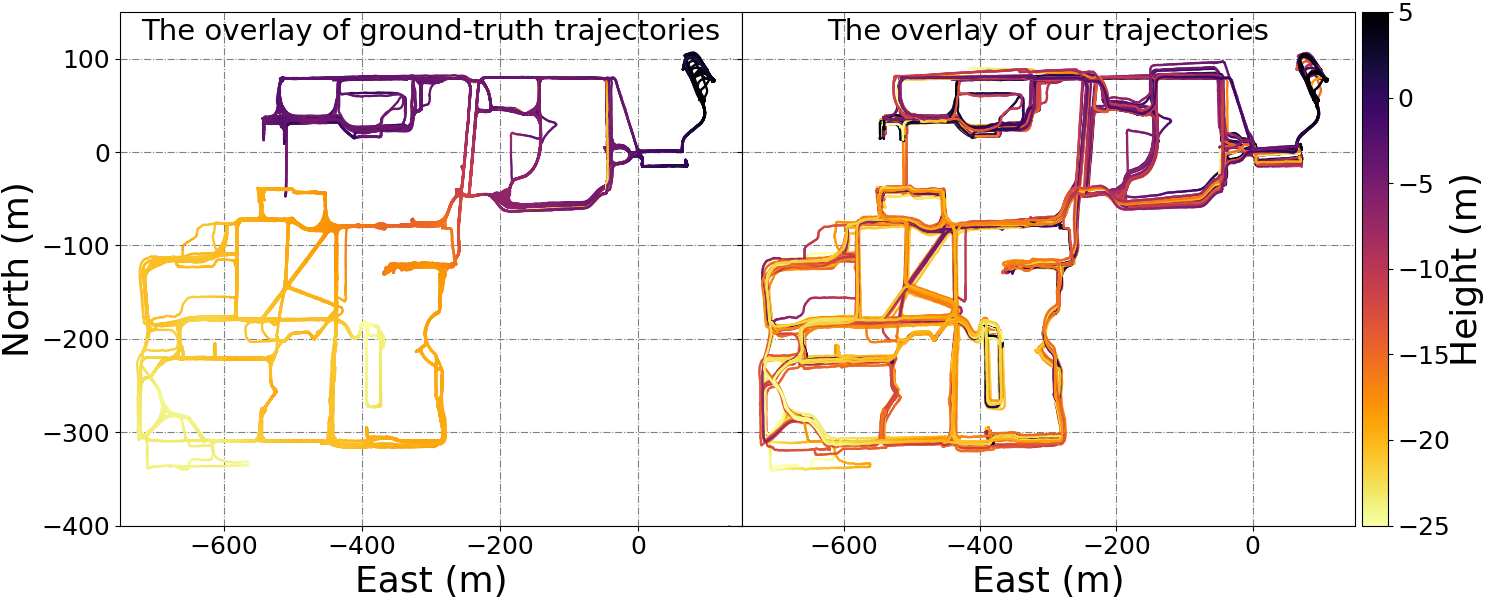}
\captionof{figure}{The overlay of ground-true trajectory and ours on NCLT-dataset.}\label{fig_overlay_figure}
\end{table*}

\subsection{Experiment-1: Evaluation of localization accuracy}\label{sect_experiment_1}

In this experiment, we benchmark the localization accuracy of our systems against other state-of-the-art SLAM systems, including \textit{LIO-SAM}\cite{shan2020lio}, \textit{LVI-SAM}\cite{lvisam}, \textit{FAST-LIO2}\cite{fastlio2}, and our previous work \textit{R$^2$LIVE}\cite{r2live}, on the NCLT-dataset\cite{carlevaris2016university}. {\textit{LIO-SAM} and \textit{FAST-LIO2} are LiDAR-inertial systems without fusing image data (Section~\ref{sect_LiDAR_Inertial}), while \textit{LVI-SAM} and \textit{R$^2$LIVE} are two feature-based LiDAR-inertial-visual systems (see Section~\ref{sect_LiDAR_Inertial_Visual}).} {Since our work is a state estimator without any loop detection and correction, we deactivated the loop closure of \textit{LIO-SAM} and \textit{LVI-SAM} for the sake of fair comparison.}  Since the camera photometric calibration of the NCLT-dataset is not available, we disable the photometric calibration modules of our VIO-subsystem by using $V(\cdot) = 1$ and $\mathbf f_i(\cdot) = 1$. 

Table~\ref{TABLE_APE_NCLT} shows the absolute position error (APE)\cite{zhang2018tutorial} of these methods, where {\it Our-LIO} is the LIO subsystem of our system. \textit{LIO-SAM} and \textit{LVI-SAM} failed in some sequences, and these sequences are excluded from the computation of the average APE. As can be seen from this table, with the average APE only \SI{8.51}{\meter}, our proposed system achieves the best overall performance than feature-based LiDAR-inertial-visual systems \textit{R$^2$LIVE} and \textit{LVI-SAM}. The performance improvement mainly come from the direct method used in the LIO subsystem and the tight-coupling of the LIO and VIO subsystems, the former can be seen by comparing the direct method {\it FAST-LIO2} to the feature-based method {\it LIO-SAM} in Table~\ref{TABLE_APE_NCLT} (and also detailed in \cite{fastlio2}), the latter improves the accuracy of the VIO subsystem (hence the complete system) by leveraging the high-accuracy geometry structure reconstructed from the LiDAR. Furthermore, the overall APE of our system is lower than its LIO subsystem {\it Our-LIO} and the other LIO systems (i.e., {\it FAST-LIO2} and {\it LIO-SAM}), which confirms the effectiveness of fusing camera data. Indeed, we found that in the evaluated sequences, the LiDAR sensor may occasionally face the sun, which creates a large number of noisy LiDAR points and adversely affect the LIO accuracy. There are certain sequences where our system is slightly outperformed by its LIO subsystem and the other LIO systems, the reason residues {in moving objects (e.g., pedestrians, bicyclists, and cars ) which may adversely affect the VIO (hence the overall system).} 
In Fig. \ref{fig_overlay_figure}, we overlay all the 25 ground-true trajectories (in the left figure) and ours (the right one). As can be seen, the overlaid trajectories estimated by our system agree with the ground-truth well and each trajectory can still be clearly distinguished without noticeable errors. Note that these 25 trajectories are collected on the same campus area across different time in a day and seasons in a year, still our system can produce consistent and reliable trajectory estimation with these illumination and scene changes, demonstrating the robustness of our system. 


\subsection{Experiment-2: Evaluation of robustness}\label{sect_experiment_2}

\begin{figure}[t]
	\centering
	\vspace{-0.45cm}
	\setcounter{figure}{8} 
	\includegraphics[width=0.93\linewidth]{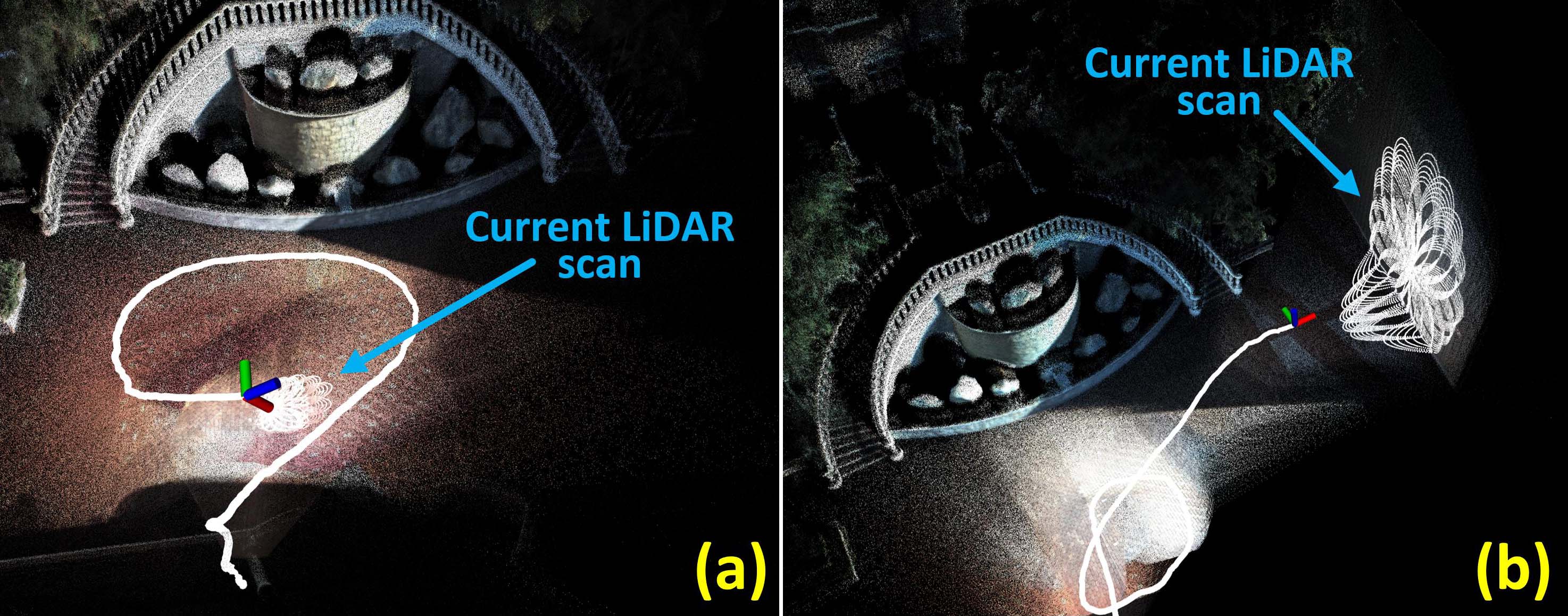}
	\includegraphics[width=0.93\linewidth]{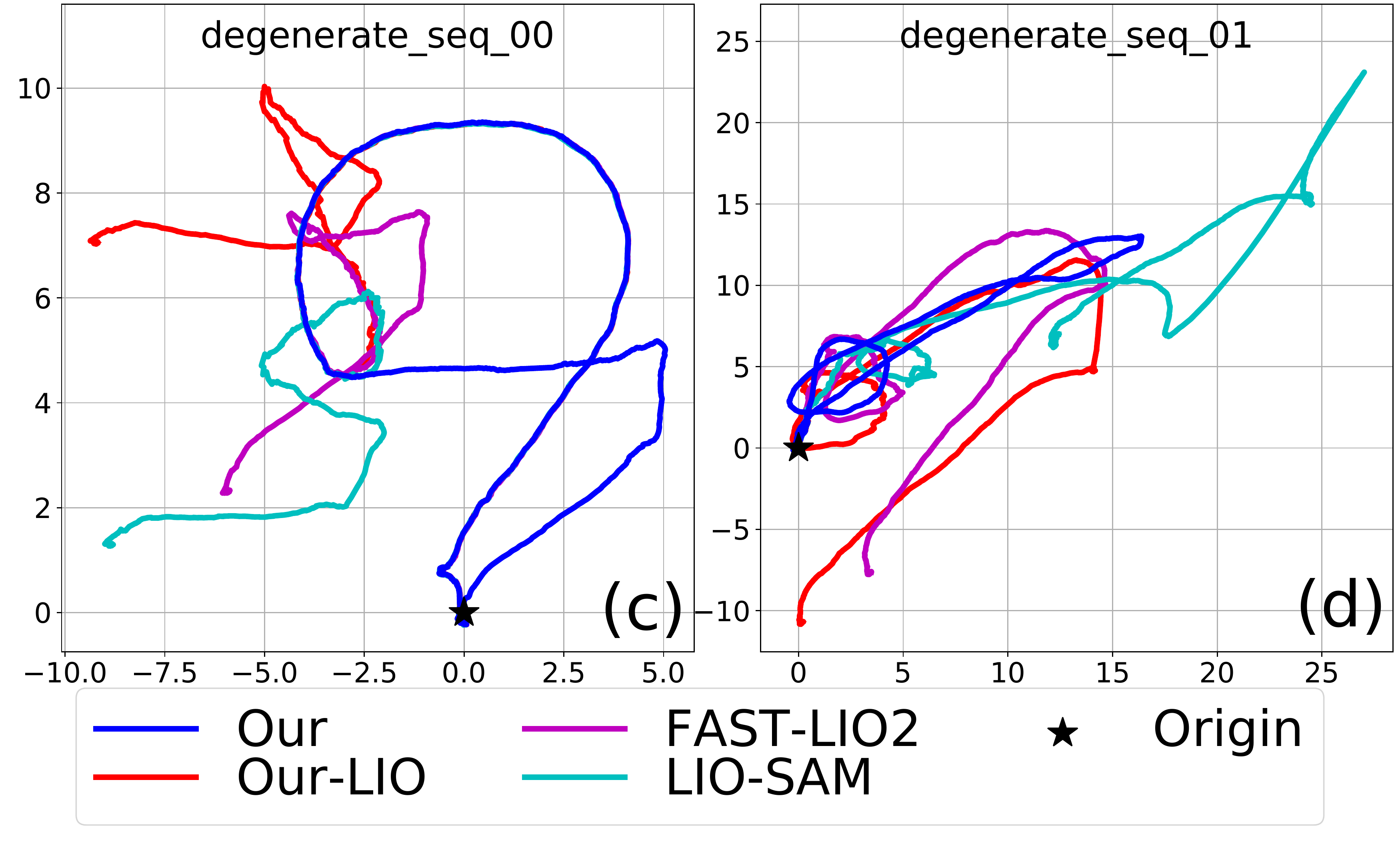}
	\caption{Tests in LiDAR degenerated environments.}
	\vspace{-0.5cm}
	\label{fig_LiDAR_degenerate}
\end{figure}

Besides illumination and scene change, we also test the robustness of our system to extreme scenarios where sensor degeneration occurs. We use the R$^3$LIVE-dataset, which contains such extreme scenarios. 

\subsubsection{Evaluation of robustness in LiDAR degenerated scenarios}
In this experiment, we evaluate the robustness of our proposed system by testing our system on the sequence ``degenerate\_seq\_00" and ``degenerate\_seq\_01" of R$^3$LIVE-dataset (see Section \ref{sect_r3live_dataset}). These two sequences were collected in front of a stairway with the LiDAR occasionally facing against the ground and a side wall (see Fig. \ref{fig_LiDAR_degenerate}(a) and (b)). When facing a wall, the LiDAR only observes a single plane, which is insufficient to determine the LiDAR pose, causing LiDAR degeneration. The device starts from and ends at the same location, enabling the evaluation of localization drift. The estimated trajectories of our proposed system, our LIO-subsystem \textit{Our-LIO},  and another two LIO systems, \textit{FAST-LIO2} and \textit{LIO-SAM}, are shown in Fig. \ref{fig_LiDAR_degenerate}(c) and (d). As can be seen, due to the LiDAR degeneration when facing a single plane, all three LiDAR-inertial odometry systems failed and did not return to the starting point. In contrast, by exploiting clues from the visual images, our proposed system works well in these two sequences and successfully returned to the starting point with drift down to {\SI{4.1}{\centi\meter} and \SI{4.6}{\centi\meter} on sequences ``degenerate\_seq\_00" and ``degenerate\_seq\_01", respectively.} To obtain a more intuitive comprehension of the LiDAR degenerated scenarios, we recommend our readers to watch the accompanying video on YouTube:  \href{https://youtu.be/qXrnIfn-7yA?t=390}{\small\tt{youtu.be/qXrnIfn-7yA?t=390}}. 

\subsubsection{Evaluate of robustness in simultaneously LiDAR degenerated and visual texture-less environments}
In this experiment, we challenge one of the most difficult scenarios in SLAM, where both LiDAR and camera degenerate. We use sequence ``degenerate\_seq\_02" of the R$^3$LIVE-dataset, where the sensor device passes through a narrow ``T"-shape passage (see Fig. \ref{fig_exp_1}) while occasionally facing against the side walls, causing LiDAR degeneration. Moreover, the visual texture on the white walls is very limited (Fig. \ref{fig_exp_1}(a) and Fig. \ref{fig_exp_1}(c)), especially for the wall-1, which has only changes in illumination. The absence of available LiDAR and visual features makes such scenarios rather challenging for both LiDAR-based and visual-based SLAM methods.

\begin{figure}[t]
	\centering
	\setcounter{figure}{10} 
	\includegraphics[width=1.0\linewidth]{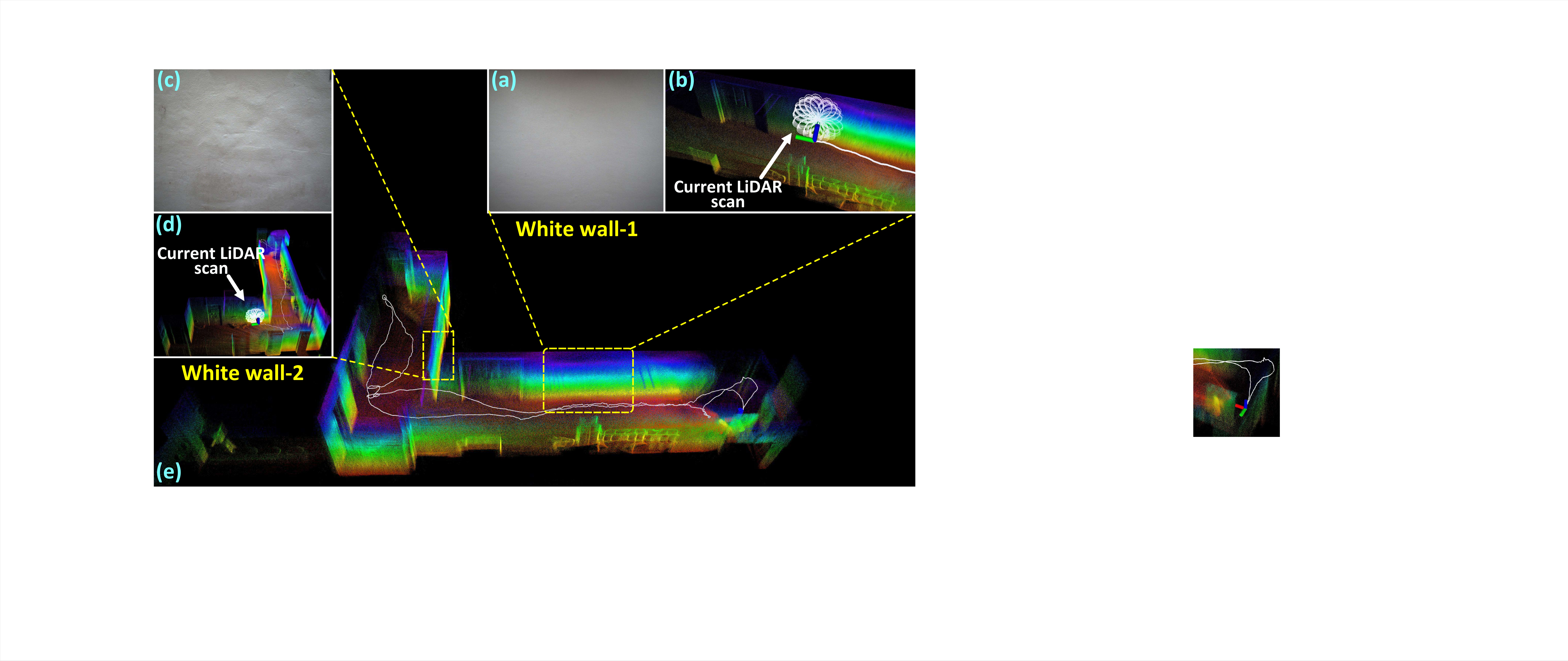}
	\vspace{-0.5cm}
	\caption{Tests in simultaneously LiDAR degenerated and visual texture-less environments.}
	\label{fig_exp_1}
	\vspace{0.2cm}
	\includegraphics[width=1.0\linewidth]{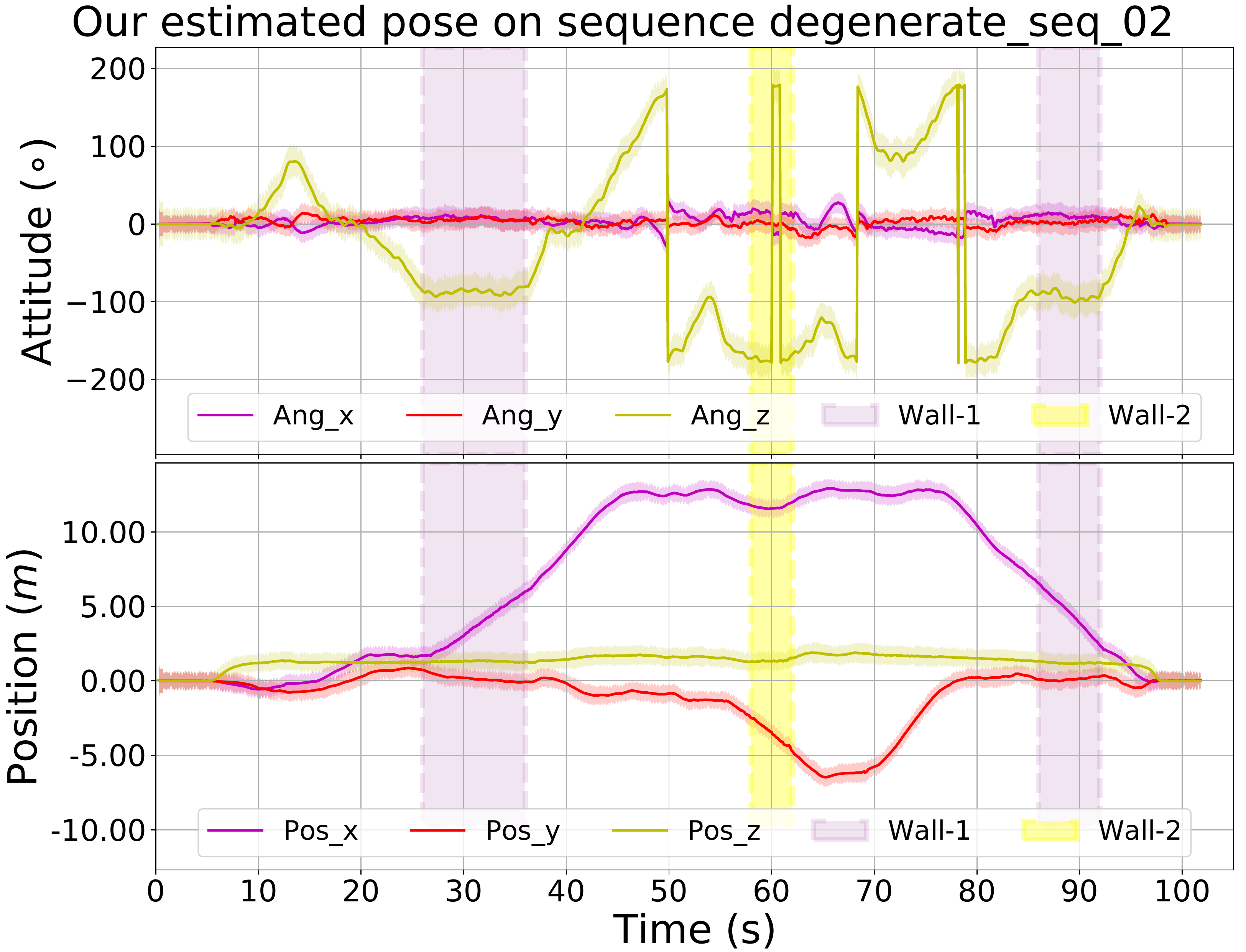}
	\vspace{-0.5cm}
	\caption{The estimated poses and their 3-$\sigma$ bound with 5 times amplification for better visualization (the light-colored area around the trajectory) of the test in simultaneously LiDAR degenerated and visual texture-less environments. The shaded areas in purple and yellow are the phases of the sensors facing against the white ``wall-1" and ``wall-2", respectively.}
	\label{fig_exp_1_pose}
	\vspace{-0.4cm}
\end{figure}

\begin{figure*}[t]
	\vspace{-0.2cm}
	\setcounter{figure}{13} 
	\includegraphics[width=1.0\linewidth]{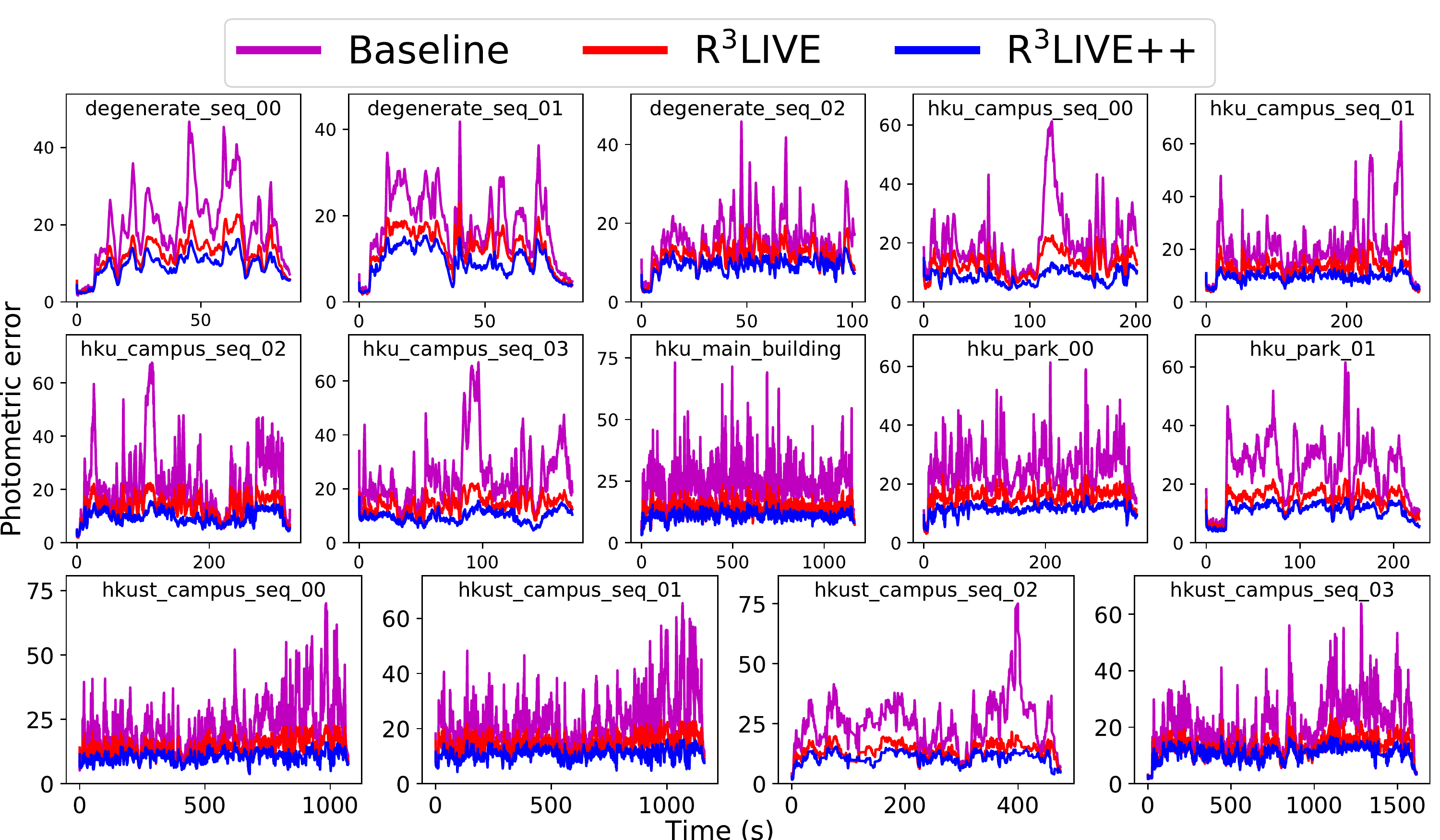}
	\vspace{-0.5cm}
	\captionof{figure}{Photometric errors between the reconstructed radiance map and image pixels.}
	\label{fig_avr_phorometric_error}
	\includegraphics[width=1.0\linewidth]{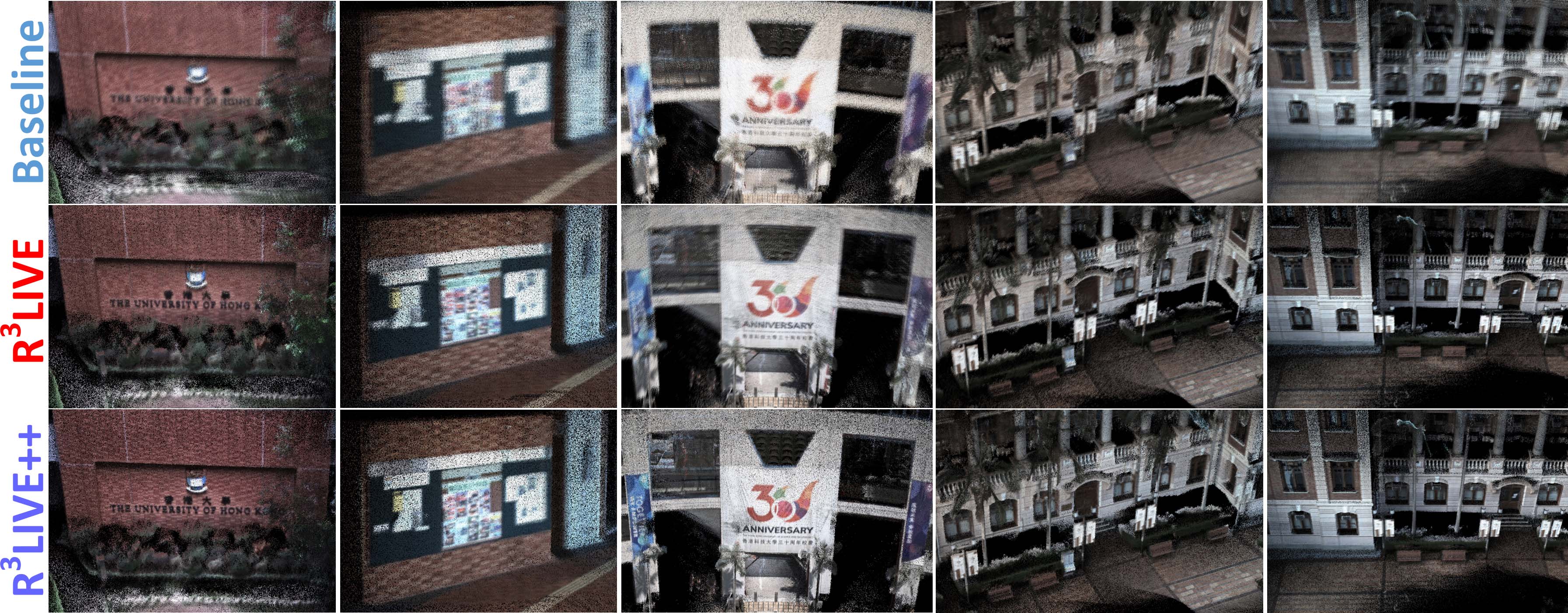}
	\vspace{-0.5cm}
	\captionof{figure}{Closeup of a few scenes in the radiance map reconstructed by the baseline, R$^3$LIVE and R$^3$LIVE++.}
	\label{fig_closeup_comparison}
	\vspace{-0.2cm}
\end{figure*}

Taking advantage of the raw pixel color information and tightly fusing it with the LiDAR point cloud measurements, our proposed algorithm can ``survive" in such extremely difficult scenarios. Fig. \ref{fig_exp_1_pose} shows our estimated pose, with the phases of passing through ``wall-1" and ``wall-2" shaded with purple and yellow, respectively. The estimated covariance is also shown in Fig. \ref{fig_exp_1_pose}, which is bounded over the entire estimated trajectory, indicating that our estimation quality is stable over the entire process. The sensor is moved to the starting point, where an ArUco marker board is used to obtain the ground-true relative pose between the starting and end poses. Compared with the ground-true end pose, our algorithm drifts  $1.62 ^\circ$ in rotation and $4.57 $ cm in translation. We recommend the readers to the accompanying video on YouTube (\href{https://youtu.be/qXrnIfn-7yA?t=461}{\tt{youtu.be/qXrnIfn-7yA?t=461}}) for better visualization of the experiment.

	
\subsection{Experiment-3: Evaluation of radiance map reconstruction}\label{sect_experiment_3}

In this experiment, we evaluate the accuracy of our proposed algorithm in reconstructing the radiance map. Since the ground-true radiance map of the environment can not be measured, we evaluate the accuracy based on two indicators: one is the estimation quality of the camera exposure time and the other is the average photometric error between the reconstructed radiance map and the measured images. 

\begin{table}[t]
	\centering
	\setcounter{figure}{12} 
	\includegraphics[width=1.0\linewidth]{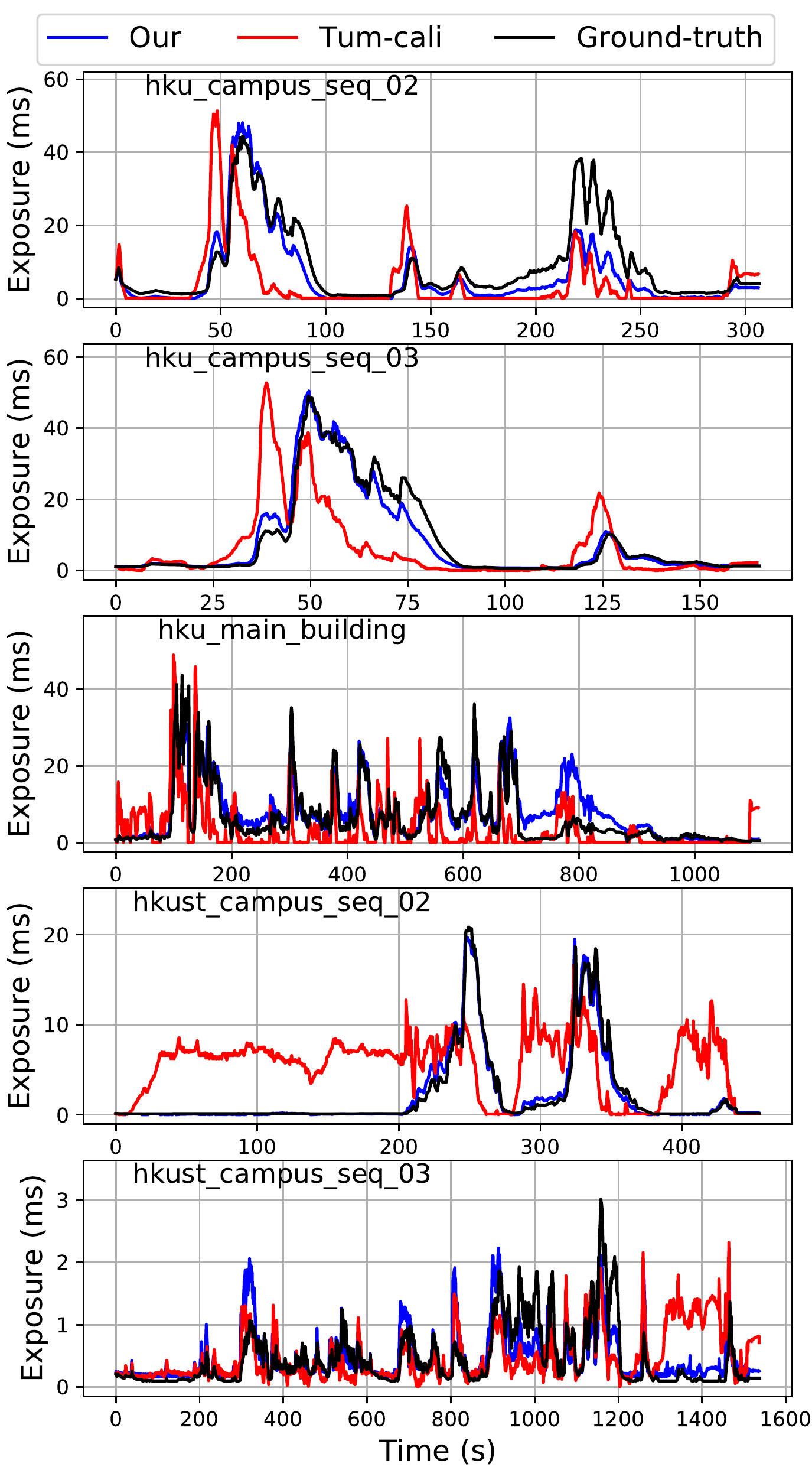}
	\captionof{figure}{Estimation of camera exposure time.}
	\label{fig_exposure_com}
	\captionof{table}{The comparison of estimation error of exposure time over 5 sequences.}
	\small
	\setlength{\tabcolsep}{4pt}     
	\renewcommand{\arraystretch}{1.15} 
	\vspace{-0.25cm}
	\begin{tabular}{ccc}
		\toprule
		\textbf{Sequence}    & \textbf{\begin{tabular}[c]{@{}c@{}}Our\\ Mean / Max (ms)\end{tabular}} & \textbf{\begin{tabular}[c]{@{}c@{}}Tum\_cali\\ Mean / Max (ms)\end{tabular}} \\ \hline
		hku\_campus\_seq\_02 & 3.460 / 20.311                                                & 7.082 / 36.175                                                      \\ 
		hku\_campus\_seq\_03 & 1.460 / 10.653                                                & 6.400 / 37.126                                                      \\ 
		hku\_main\_building   & 2.572 / 16.855                                                & 5.196 / 26.775                                                      \\ 
		hkust\_campus\_seq\_02       & 0.302 / 3.514                                                 & 5.225 / 13.361                                                      \\ 
		hkust\_campus\_seq\_03       & 0.189 / 1.185                                                 & 0.341 / 1.451                                                       \\ 
		\toprule
		\end{tabular}
	\label{Table_exp_error}
	\vspace{-1.5cm}
\end{table}

\subsubsection{Evaluation of exposure time estimation}
In this experiment, we evaluate the accuracy of the estimated camera exposure time by comparing it with the ground-true value read from the camera's API. We use four sequences (see Fig.~\ref{fig_exposure_com}) of the R$^3$LIVE-dataset, where the data were collected by traveling through both interior and exterior of a complex building to ensure significant changes in lighting conditions. We compare our estimated results with \textit{Tum-cali} \cite{bergmann2017online}, which is currently the only work that can estimate the camera's exposure time online to our knowledge. Both our system and \cite{bergmann2017online} are initialized by assuming the exposure time of the first image frame is at a default value \SI{1}{\milli\second}. 

The results are shown in Fig.~\ref{fig_exposure_com}, where the estimated exposure time of both our method and \textit{Tum-cali} \cite{bergmann2017online} is re-scaled to match with the ground-truth for better visualization. The average and maximum error of estimated exposure time w.r.t. the ground-truth is listed in Table~\ref{Table_exp_error}. As shown in Fig.~\ref{fig_exposure_com} and Table~\ref{Table_exp_error}, our proposed method shows significantly lower estimation error than \cite{bergmann2017online}. This is mainly because our method estimates the exposure time by minimizing the scan-to-map radiance error, while \cite{bergmann2017online} recovers the exposure times from consecutive frames. The consequence is that our method can better utilize longer-term temporal intensity changes to restrain the drift of the exposure time estimation.

\subsubsection{Evaluation of radiance map}

\begin{figure*}[t]
	\centering
	\setcounter{figure}{15} 
	\includegraphics[width=1.00\linewidth]{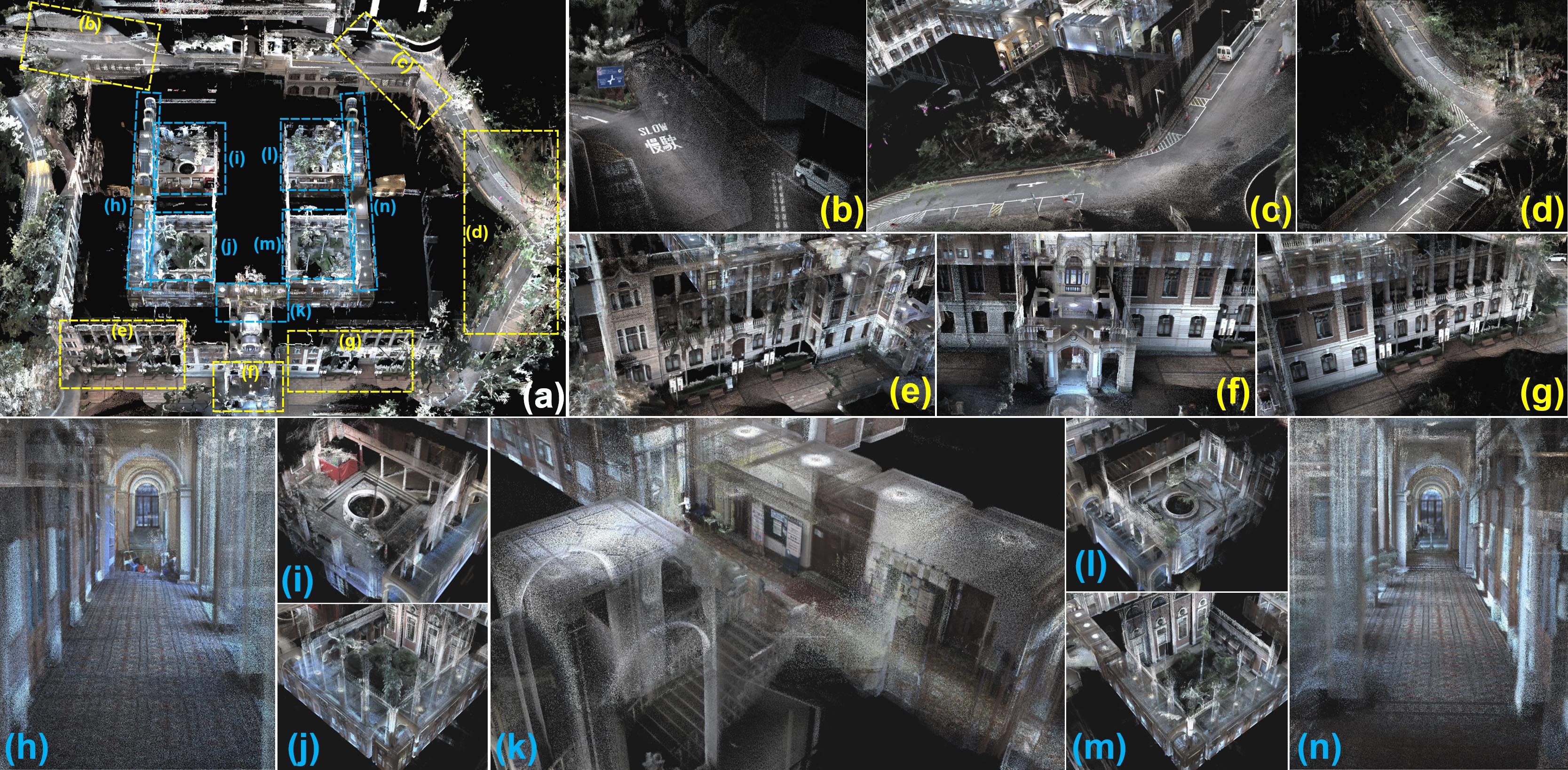}
	\vspace{-0.5cm}
	\captionof{figure}{Our reconstructed radiance map of the main building of HKU. (a) The bird's view of the map, with its details shown in (b$\sim$n). (b$\sim$g) closeup of outdoor scenarios and (h$\sim$n) closeup of indoor scenarios. To see the real-time reconstruction process of the map, please refer to the video on YouTube:  \href{https://youtu.be/qXrnIfn-7yA?t=55}{\tt{youtu.be/qXrnIfn-7yA?t=55}}.}
	\label{fig_reconstruct_main_building}
	\vspace{-0.5cm}
\end{figure*}

\begin{table}[]
	\scriptsize
	\centering
	\vspace{-0.35cm}
	\setlength{\tabcolsep}{8pt}     
	\renewcommand{\arraystretch}{1.15} 
	\caption{The average photometric error among all sequences of R$^3$LIVE-dataset.}
	\vspace{-0.2cm}
	\begin{tabular}{ccccc}
		\toprule
		\textbf{Sequence}    & \textbf{Frames} & \textbf{baseline} & \textbf{R$^3$LIVE} & \textbf{R$^3$LIVE++} \\ \hline
		degenerate\_seq\_00  & 1694                   & 30.58             & 21.36           & 16.19             \\ 
		degenerate\_seq\_01  & 1715                   & 34.24             & 21.28           & 16.55             \\ 
		degenerate\_seq\_02  & 3315                   & 27.14             & 20.30           & 15.97             \\ 
		hku\_campus\_seq\_00 & 3016                   & 34.78             & 22.56           & 14.57             \\ 
		hku\_campus\_seq\_01 & 4502                   & 34.97             & 22.47           & 16.30             \\ 
		hku\_campus\_seq\_02 & 2595                  & 39.01             & 23.73           & 16.85             \\ 
		hku\_campus\_seq\_03 & 4845                   & 42.95             & 24.78           & 16.93             \\ 
		hku\_main\_building  & 12157                  & 43.12             & 22.26           & 19.04             \\ 
		hku\_park\_00        & 5251                   & 43.86             & 27.01           & 20.07             \\ 
		hku\_park\_01        & 3410                   & 43.29             & 25.17           & 19.02             \\ 
		hkust\_campus\_00    & 16075                  & 37.64             & 24.19           & 18.00             \\ 
		hkust\_campus\_01    & 17426                  & 39.09             & 24.67           & 18.53             \\ 
		hkust\_campus\_02    & 4031                   & 41.74             & 23.77           & 18.92             \\ 
		hkust\_campus\_03    & 24270                  & 35.60             & 22.37           & 18.65             \\ \hline
		\textbf{Average}     & 7413.57                & 38.60             & 23.58           & 18.01             \\ \toprule
	\end{tabular}
	\label{TAB_ERROR_EXP}
	\vspace{-1.3cm}
\end{table}

\begin{figure}[h]
	\centering
	\includegraphics[width=0.93\linewidth]{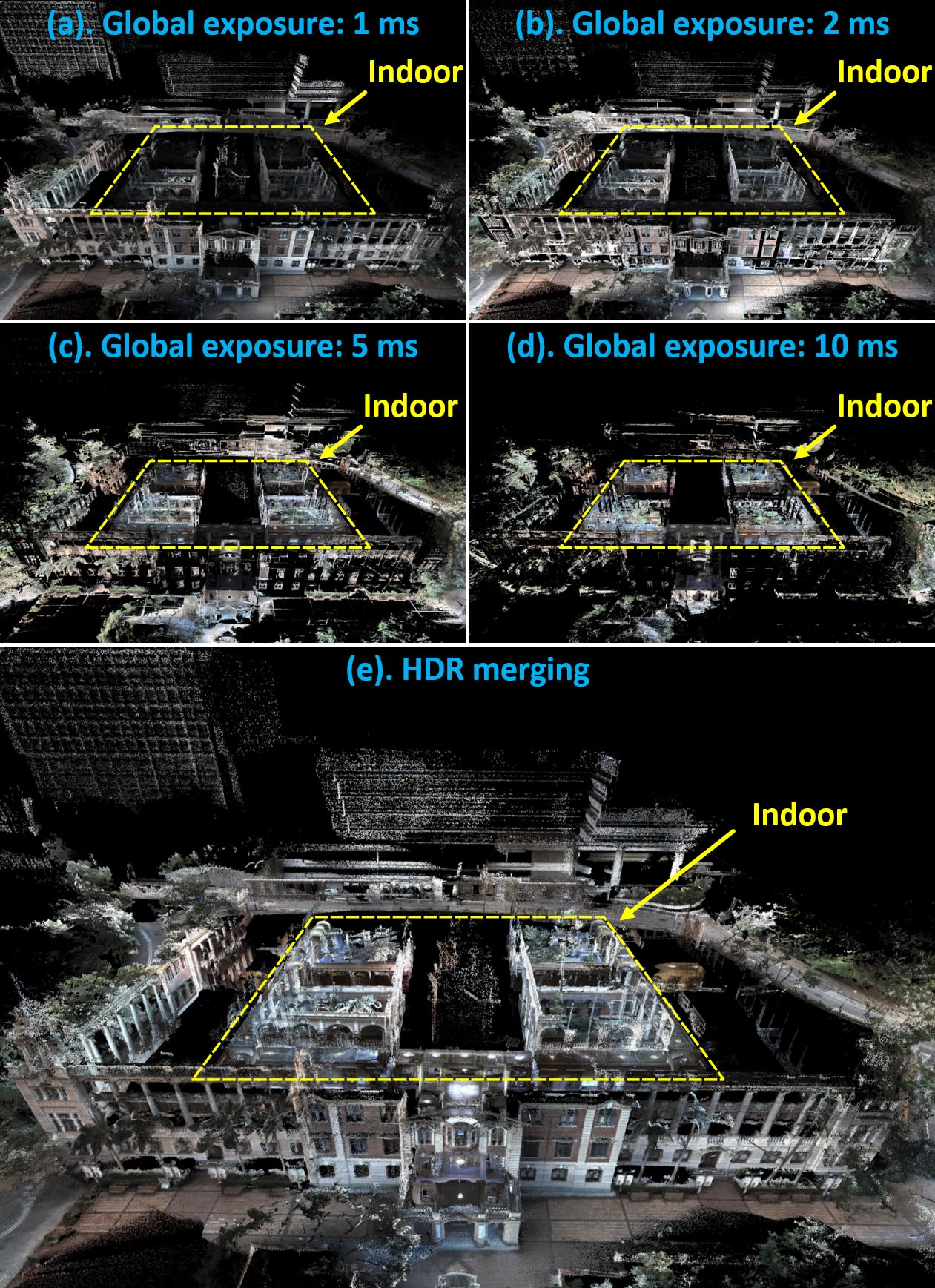}
	\vspace{-0.2cm}
	\caption{ (a$\sim$~d) Images rendered from the reconstructed radiance map at different exposure time: \SI{1}{\milli\second}, \SI{2}{\milli\second}, \SI{5}{\milli\second} and \SI{10}{\milli\second}. (e) The HDR image merged from (a$\sim$d). Notice that for the sake of better visualization, those points that are over-exposure are not displayed.}
	\label{fig_hdr_image_merge}
	\vspace{-0.6cm}
\end{figure}

In this experiment, we evaluate the accuracy of our proposed algorithm in reconstructing the radiance map. Currently, LiDAR point cloud colorization remains one of the most challenging problems in the field of 3D reconstruction. The most common way is using the most recent image frame in time to give the color of each LiDAR frame \cite{camvox, fastlivo}. The preliminary implementation of our system R$^3$LIVE published previously \cite{r3live} colorized the point cloud by minimizing the photometric error similar to our current system but does not consider any exposure time estimation or photometric calibration. In this experiment, we compare our system against the previous implementation R$^3$LIVE \cite{r3live} to show the effectiveness of the exposure time estimation and photometric calibration and against the current baseline\cite{camvox, fastlivo} to show the advantage of the overall system. 

To assess the radiance reconstruction error, after the map reconstruction, we re-project all points in the map with radiance information to each image frame with the estimated camera pose and calibrated photometric parameters. Then, we calculate the photometric error between the map point color and the RGB values of the image at the projected pixel location. The average photometric error of each image frame is calculated to evaluate the accuracy of the reconstructed radiance map. We perform the evaluation on all R$^3$LIVE-dataset sequences, with the results of each sequence are given in Fig.~\ref{fig_avr_phorometric_error}, and the average photometric of each sequence is listed in Table~\ref{TAB_ERROR_EXP}. As can be seen, our system has consistently achieved the lowest photometric errors in all sequences and the next best is R$^3$LIVE. Fig. \ref{fig_closeup_comparison} shows a few closeups of the reconstructed radiance map, from which we can clearly tell the words on objects. Moreover, in Fig.~\ref{fig_reconstruct_main_building}, we present the reconstructed radiance map of the sequence ``hku\_main\_building" in R$^3$LIVE-dataset, in which we collect the data in both interior and exterior of the main building of HKU. As shown in Fig.~\ref{fig_reconstruct_main_building}, both the indoor and outdoor details (e.g., marks on the road) are very clear, demonstrating that our proposed algorithm is of high accuracy.
What worth mentioning is, this 3D radiance map is reconstructed on the fly as the data is being acquired (see the accompanying video on YouTube:  \href{https://youtu.be/qXrnIfn-7yA?t=55}{\small\tt{youtu.be/qXrnIfn-7yA?t=55}}). For more qualitative results of other sequences, we refer our readers to our 
 Supplementary Material: \href{https://github.com/hku-mars/r3live/blob/master/supply/r3live_plus_plus_supplementary_material.pdf}{\small\tt{github.com/hku-mars/r3live/blob/master/supply/\\r3live\_plus\_plus\_supplementary\_material.pdf}}

\subsection{Run time analysis}

In this section, We investigate the average time consumption of our proposed system on a CPU-Only PC (equipped with an Intel i7-9700K CPU and \SI{64}GB RAM). We counts the average time consumption on all sequences of both datasets (i.e., the NCLT-dataset and R$^3$LIVE-dataset), whose results are shown in Table~\ref{tab_rumtime_nclt} and Table~\ref{tab_rumtime_r3live_dataset}, respectively. {For the NCLT-dataset, each LiDAR scan takes an average of \SI{34.3}{\milli\second} and each camera image takes an average of \SI{16.6}{\milli\second} processing time. Since the data rate of the LiDAR and camera sensors are  \SI{10}{\hertz} and \SI{5}{\hertz}, the total processing time per second is \SI{426}{\milli\second}, comprising of the time for processing 10 lidar scans (i.e., \SI{343}{\milli\second}) and 5 images (\SI{83}{\milli\second}).  For R$^3$LIVE-dataset, each LiDAR scan takes an average of \SI{22.7}{\milli\second} and each camera image takes an average of \SI{16.2}{\milli\second} processing time. Since the data rate of the LiDAR and camera sensors are \SI{10}{\hertz} and \SI{15}{\hertz}, the total processing time per second is \SI{470}{\milli\second}, comprising of the time for processing 10 lidar scans (i.e., \SI{235}{\milli\second}) and 15 images (\SI{244}{\milli\second}). In both cases, the processing time required per second is below half a second, indicating that our system runs in real-time (even two times faster than real-time) in both pose estimation and radiance map reconstruction. In addition, the processing time per image is similar across the two datasets while that for each lidar scan is quite different. The reason is that the lidar in R$^3$LIVE-dataset has a much lower data rate than that of the NCLT-dataset {(\SI{240}{k} versus \SI{695}{k} points per second)} while the cameras have similar resolution. }

\begin{table}[]
	\centering
	\caption{The average time consumption per LiDAR or camera frame of our system on NCLT-dataset.}
	\setlength{\tabcolsep}{10pt}     
	\renewcommand{\arraystretch}{1.05} 
	\label{tab_rumtime_nclt}
	\vspace{-0.25cm}
		\begin{tabular}{ccc}
			\toprule
			\textbf{\begin{tabular}[c]{@{}c@{}}NCLT-dataset\\ Sequence (date)\end{tabular}} & \textbf{\begin{tabular}[c]{@{}c@{}}LiDAR frame\\ Mean / Std (ms)\end{tabular}} & \textbf{\begin{tabular}[c]{@{}c@{}}Camera frame\\ Mean / Std  (ms)\end{tabular}} \\ \hline
			2012-01-08                                                         & 33.852 / 9.110                                                                 & 15.911 / 4.180                                                                   \\ 
			2012-01-15                                                         & 35.517 / 11.602                                                                & 17.613 / 4.275                                                                   \\ 
			2012-01-22                                                         & 34.392 / 11.719                                                                & 16.551 / 4.591                                                                   \\ 
			2012-02-02                                                         & 33.812 / 11.188                                                                & 16.926 / 4.361                                                                   \\ 
			2012-02-04                                                         & 32.599 / 10.498                                                                & 15.09 / 4.164                                                                    \\ 
			2012-02-05                                                         & 34.823 / 11.147                                                                & 17.09 / 4.276                                                                    \\ 
			2012-02-12                                                         & 32.738 / 11.765                                                                & 17.071 / 3.846                                                                   \\ 
			2012-02-18                                                         & 37.284 / 10.169                                                                & 17.087 /  3.973                                                                  \\ 
			2012-02-19                                                         & 38.004 / 9.928                                                                 & 18.179 / 4.099                                                                   \\ 
			2012-03-17                                                         & 32.196 / 11.154                                                                & 17.304 / 4.370                                                                   \\ 
			2012-03-31                                                         & 36.283 / 10.377                                                                & 16.228 / 4.312                                                                   \\ 
			2012-04-29                                                         & 33.652 / 9.014                                                                 & 16.487 / 4.018                                                                   \\ 
			2012-05-11                                                         & 34.044 / 8.964                                                                 & 17.357 / 3.811                                                                   \\ 
			2012-05-26                                                & 37.623 / 11.315                                                                & 15.228 / 4.384                                                                   \\ 
			2012-06-15                                                         & 29.07 / 8.807                                                                  & 17.254 / 3.881                                                                   \\ 
			2012-08-04                                                         & 29.231 / 10.427                                                                & 17.267 / 4.629                                                                   \\ 
			2012-08-20                                                         & 28.561 / 10.952                                                                & 15.73 / 3.931                                                                    \\ 
			2012-09-28                                                         & 36.692 / 10.738                                                                & 15.458 / 4.346                                                                   \\ 
			2012-10-28                                                         & 36.147 / 11.497                                                                & 16.081 / 4.102                                                                   \\ 
			2012-11-04                                                         & 37.066 / 11.430                                                                & 16.242 / 4.061                                                                   \\ 
			2012-11-17                                                         & 38.153 / 9.187                                                                 & 16.931 / 3.904                                                                   \\ 
			2012-12-01                                                         & 36.357 / 11.660                                                                & 16.492 / 4.165                                                                   \\ 
			2013-01-10                                                         & 32.977 / 10.355                                                                & 18.328 / 3.971                                                                   \\ 
			2013-02-23                                                         & 36.171 / 10.720                                                                & 15.368 / 4.358                                                                   \\ 
			2013-04-05                                                         & 29.542 / 10.052                                                                & 15.725 / 4.189                                                                   \\ \hline
			\textbf{Average}                                                   & 34.271 / 10.551                                                                & 16.600 / 4.168                                                                   \\ \toprule
	\end{tabular}

	\centering
	\vspace{0.4cm}
	\caption{The average time consumption per LiDAR or camera frame of our system on R$^3$LIVE-dataset. }
	\renewcommand{\arraystretch}{1.05} 
	\setlength{\tabcolsep}{8pt}
	\label{tab_rumtime_r3live_dataset}
	\vspace{-0.25cm}
		\begin{tabular}{ccc}
		\toprule
		\textbf{\begin{tabular}[c]{@{}c@{}}Sequence\\ (date)\end{tabular}} & \textbf{\begin{tabular}[c]{@{}c@{}}LiDAR frame\\ Mean / Std (ms)\end{tabular}} & \textbf{\begin{tabular}[c]{@{}c@{}}Camera frame\\ Mean / Std  (ms)\end{tabular}} \\ \hline
		degenerate\_seq\_00                                                & 17.927 /  9.080                                                                & 13.057 / 2.675                                                                   \\ 
		degenerate\_seq\_01                                                & 8.111 / 3.622                                                                  & 12.111 / 2.502                                                                   \\ 
		degenerate\_seq\_02                                                & 21.638 / 6.793                                                                 & 13.543 / 3.014                                                                   \\ 
		hku\_campus\_seq\_00                                               & 27.659 / 8.159                                                                 & 18.829 / 3.719                                                                   \\ 
		hku\_campus\_seq\_01                                               & 25.328 / 11.246                                                                & 18.187 / 2.860                                                                   \\ 
		hku\_campus\_seq\_02                                               & 20.757 / 5.109                                                                 & 16.928 / 2.518                                                                   \\ 
		hku\_campus\_seq\_03                                               & 21.705 /  4.903                                                                & 16.040 /  2.338                                                                  \\ 
		hku\_main\_building                                                & 25.123 / 10.246                                                                & 17.023 / 2.512                                                                   \\ 
		hku\_park\_00                                                      & 26.908 /  7.196                                                                & 17.882 / 2.545                                                                   \\ 
		hku\_park\_01                                                      & 24.412 / 6.598                                                                 & 17.479 / 2.485                                                                   \\ 
		hkust\_campus\_seq\_00                                             & 28.160 / 9.703                                                                 & 16.624 /  2.756                                                                  \\ 
		hkust\_campus\_seq\_01                                             & 27.058 / 10.135                                                                & 16.138 /  2.813                                                                  \\ 
		hkust\_campus\_seq\_02                                             & 22.857 / 6.135                                                                 & 16.518 / 2.618                                                                   \\ 
		hkust\_campus\_seq\_03                                             & 30.757 / 5.109                                                                 & 16.928 / 2.518                                                                   \\ \hline
		\textbf{Average}                                                   & 23.453 / 7.431		                                                            & 16.234 / 2.705                                                                  \\ 
		\toprule		
	\end{tabular}
	\vspace{-1.2cm}
\end{table}

\section{Applications with R$^3$LIVE}

\subsection{High dynamic range (HDR) imaging}
After the reconstruction of the radiance map, we are able to render an image by projecting the map to an image plane with a given pose and exposure time with equation (\ref{eq_image_output}). Taking the sequence ``hku\_main\_building" as an example, Fig.~\ref{fig_hdr_image_merge}(a), (b), (c) and (d) are the rendered images with global exposure time of \SI{1}{\milli\second}, \SI{2}{\milli\second}, \SI{5}{\milli\second} and \SI{10}{\milli\second}, respectively.  These images rendered at different exposure times can be merged into a HDR image shown in Fig.~\ref{fig_hdr_image_merge} (e).

\subsection{Mesh reconstruction and texturing}
\begin{figure}[h]
	\centering
	\includegraphics[width=1.0\linewidth]{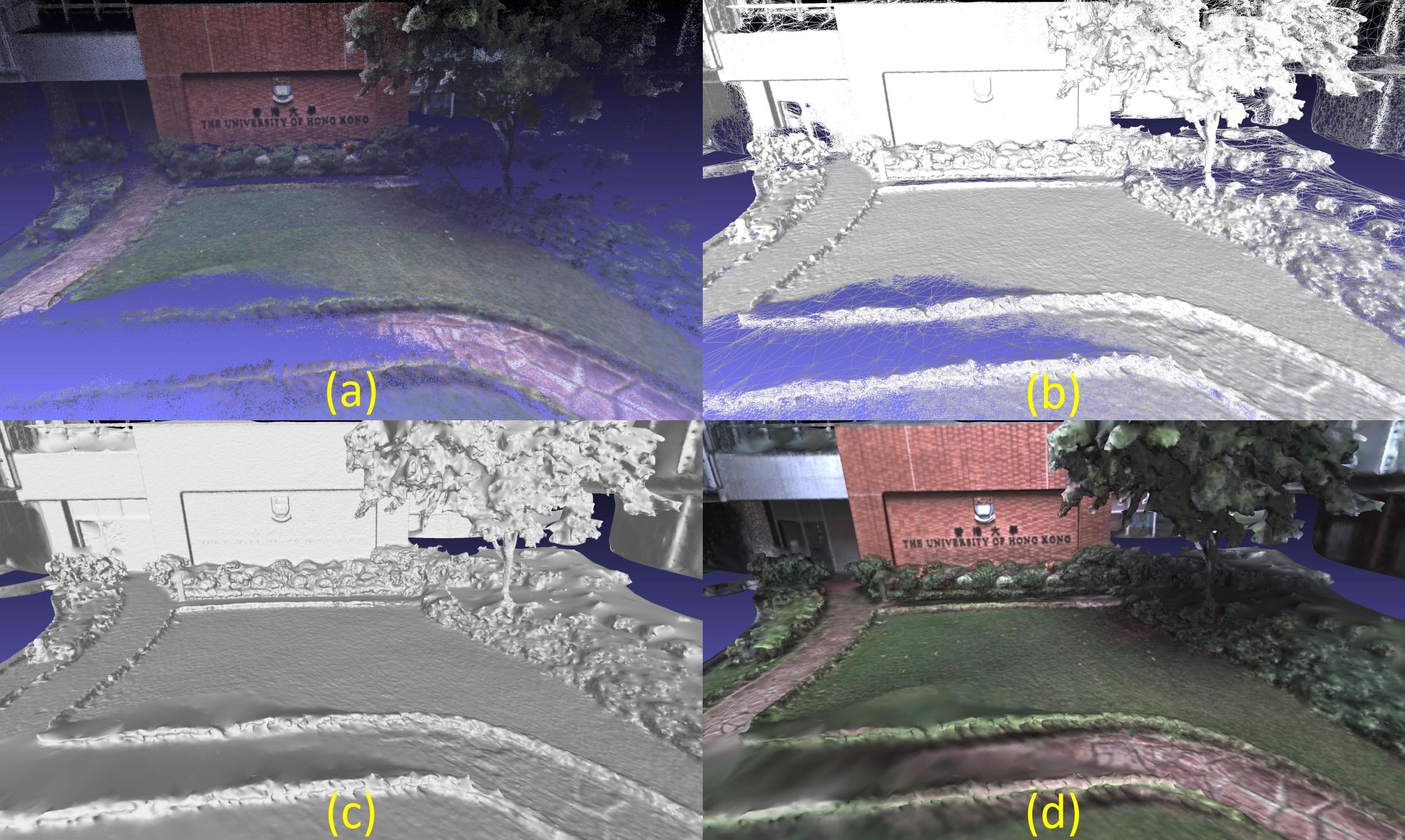}
	\caption{ (a) show the RGB-colored 3D points reconstructed by R$^3$LIVE++. (b) and (c) show the wireframe and surface of our reconstructed mesh. (d) show the mesh after texture rendering.}
	\vspace{-0.0cm}
	\label{fig_mesh}
\end{figure}
While R$^3$LIVE++ reconstructs the colored 3D map in real-time, we also develop software utilities to mesh and texture the reconstructed map offline (see Fig. \ref{fig_mesh}). For meshing, we make use of the Delaunay triangulation and graph cuts \cite{labatut2007efficient} implemented in CGAL\cite{fabri2009cgal} and openMVS\cite{openmvs2020}. After the mesh construction, we texture the mesh with the vertex (point) colors, with are rendered by our VIO subsystem. 

Our developed utilities also export the colored point map from R$^3$LIVE++ or the offline meshed map into commonly used file formats such as ``pcd", ``ply", ``obj", etc. As a result, the maps reconstructed by R$^3$LIVE++ can be imported into various 3D software, including but not limited to CloudCompare\cite{girardeau2016cloudcompare}, Meshlab\cite{cignoni2011meshlab}, AutoDesk 3ds Max\cite{autodesk3dsmax}.

\subsection{Toward various of 3D applications}
With the developed software utilities, we can export the reconstructed 3D maps to Unreal Engine\cite{unrealengine} to enable a series of 3D applications. For example, in Fig. \ref{fig_ue4_game}(a), we built a car simulator with the AirSim\cite{shah2018airsim}. In Fig. \ref{fig_ue4_game}(b) and Fig. \ref{fig_ue4_game}(c), we used our reconstructed maps to develop video games for mobile platforms and desktop PCs, respectively. To get more details about our demos, we refer the readers to watch our video on YoutuBe: \href{https://youtu.be/qXrnIfn-7yA?t=516}{\small\tt{youtu.be/qXrnIfn-7yA?t=516}}.

\begin{figure}[h]
	\centering
	\includegraphics[width=1.0 \linewidth]{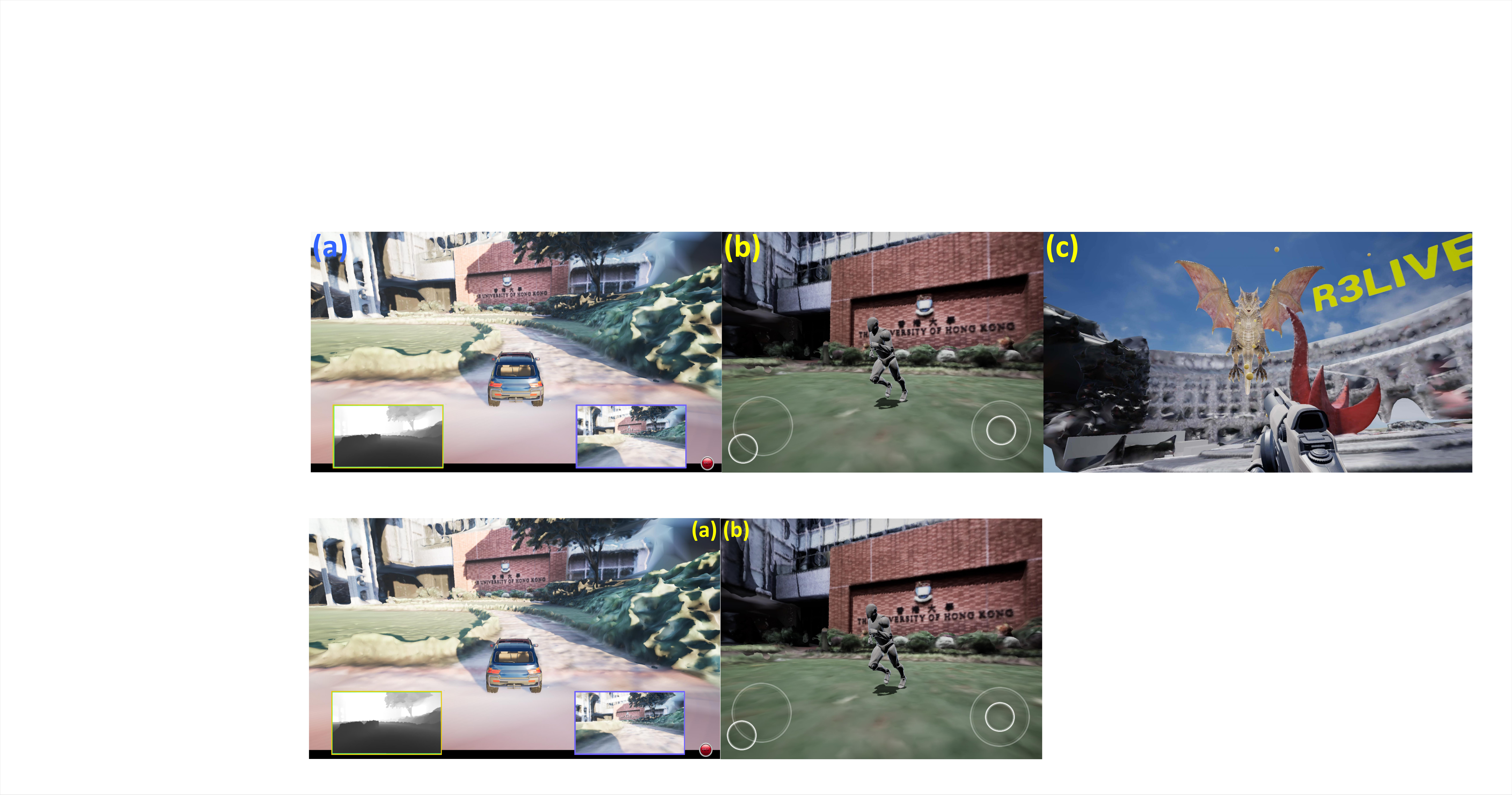}
	\caption{In (a), we built a car simulator with our maps and AirSim. The images in yellow and blue frame-boxes are the depth and RGB image query from the AirSim's API. In (b) and (c), we developed video games for mobile platforms and desktop PCs. The player in (b) is controlling the actor to explore the campus of HKU, and in (c) is fighting against a dragon by shooting rubber balls at HKUST.}
	\label{fig_ue4_game}
\end{figure}

\section{Conclusions and future work}
		\subsection{Conclusions}
		In this paper, we proposed a novel LiDAR-inertial-visual fusion framework termed R$^3$LIVE++ to achieve robust and accurate state estimation while simultaneously reconstructing the radiance map on the fly. This framework consists of two subsystems (i.e., the LIO and the VIO) that jointly and incrementally build a 3D radiance map of the environment in real-time. By tightly fusing the measurement of three different types of sensors, R$^3$LIVE++ can achieve higher localization accuracy while being robust enough to scenarios with sensor degenerations. 
		
		In our experiments, we extensively validated our proposed algorithm with real-world experiments in terms of localization accuracy, robustness, and radiance map reconstruction accuracy. The benchmark results on 25 sequences from an open dataset (the NCLT-dataset) showed that R$^3$LIVE++ achieved the highest overall accuracy among all other state-of-the-art SLAM systems under comparison. The evaluation on a private dataset (the R$^3$LIVE-dataset) showed that R$^3$LIVE++ was robust to extremely challenging scenarios that LiDAR and/or camera measurements degenerate (e.g., when the device is facing a single texture-less wall). Finally, compared with other counterparts, R$^3$LIVE++ estimates the camera exposure time more accurately and reconstructs the true radiance information of the environment with significantly smaller errors when compared to the measured values in images. 
		
		To demonstrate the extendability of our work, we developed several applications based on our reconstructed radiance maps, such as high dynamic range (HDR) imaging, virtual environment exploration, and 3D video gaming. Finally, to share our findings and make contributions to the community, we made our codes, hardware design, and dataset publicly available on our Github.

		\subsection{Future work}
		In R$^3$LIVE++, the radiance map is reconstructed with 3D points that contain radiance information, which prevents us from rendering high-resolution images from the radiance map due to the limited point cloud density of the radiance map (\SI{1}{\centi\meter} in our current implementation). While further increasing this point cloud density is possible, it will further increase the processing time. This point density is also limited by the density of the raw points measured by lidar sensors. Noticing that images often have much higher resolution, in the future, we could explore how to make full use of such high-resolution images in the fusion framework.


%
%
%
%

\ifCLASSOPTIONcaptionsoff
  \newpage
\fi

\bibliography{r3live_pp}
\vspace{-1cm}
\begin{IEEEbiography}
	[{\includegraphics[width=1in,height=1.25in,clip,keepaspectratio]{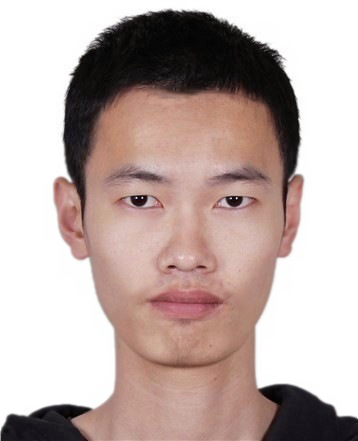}}]{Jiarong Lin} 
	received a B.S. degree in Optical Information Science and Technology from the University of Electronic Science and Technology of China (UESTC) in 2015. He is currently a Ph.D. candidate in the Department of Mechanical Engineering, the University of Hong Kong (HKU), Hong Kong, China. 

	His research interests include light detection and ranging (LiDAR) mapping and sensor fusion.
\end{IEEEbiography}
\vspace{-1.2cm}
\begin{IEEEbiography}
	[{\includegraphics[width=1in,height=1.25in,clip,keepaspectratio]{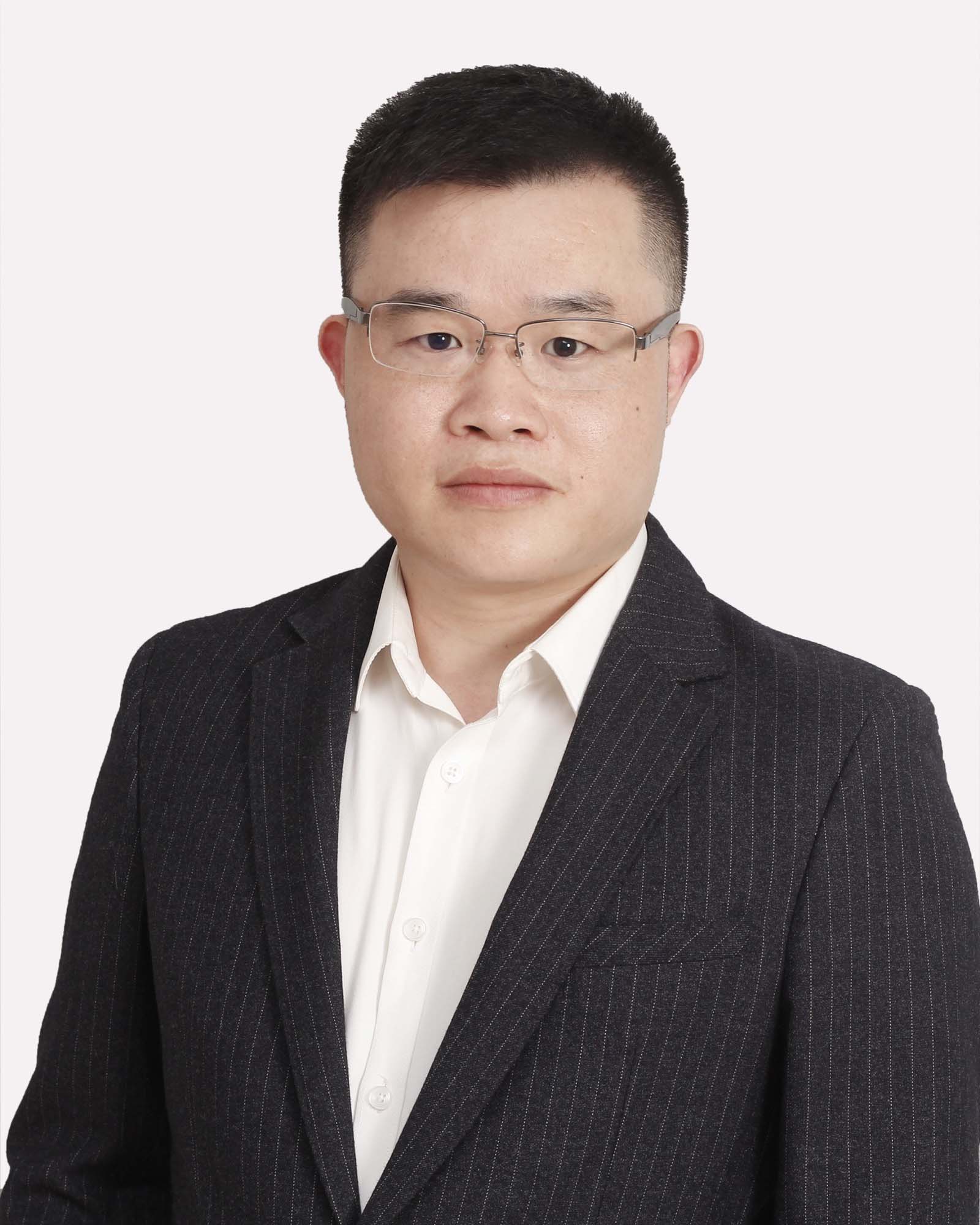}}]{Fu Zhang}
	received the B.E. degree in automation from the University of Science and Technology of China (USTC), Hefei, Anhui, China, in 2011, and the Ph.D. degree in controls from the University of California at Berkeley, Berkeley, CA, USA, in 2015. 
	
	He joined the Department of Mechanical Engineering, The University of Hong Kong (HKU), Hong Kong, as an Assistant Professor in August 2018. His current research interests are on robotics and controls, with a focus on unmanned aerial vehicle (UAV) design, navigation, control, and light detection and ranging (LiDAR)-based simultaneous localization and mapping (SLAM).
\end{IEEEbiography}


\vfill
\fi

\newpage
\clearpage
\setcounter{equation}{0}
\setcounter{figure}{0}
\setcounter{table}{0}
\setcounter{page}{1}
\setcounter{section}{1}
\setcounter{section}{0}%
\setcounter{subsection}{0}%
\setcounter{subsubsection}{0}%
\setcounter{paragraph}{0}%


\onecolumn
\begin{center}
	\textbf{\large Supplementary Material for R$^3$LIVE++: The qualitative results of our reconstructed radiance map on R$^3$LIVE-dataset}
\end{center}
\begin{figure}[h]
	\centering
	\begin{minipage}{0.85\linewidth}
	\includegraphics[width=1.0\linewidth]{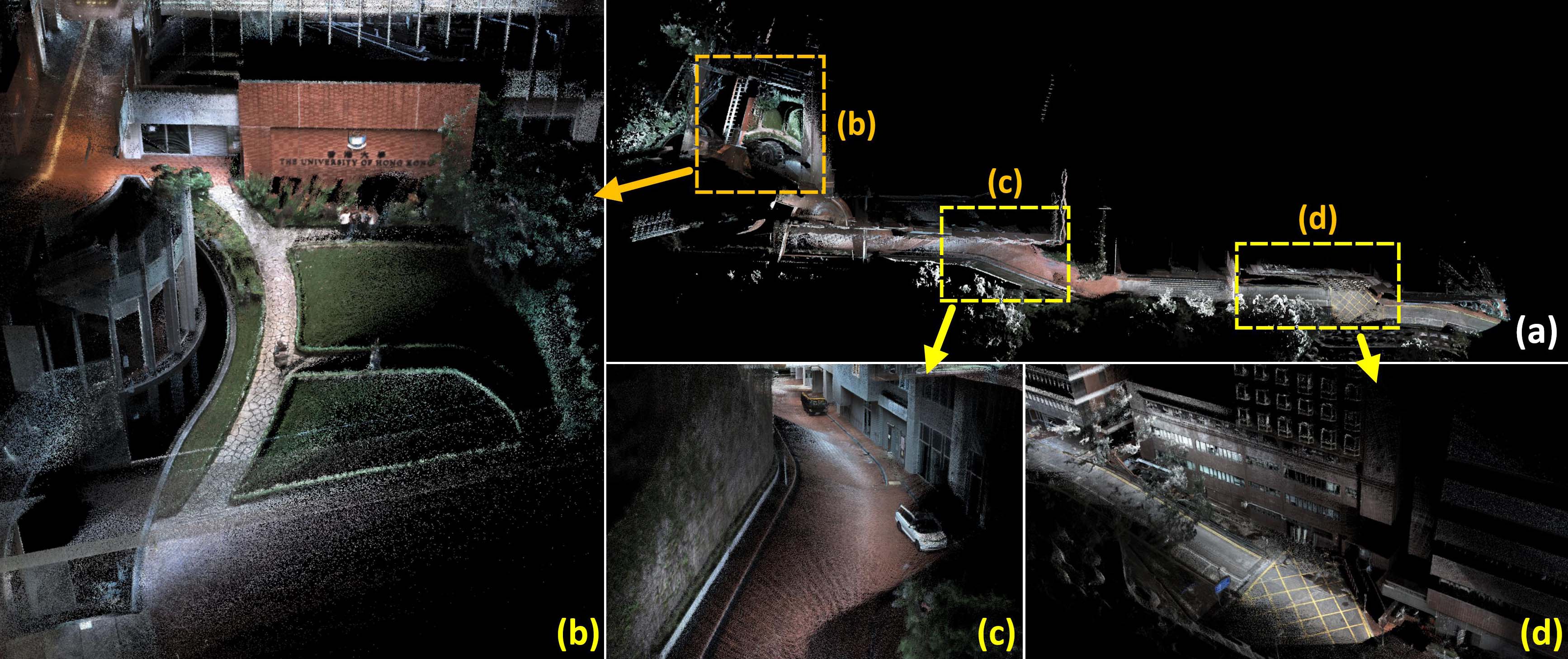}
	\caption{Sequence ``\textbf{hku\_campus\_seq\_01}" are collected by walking along the drive way of the HKU campus. (a) is the birdview of the whole radiance map, with its details shown in (b$\sim$ d). }
	\vspace{0.4cm}
	\includegraphics[width=1.0\linewidth]{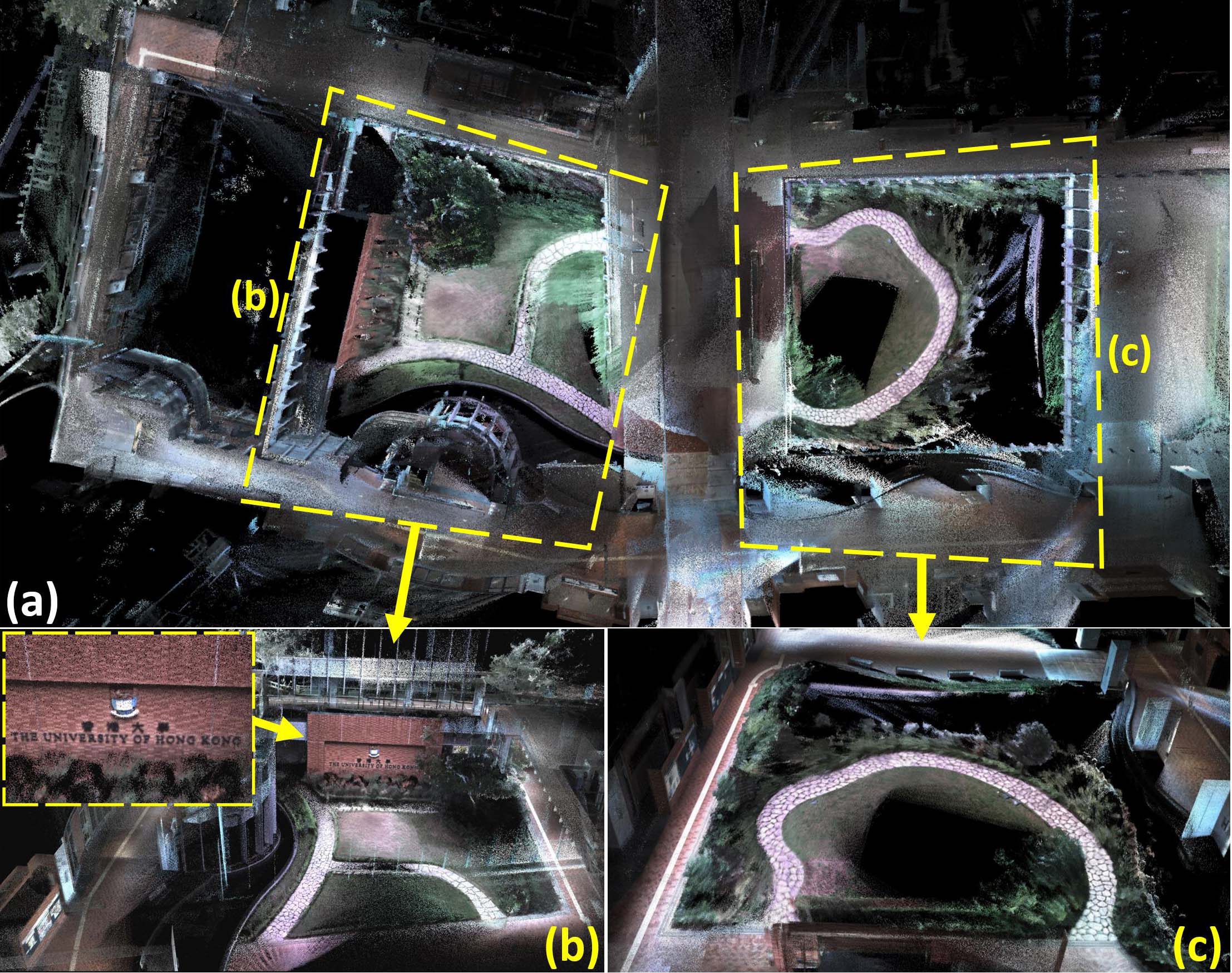}
	\caption{Sequence ``\textbf{hku\_campus\_seq\_00/02/03}" are sampled at the same place but at different times of day (evening, noon and morning, respectively) and with different traveling trajectories. (a) is the birdview of map of sequence ``\textbf{hku\_campus\_seq\_02}", with the closeup view of details are shown in (b) and (c).}
	\end{minipage}
\end{figure}

\begin{figure}[h]
	\centering
	\begin{minipage}{0.85\linewidth}
	\includegraphics[width=1.0\linewidth]{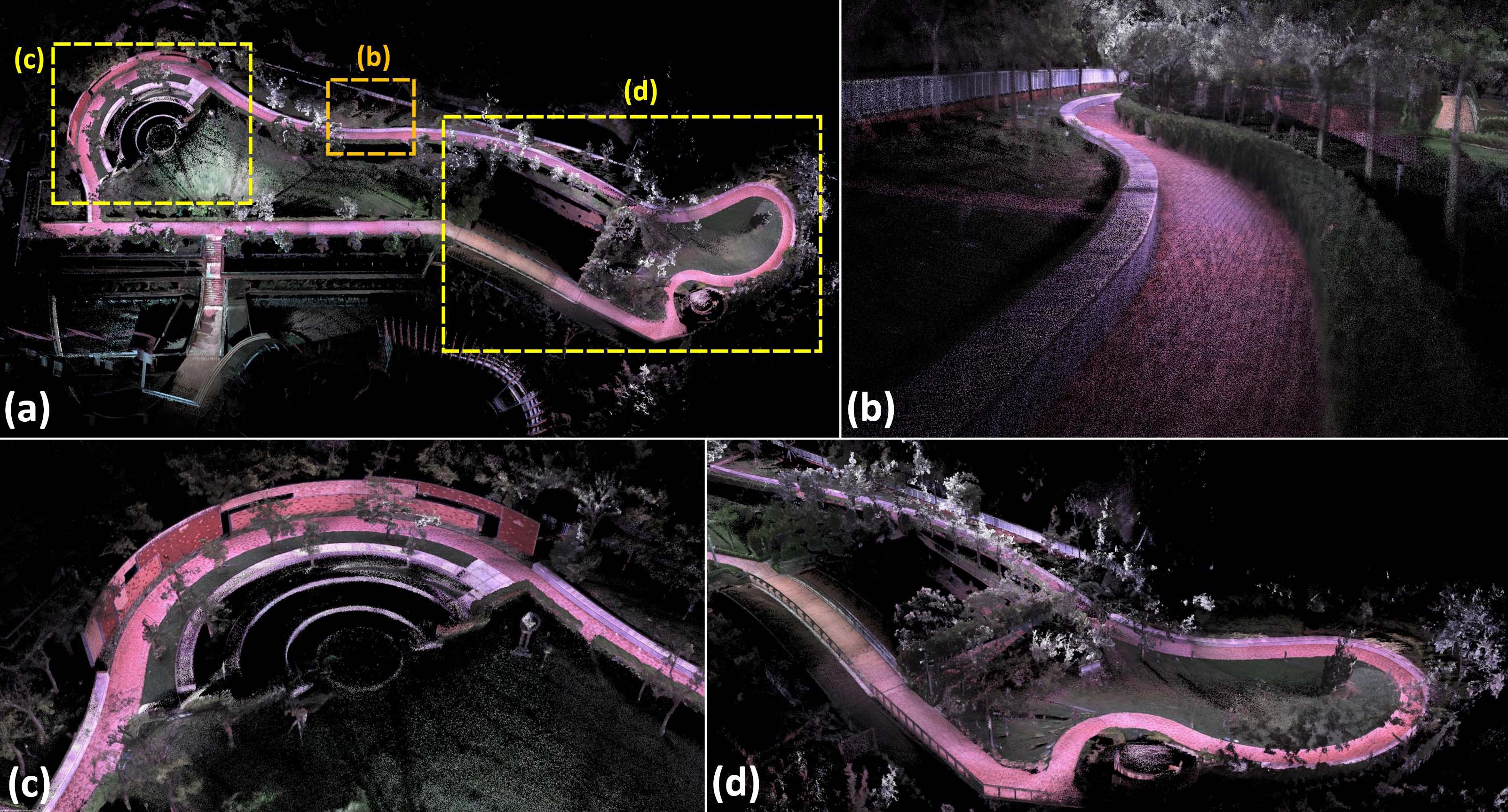}
	\caption{Sequence ``\textbf{hku\_park\_00}" is collected by walking along the pathway of a garden of HKU. (a) is the birdview of the whole radiance map, with its details shown in (b$\sim$ d).}	
	\vspace{0.4cm}
	\includegraphics[width=1.0\linewidth]{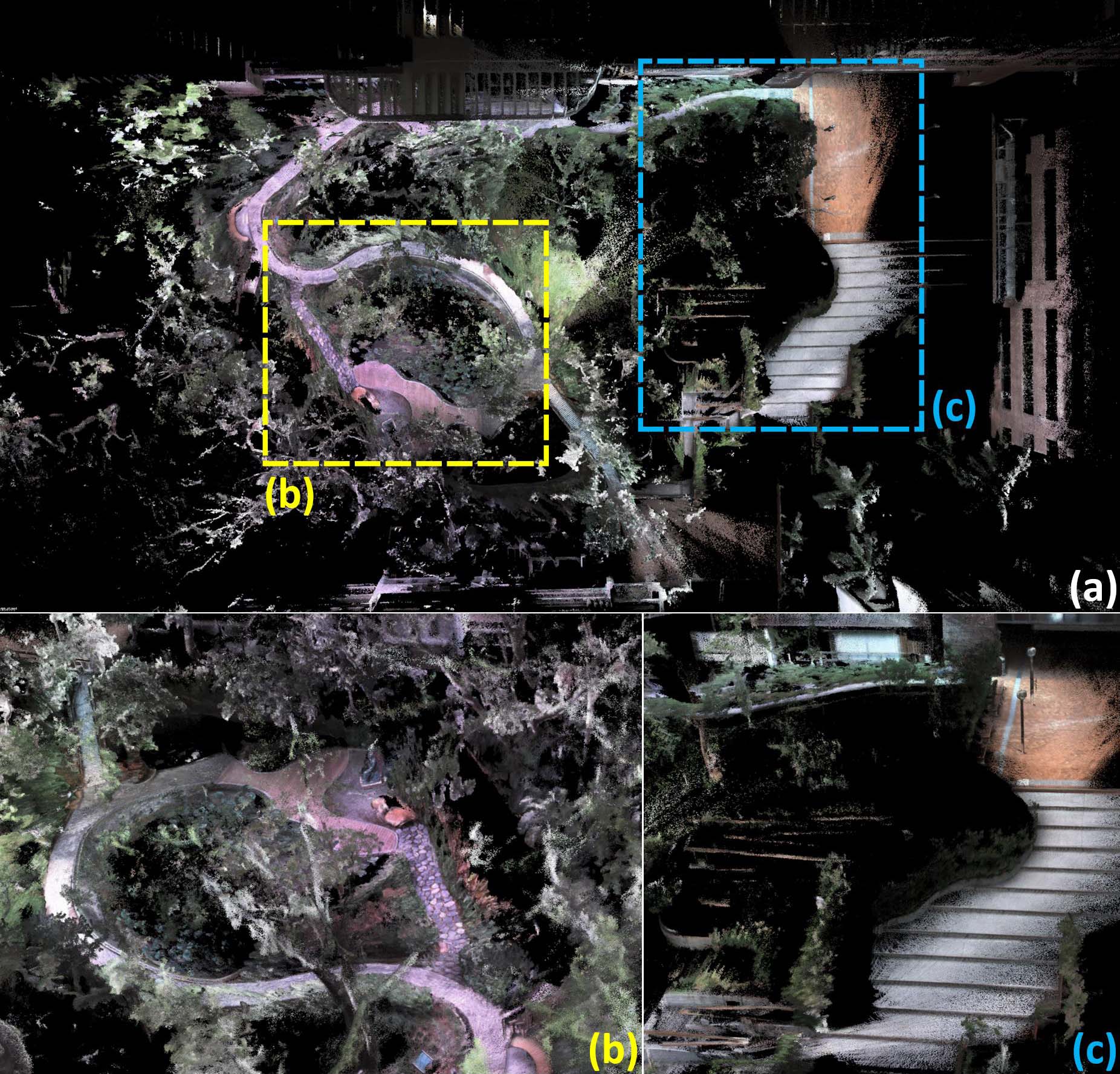}
	\caption{Sequence ``\textbf{hku\_park\_01}" is collected in a cluttered environment with many trees and bushes. (a) is the birdview of the whole radiance map, with its details are shown in (b) and (c).}
	\end{minipage}
\end{figure}

\begin{figure}[h]
	\centering
	\begin{minipage}{0.85\linewidth}
	\includegraphics[width=1.0\linewidth]{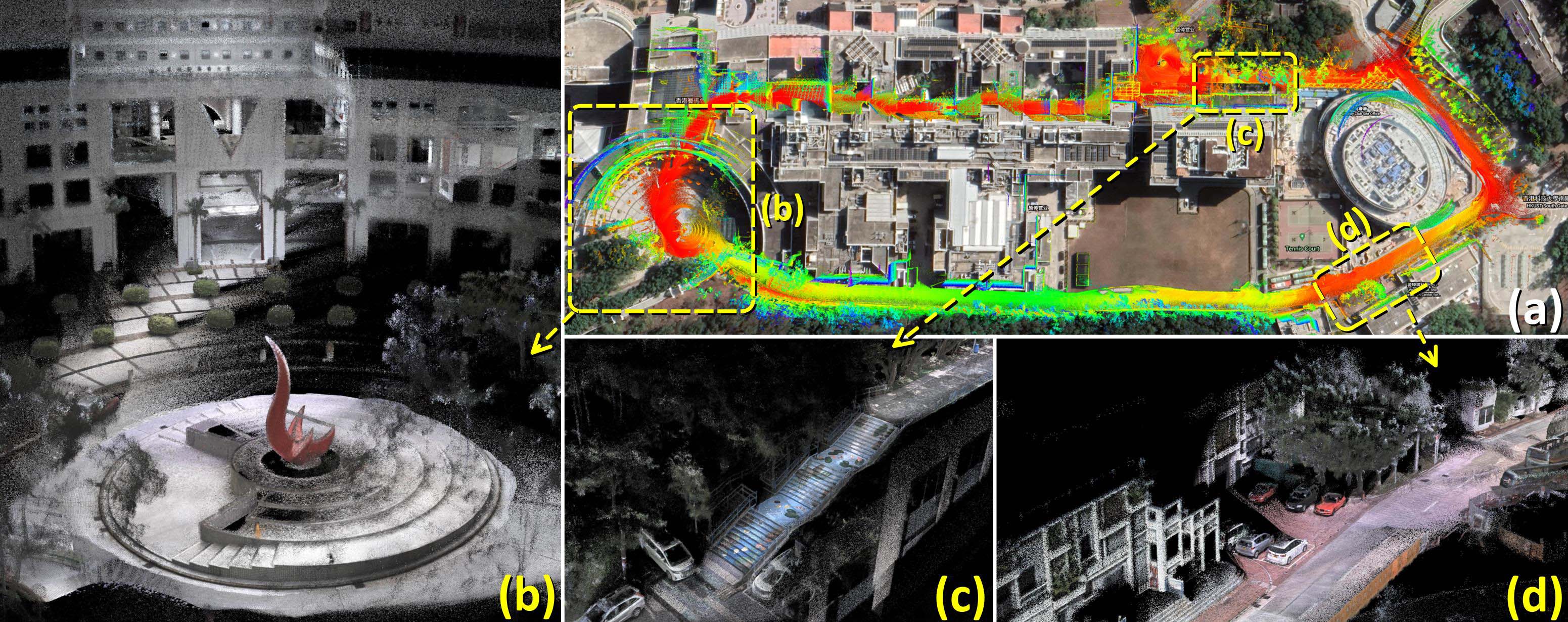}
	\caption{Sequence ``\textbf{hku\_campus\_seq\_00/01}" are collected within the campus of HKUST with two different traveling trajectories. In (a), we merge the point cloud of sequence ``\textbf{hku\_campus\_seq\_00}" with the GoogleEarth satellite image and find them aligned well. The details of our reconstructed radiance map are selectively shown in (b$\sim$d).}
	\vspace{0.4cm}
	\includegraphics[width=1.0\linewidth]{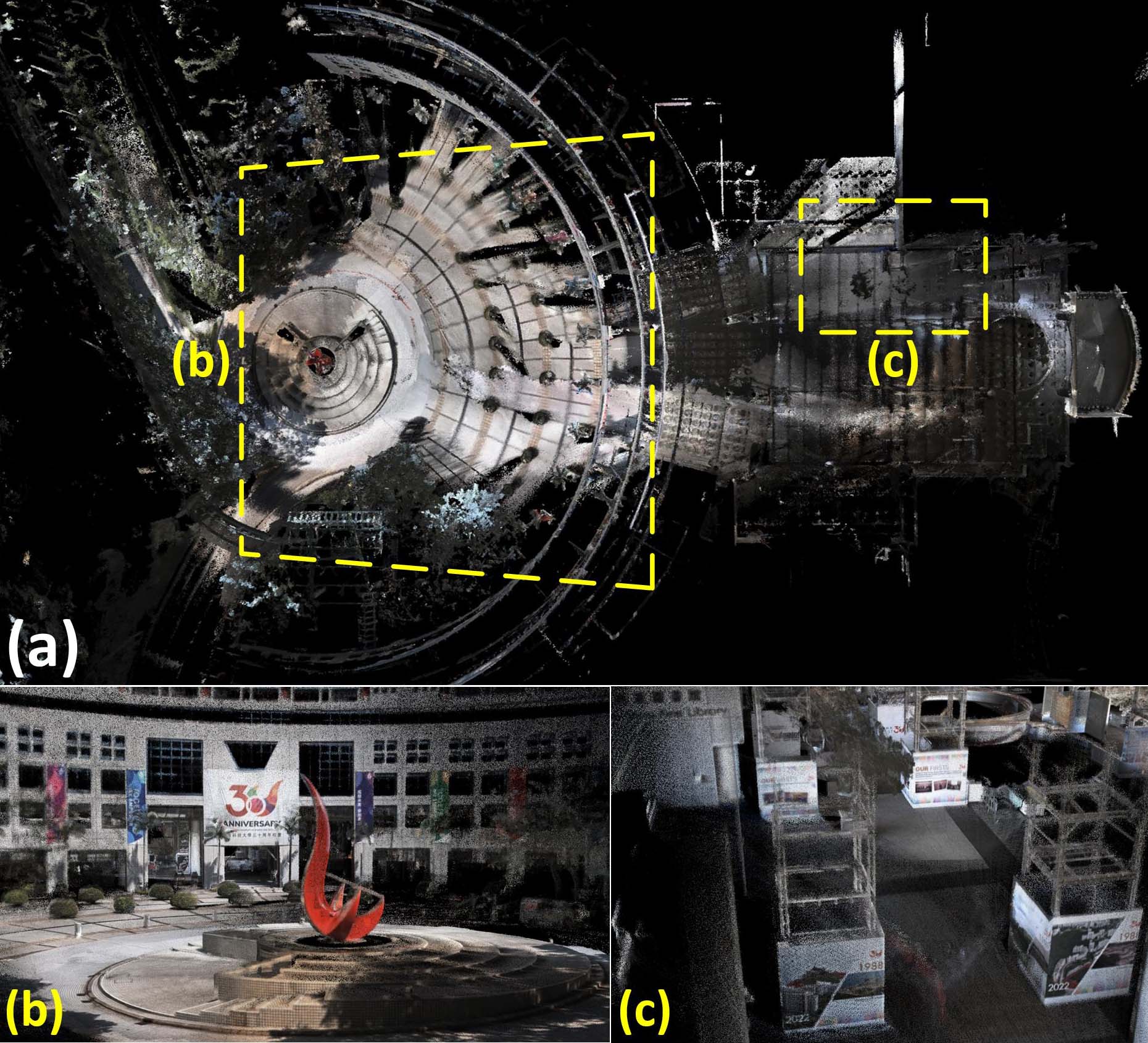}
	\caption{Sequence ``\textbf{hku\_campus\_seq\_02}" is collected by exploring the entrance piazza of HKUST, traveling both the interior and exterior of the buildings. (a) is the birdview of the whole radiance map, with the outdoor and indoor scenarios selectively shown in (b) and (c), respectively.}
	\end{minipage}
\end{figure}

\newpage

\newpage
\begin{figure}[ht]
	\centering
	\begin{turn}{-90}
		\begin{minipage}{1.4\linewidth}
			\centering
			\includegraphics[width=1.0\linewidth]{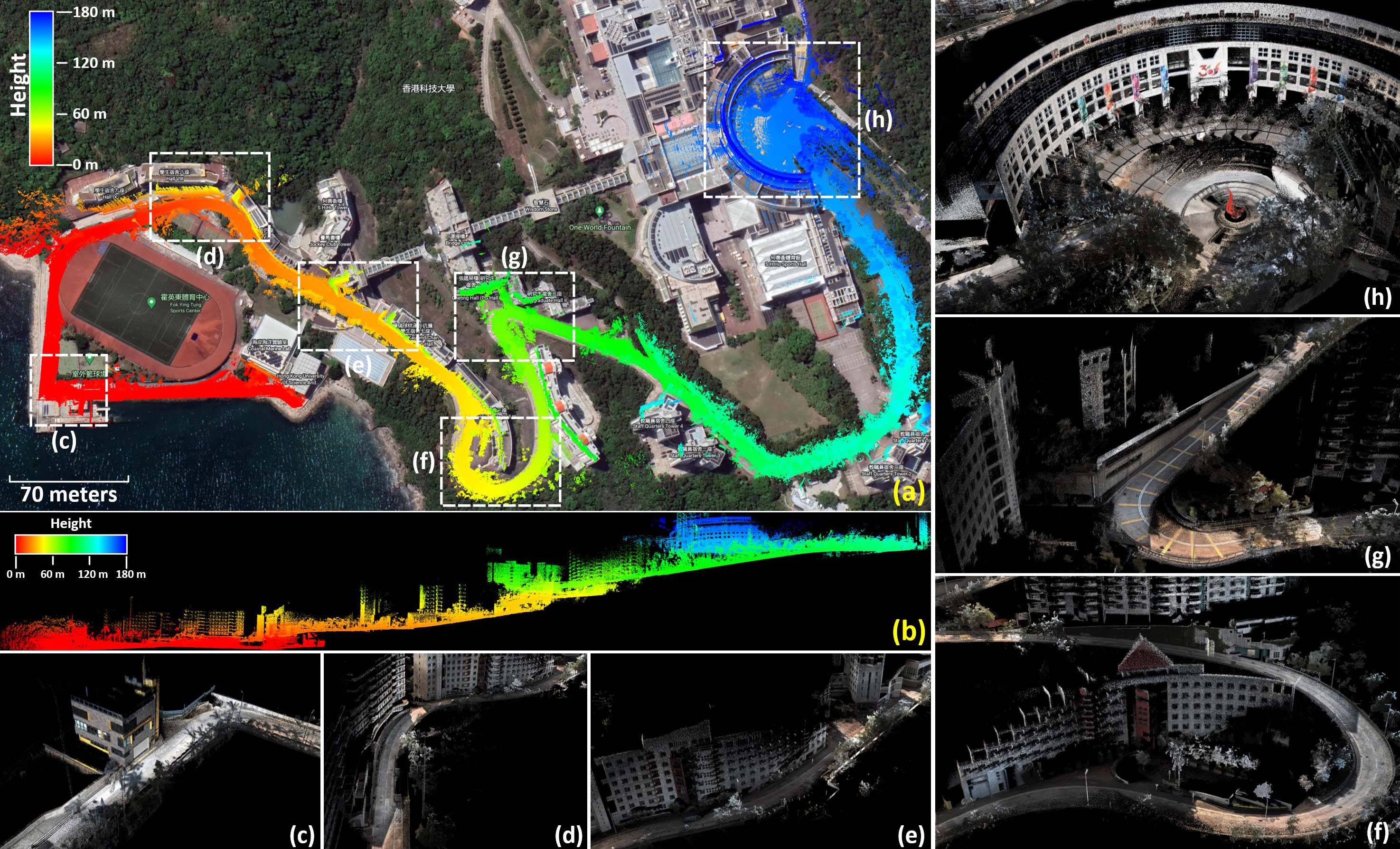}
			\caption{Sequence ``\textbf{hkust\_campus\_seq\_03}" captures most part of the HKUST's campus,  with the traveling length reaching \SI{2.1}{\kilo\meter}. We collected the data starting from the sea front (see the lower left of (a))  and ending at the entrance piazza (the upper right of (a)) of HKUST. In (a), we merge our reconstructed point cloud map (points are colored by their heigh)  with the Google Earth satellite image and find them aligned well. (b)  shows the side view of the map. (c$\sim$h) are the closeup views of the details marked in (a). To see the real-time reconstruction process of the map, please refer to the video on YouTube: 
			\href{https://youtu.be/qXrnIfn-7yA?t=261}{\tt{youtu.be/qXrnIfn-7yA?t=261}}.}
			\label{fig:bear}
		\end{minipage}
	\end{turn}
\end{figure}

\twocolumn

\iftrue

\fi




\end{document}